\documentclass[runningheads]{llncs}
\usepackage{graphicx}
\usepackage{amsmath,amssymb} % define this before the line numbering.
\usepackage{color}
\usepackage{times}
\usepackage{epsfig}
\usepackage{tabularx}
\usepackage{subcaption}
\usepackage{multirow}
\usepackage[ruled,lined,longend,linesnumbered,noend,boxed]{algorithm2e}
\usepackage[table]{xcolor}
\usepackage{enumerate}

\SetCommentSty{mycommfont}
\SetKwComment{tcp}{$\triangleright$}{}
\usepackage[page]{appendix}
\usepackage{stackengine}
\usepackage{hyperref}
\graphicspath{{figures/}}

\DeclareMathOperator*{\argmin}{argmin}
\newcommand{\mes}[2]{m^{#1}_{#2}}
\newcommand{\hmes}[2]{\hat{m}^{#1}_{#2}}
\newcommand{\mesnew}[3]{m^{#1}_{#2}({#3})}
\newcommand{\hmesnew}[3]{\hat{m}^{#1}_{#2}({#3})}
\newcommand{\minisubcaption}[2]{\centering \text{(#1) \scriptsize #2}}
\newcommand{\lgray}{\cellcolor{gray!30}}
\def\myref#1{{\color{black}{#1}}}
\def\eqref#1{Eq.~{\color{black}{(\ref{#1})}}}
\def\plaineqref#1{{\color{black}{(\ref{#1})}}}
\def\myfig#1{Figure~{\color{black}{\ref{#1}}}}
\def\mytb#1{Table~{\color{black}{\ref{#1}}}}
\def\myblue#1{{\color{black}{#1}}}
\def\myalg#1{Algorithm~{\color{black}{\ref{#1}}}}
\def\ie{\textit{i.e.}}
\def\etc{\textit{etc.}}
\newcolumntype{L}[1]{>{\raggedright\let\newline\\\arraybackslash}m{#1}}
\newcolumntype{C}[1]{>{\centering\let\newline\\\arraybackslash}m{#1}}
\newcolumntype{R}[1]{>{\raggedleft\let\newline\\\arraybackslash}m{#1}}
\hypersetup{colorlinks=true, linkcolor=red, filecolor=magenta, urlcolor=blue}
\newif\ifsupp
\suppfalse  % option 2

\begin{document}
\pagestyle{headings}
\mainmatter

\ifsupp
    \title{Fast and Differentiable Message Passing on \\Pairwise Markov Random Fields \\ -- Supplementary Material --}
\else
    \title{Fast and Differentiable Message Passing on \\Pairwise Markov Random Fields}
\fi

\titlerunning{Fast and Differentiable Message Passing on Pairwise Markov Random Fields}
\authorrunning{Z. Xu et al.}
\author{Zhiwei Xu\inst{1,2}\orcidID{0000-0001-8283-6095} \and
Thalaiyasingam Ajanthan\inst{1}\orcidID{0000-0002-6431-0775} \and
Richard Hartley\inst{1}\orcidID{0000-0002-5005-0191}}
\institute{Australian National University and Australian Centre for Robotic Vision \and
Data61, CSIRO, Canberra, Australia\\
\email{\{firstname.lastname\}@anu.edu.au}}
\maketitle
\ifsupp
    \setcounter{equation}{13}
    \setcounter{figure}{6}
    \setcounter{table}{5}
    \setcounter{algocf}{2}
    \appendix
    
%==========================================================================
\section{Pseudocode of Backpropagation of ISGMR and TRWP}

Due to the limited space of the main paper, we provide
the pseudocode of backpropagation of ISGMR and TRWP in this appendix, in
Algorithms~\ref{alg:isgmr-back}-\ref{alg:trwp-back} respectively.

\setlength{\textfloatsep}{0.5ex}
\begin{algorithm}
\KwIn{Partial energy parameters $\{\theta_{i,j}\}$, gradients of final costs
$\mathbf{\nabla c} = \{\nabla c_{i}{(\lambda)}\}$, set of nodes $\mathcal{V}$,
edges $\mathcal{E}$, directions $\mathcal{R}$, indices $\{p^r_{k,i}{(\lambda)}\}$,
$\{q^r_{k,i}\}$, iteration number $K$.
We replace $\nabla m^{r,k+1}$ by $\nabla \hat{m}^r$ and
$\nabla m^{r,k}$ by $\nabla m^r$ for simplicity.}
\KwOut{Gradients $\{\nabla \theta_{i}, \nabla \theta_{i,j}(\cdot, \cdot)\}$.}
\caption{Backpropagation of ISGMR} \label{alg:isgmr-back}
$\nabla \mathbf{m}^{\mathbf{r}} \gets \nabla \mathbf{\Theta_i} \gets \nabla
\mathbf{c}$, $\nabla \mathbf{\Theta_{i,j}} \gets 0$
% \tcp*[f]{back \eqref{eq:isgmr-aggregation}}
\tcp*[f]{back Eq. (7)}
\\
$\nabla \hat{\mathbf{m}}^{\mathbf{r}} \gets \nabla \mathbf{m}^{\mathbf{r}}$
\tcp*[f]{back message updates} \\
\For{\textup{iteration $k \in \{K, ..., 1\}$}}{
  $\nabla \mathbf{m}^{\mathbf{r}} \gets 0$ \tcp*[f]{zero-out} \\
    \ForAll(\tcp*[f]{\textbf{parallel}}){\textup{directions $r \in
\mathcal{R}$}}{
      \ForAll(\tcp*[f]{\textbf{parallel}}){\textup{scanlines $t$ in direction
$r$}}{
        \For(\tcp*[f]{\textbf{sequential}}){\textup{node $i$ in scanline $t$}}{
          $\lambda^* \gets q^r_{k,i} \in \mathcal{L}$ \tcp*[f]{extract index}
\\
          $\nabla \hmesnew{r}{i}{\lambda^*} \mathrel{-}= \sum_{\lambda \in
\mathcal{L}}{\nabla \hmesnew{r}{i}{\lambda}}$
% \tcp*[f]{back \eqref{eq:isgmr-norm}}
\tcp*[f]{back Eq. (5)}
\\
          \For{\textup{label $\lambda \in \mathcal{L}$}}{
            $\mu^* \gets p^r_{k,i}{(\lambda)} \in \mathcal{L}$ \tcp*[f]{extract
index} \\
            $\nabla \theta_{i-r}{(\mu^*)} \mathrel{+}= \nabla
\hmesnew{r}{i}{\lambda}$
% \tcp*[f]{back \eqref{eq:isgmr}}
\tcp*[f]{back Eq. (9)}
\\
            $\nabla \hmesnew{r}{i-r}{\mu^*} \mathrel{+}= \nabla
\hmesnew{r}{i}{\lambda}$ \\
            $\nabla \mesnew{d}{i-r}{\mu^*} \mathrel{+}= \nabla
\hmesnew{r}{i}{\lambda}, \forall d \in \mathcal{R} \setminus \{r,r^-\}$ \\
            $\nabla \theta_{i-r,i}(\mu^*, \lambda) \mathrel{+}= \nabla
\hmesnew{r}{i}{\lambda}$ \\
          }
        }
      }
    $\nabla \hat{\mathbf{m}}^r \gets 0$ \tcp*[f]{zero-out} \\
  }
  $\nabla \mathbf{m^r} \mathrel{+}= \nabla \hat{\mathbf{m}}^\mathbf{r}$
\tcp*[f]{gather history gradients} \\
  $\nabla \hat{\mathbf{m}}^\mathbf{r} \gets \nabla \mathbf{m^r}$ \tcp*[f]{back
message updates after iteration} \\
}
\end{algorithm}

\setlength{\textfloatsep}{0.5ex}
\begin{algorithm}
\KwIn{Partial energy parameters $\{\theta_{i,j}\}$, gradients of final costs
$\mathbf{\nabla c} = \{\nabla c_{i}{(\lambda)}\}$,  tree decomposition coefficients
$\{\rho_{i,j}\}$, set of nodes $\mathcal{V}$, edges $\mathcal{E}$, directions
$\mathcal{R}$, indices $\{p^r_{k,i}{(\lambda)}\}$, $\{q^r_{k,i}\}$, iteration number $K$.}
\KwOut{Gradients $\{\nabla \theta_{i}, \nabla \theta_{i,j}(\cdot, \cdot)\}$.}
\caption{Backpropagation of TRWP} \label{alg:trwp-back}
$\nabla \mathbf{m}^{\mathbf{r}} \gets \nabla \mathbf{\Theta_i} \gets
d\mathbf{c}$, $d\mathbf{\Theta_{i,j}} \gets 0$
% \tcp*[f]{back \eqref{eq:isgmr-aggregation}}
\tcp*[f]{back Eq. (7)}
\\
\For{\textup{iteration $k \in \{K, ..., 1\}$}}{
  \For(\tcp*[f]{\textbf{sequential}}){\textup{direction $r \in \mathcal{R}$}}{
    \ForAll(\tcp*[f]{\textbf{parallel}}){\textup{scanlines $t$ in direction
$r$}}{
      \For(\tcp*[f]{\textbf{sequential}}){\textup{node $i$ in scanline $t$}}{
        $\lambda^* \gets q^{r}_{k,i} \in \mathcal{L}$ \tcp*[f]{extract index}
\\
        $\nabla \mesnew{r}{i}{\lambda^*} \mathrel{-}= \sum_{\lambda \in
\mathcal{L}}{\nabla \mesnew{r}{i}{\lambda}}$
% \tcp*[f]{back \eqref{eq:isgmr-norm}}
\tcp*[f]{back Eq. (5)}
\\
        \For{\textup{label $\lambda \in \mathcal{L}$}}{
          $\mu^* \gets p^{r}_{k,i}{(\lambda)} \in \mathcal{L}$ \tcp*[f]{extract
index} \\
          $\nabla \theta_{i-r}{(\mu^*)} \mathrel{+}= \rho_{i-r,i} \nabla
\mesnew{r}{i}{\lambda}$
% \tcp*[f]{back \eqref{eq:trwp}}
\tcp*[f]{back Eq. (11)}
\\
          $\nabla \mesnew{d}{i-r}{\mu^*} \mathrel{+}= \rho_{i-r,i} \nabla
\mesnew{r}{i}{\lambda}, \forall d \in \mathcal{R}$ \\
          $\nabla \mesnew{r^-}{i-r}{\mu^*} \mathrel{-}= \nabla \mesnew{r}{i}{\lambda}$
\\
          $\nabla \theta_{i-r,i}(\mu^*, \lambda) \mathrel{+}= \nabla
\mesnew{r}{i}{\lambda}$ \\
        }
      }
    }
    $\nabla \mathbf{m}^{r} \gets 0$ \tcp*[f]{zero-out} \\
  }
}
\end{algorithm}

%==========================================================================
\section{Maintaining Energy Function in Iterations}
With the same notations in Eq.\myblue{(1)} and Eq.\myblue{(9)} in the
main paper, let a general energy function in a MRF defined as

\begin{equation}
\label{eq:supp-energy-function}
\begin{aligned}
  E(\mathbf{x} | \mathbf{\Theta}) = \sum_{i \in \mathcal{V}} \theta_{i}(x_i)
  + \sum_{(i, j) \in \mathcal{E}} \theta_{i,j}(x_i, x_j)\ .
\end{aligned}
\end{equation}
In the standard SGM and ISGMR, given a node $i$ and an edge from nodes $j$
to $i$, the message will be updated at $k$th iteration as follows,

\begin{equation}
\centering
\label{eq:supp-isgmr}
\begin{aligned}
\mesnew{r,k+1}{i}{\lambda} = \min_{\mu \in \mathcal{L}}
\big(\theta_{i-r}{(\mu)}
+ \theta_{i-r,i}(\mu,\lambda)
+ \mesnew{r,k+1}{i-r}{\mu} &+ \sum_{d \in \mathcal{R} \setminus \{r,r^-\}}
\mesnew{d,k}{i-r}{\mu}\big) \ .
\end{aligned}
\end{equation}
In \myfig{fig:supp-passing}, however, if we add a term $m_{i}{(\lambda)}$
to node $i$ at label $\lambda$ via $m_{ji}{(\lambda)}$
from node $j$ to node $i$ at label $\lambda$, the same value should
be subtracted along all edges connecting this node $i$, that is $\forall (i,j)
\in \mathcal{E}$, in order to maintain the
same \eqref{eq:supp-energy-function} in optimization.
This supports the exclusion of $r^-$ from $\mathcal{R}$ in \eqref{eq:supp-isgmr}.
This is important for multiple iterations because the non-zero messages after
the 1st iteration, as additional terms, will change the energy function via
\eqref{eq:supp-isgmr}.
Hence, a simple combination of many standard SGMs will change the energy
function due to the lack of the subtraction above.

\begin{figure}[t]
\centering
\includegraphics[width=0.25\linewidth]{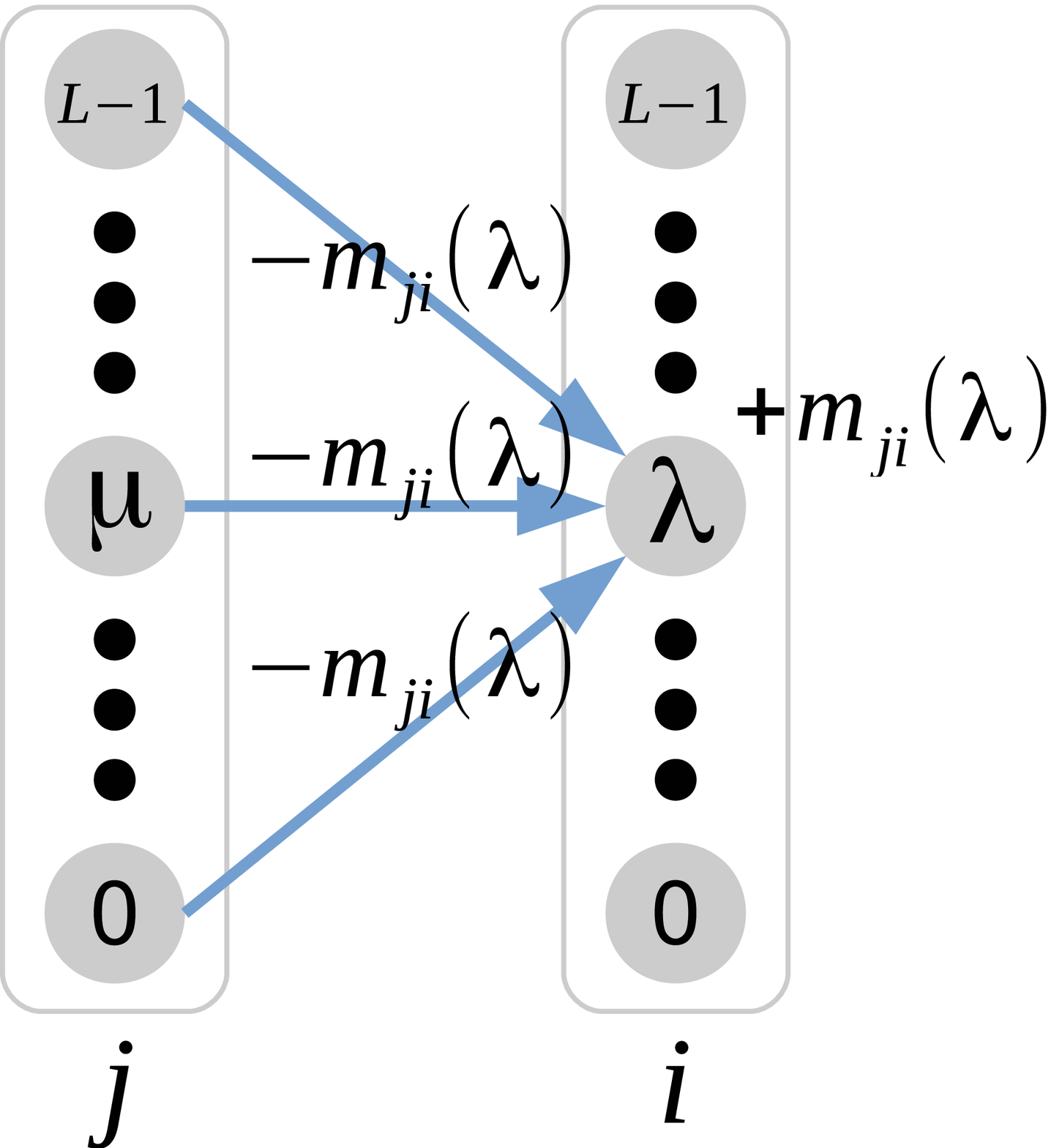}
\caption{Energy function maintained in iterative message passing.
When adding a term $m_{ji}{(\lambda)}$ to node $i$ at label $\lambda$,
the same value should be subtracted on all edges connecting node $i$ at
label $\lambda$.}
\label{fig:supp-passing}
\end{figure}

%==========================================================================
\section{Indexing First Nodes by Interpolation}
Tree graphs contain horizontal, vertical, and diagonal (including symmetric,
asymmetric wide, and asymmetric narrow) trees, shown in
\myfig{fig:supp-paths}.
Generally, the horizontal and vertical trees are for 4-connected graphs,
symmetric trees are for 8-connected graphs, and asymmetric
trees are for more
than 8-connected graphs, resulting in different ways of indexing the
first nodes
for parallelization. In the following, we denote an image size with height
$H$ and
width $W$, coordinates of the first node in vertical and horizontal
directions
as $p_h$ and $p_w$ respectively, and scanning steps in vertical and
horizontal
directions as $S_h$ and $S_w$ respectively.

\begin{figure}
\centering
\begin{subfigure}[b]{0.15\textwidth}
\includegraphics[width=\textwidth]{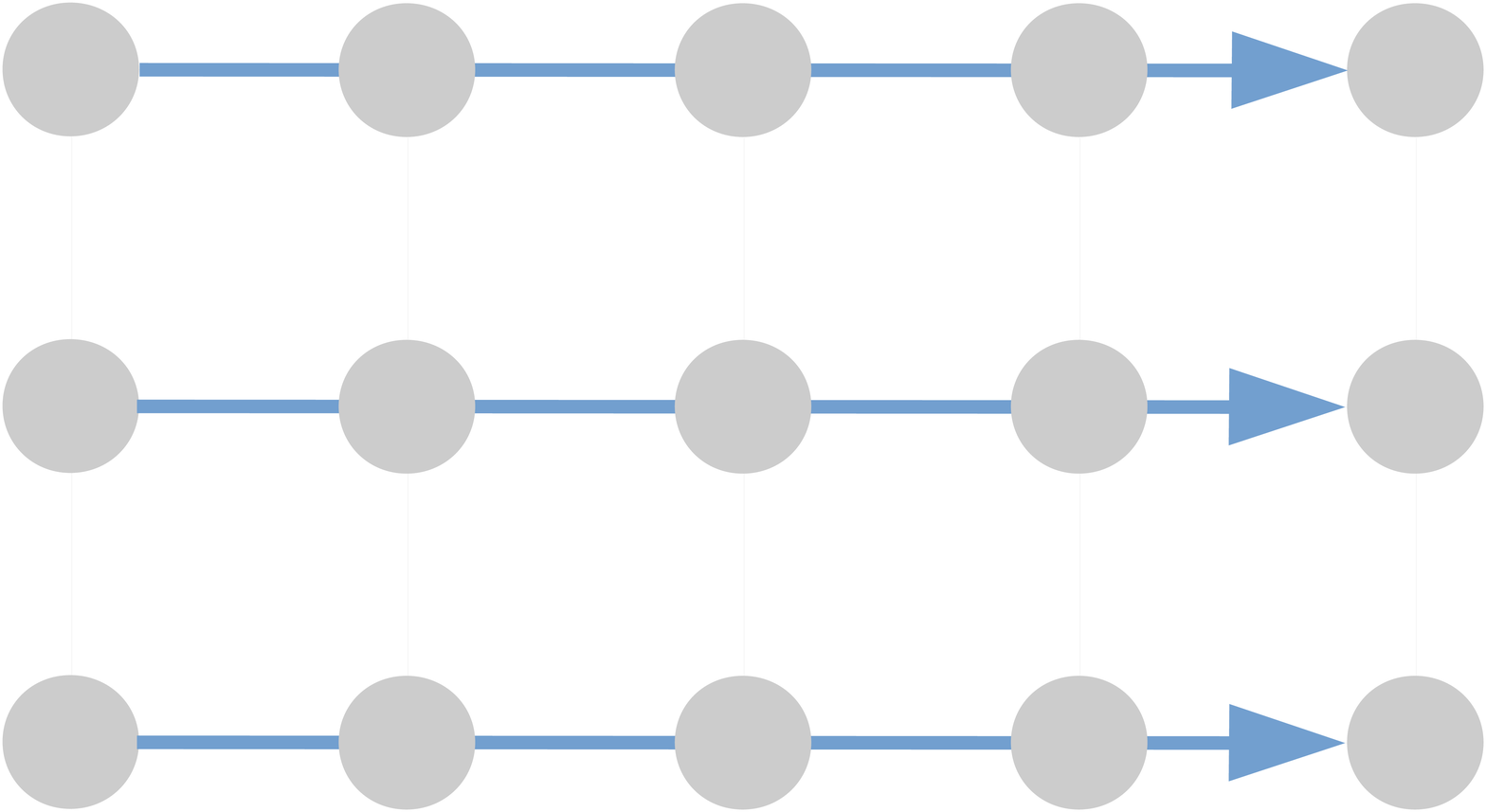}
\caption{}
\end{subfigure}
~
\begin{subfigure}[b]{0.15\textwidth}
\includegraphics[width=\textwidth]{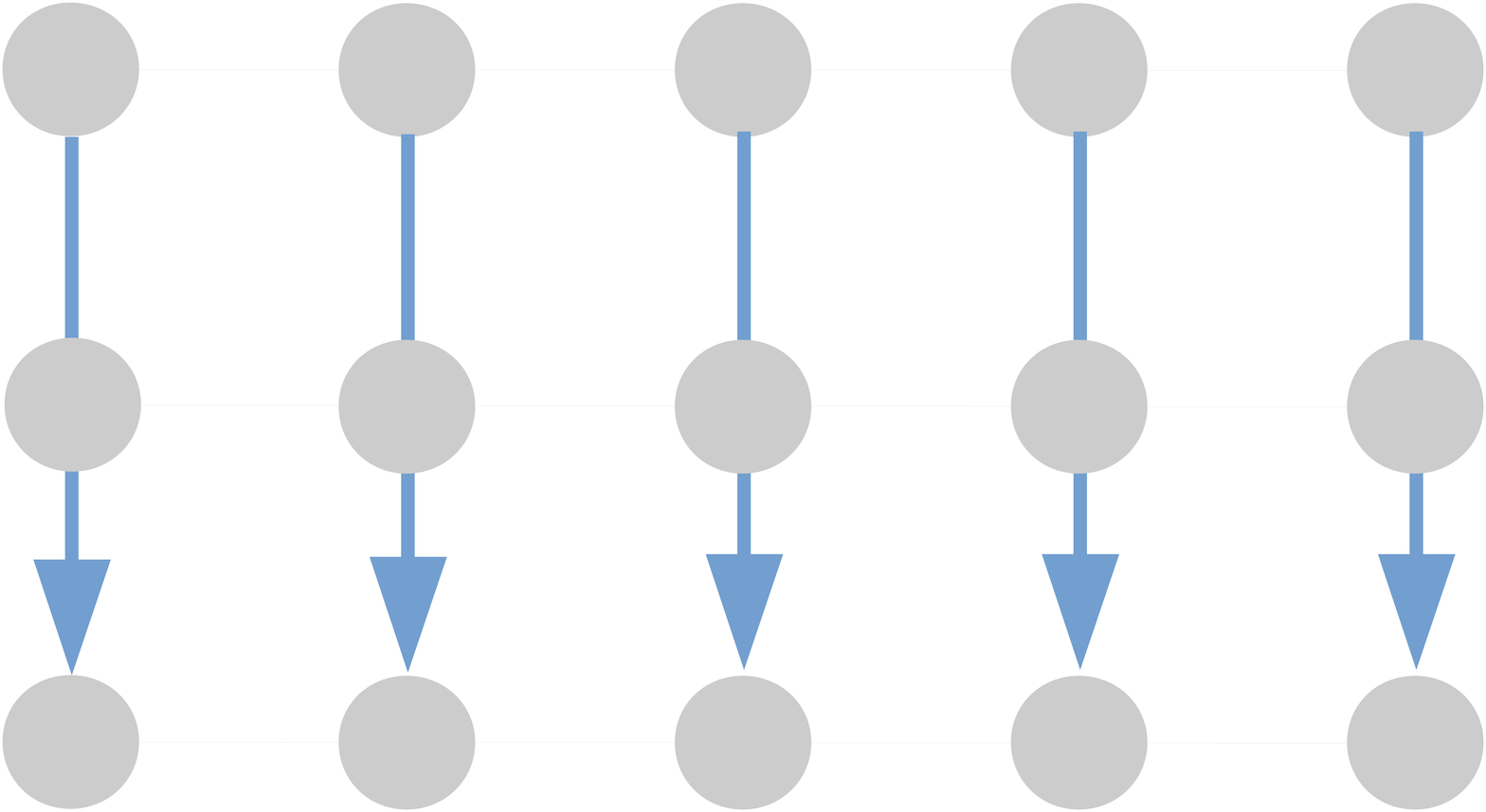}
\caption{}
\end{subfigure}
~
\begin{subfigure}[b]{0.15\textwidth}
\includegraphics[width=\textwidth]{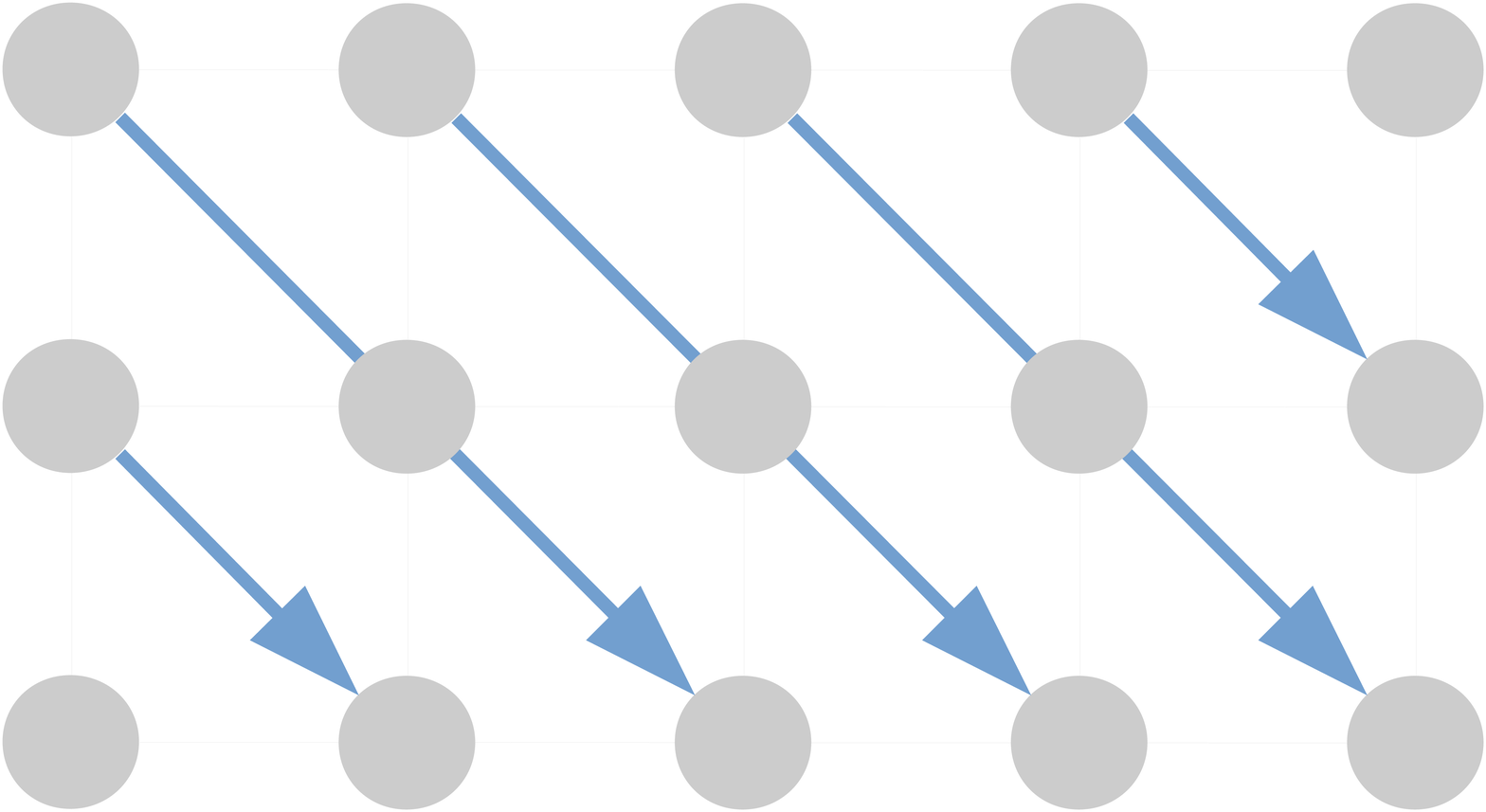}
\caption{}
\end{subfigure}
~
\begin{subfigure}[b]{0.15\textwidth}
\includegraphics[width=\textwidth]{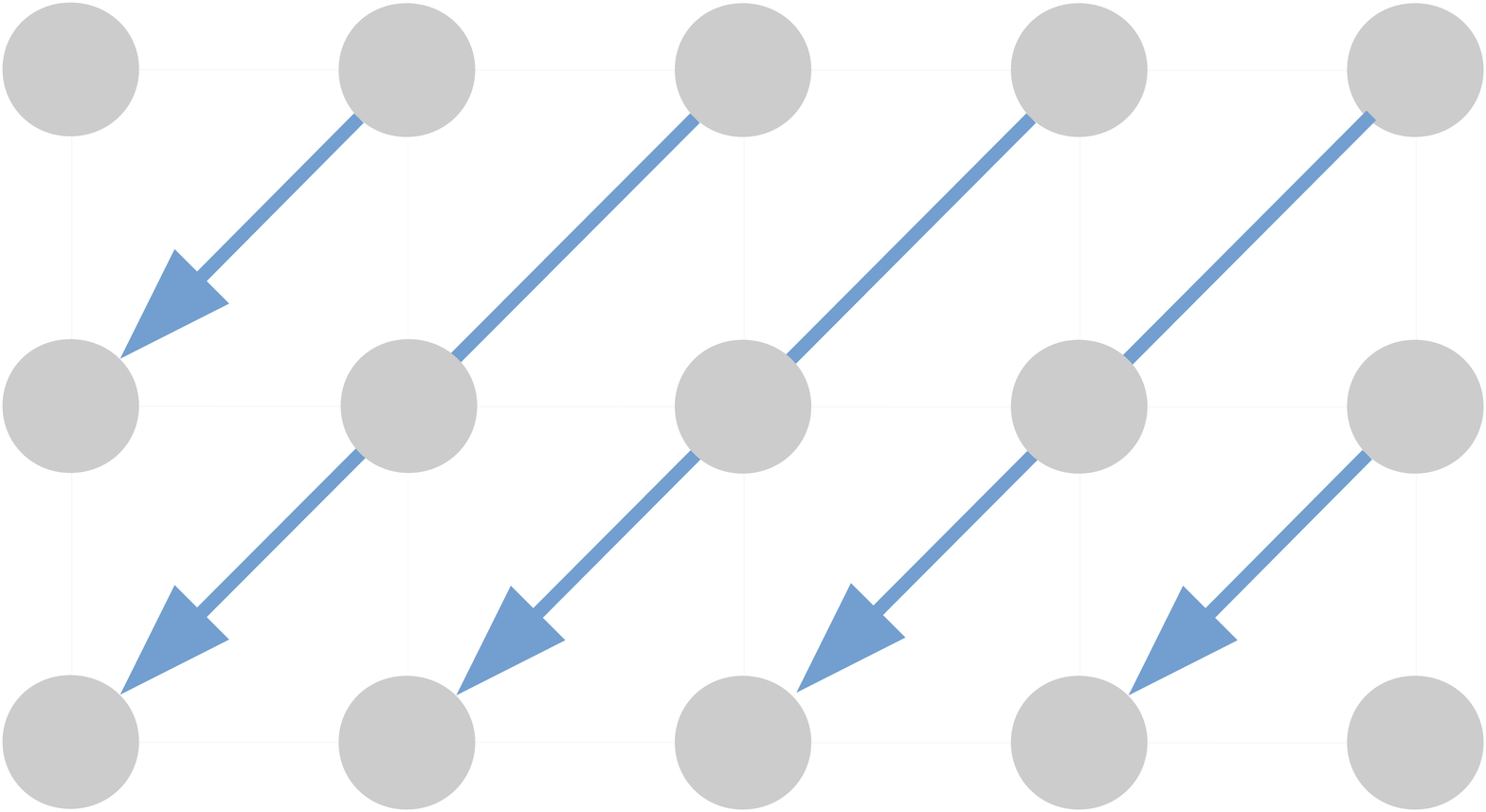}
\caption{}
\end{subfigure}
~
\begin{subfigure}[b]{0.15\textwidth}
\includegraphics[width=\textwidth]{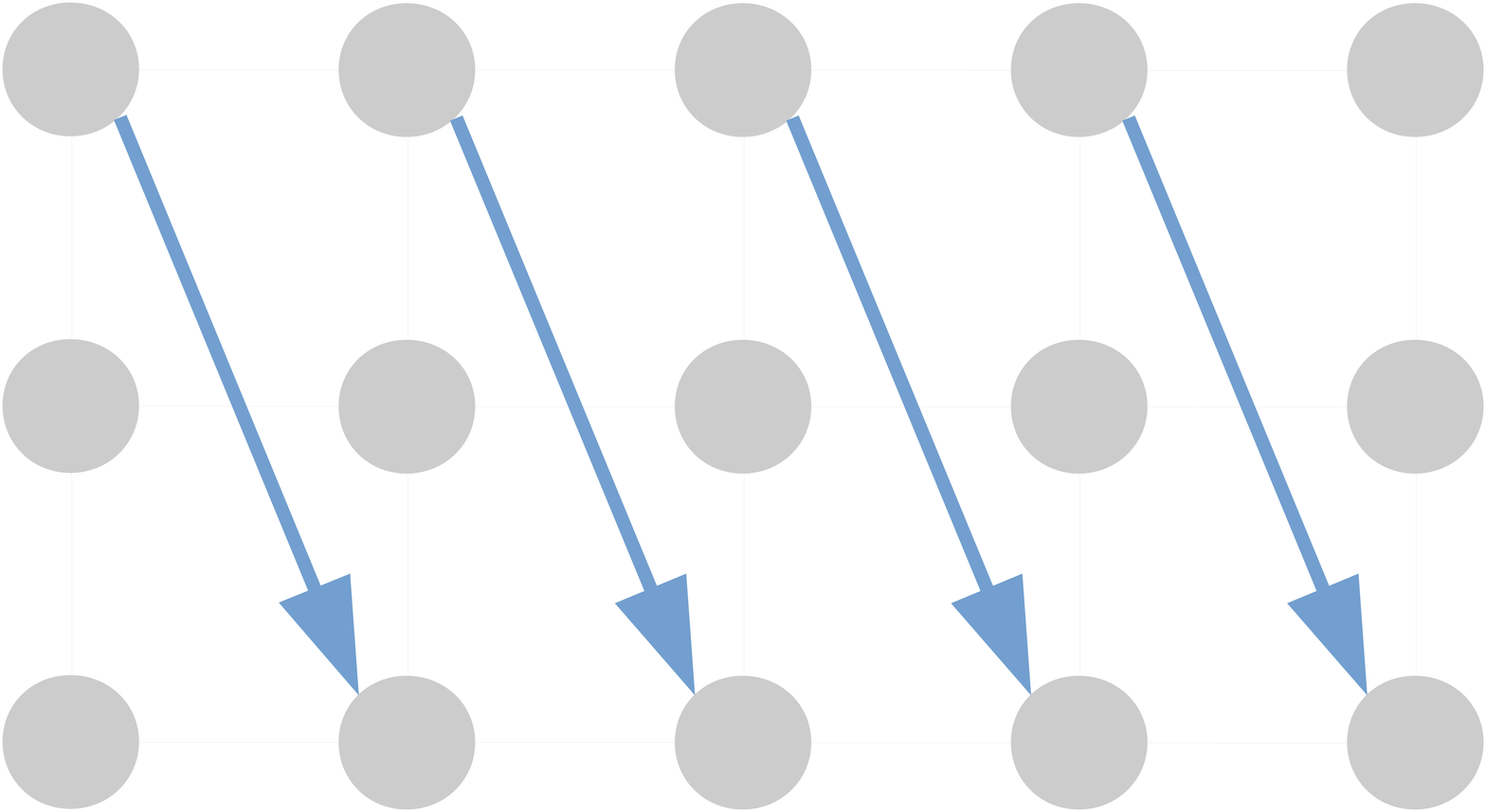}
\caption{}
\end{subfigure}
~
\begin{subfigure}[b]{0.15\textwidth}
\includegraphics[width=\textwidth]{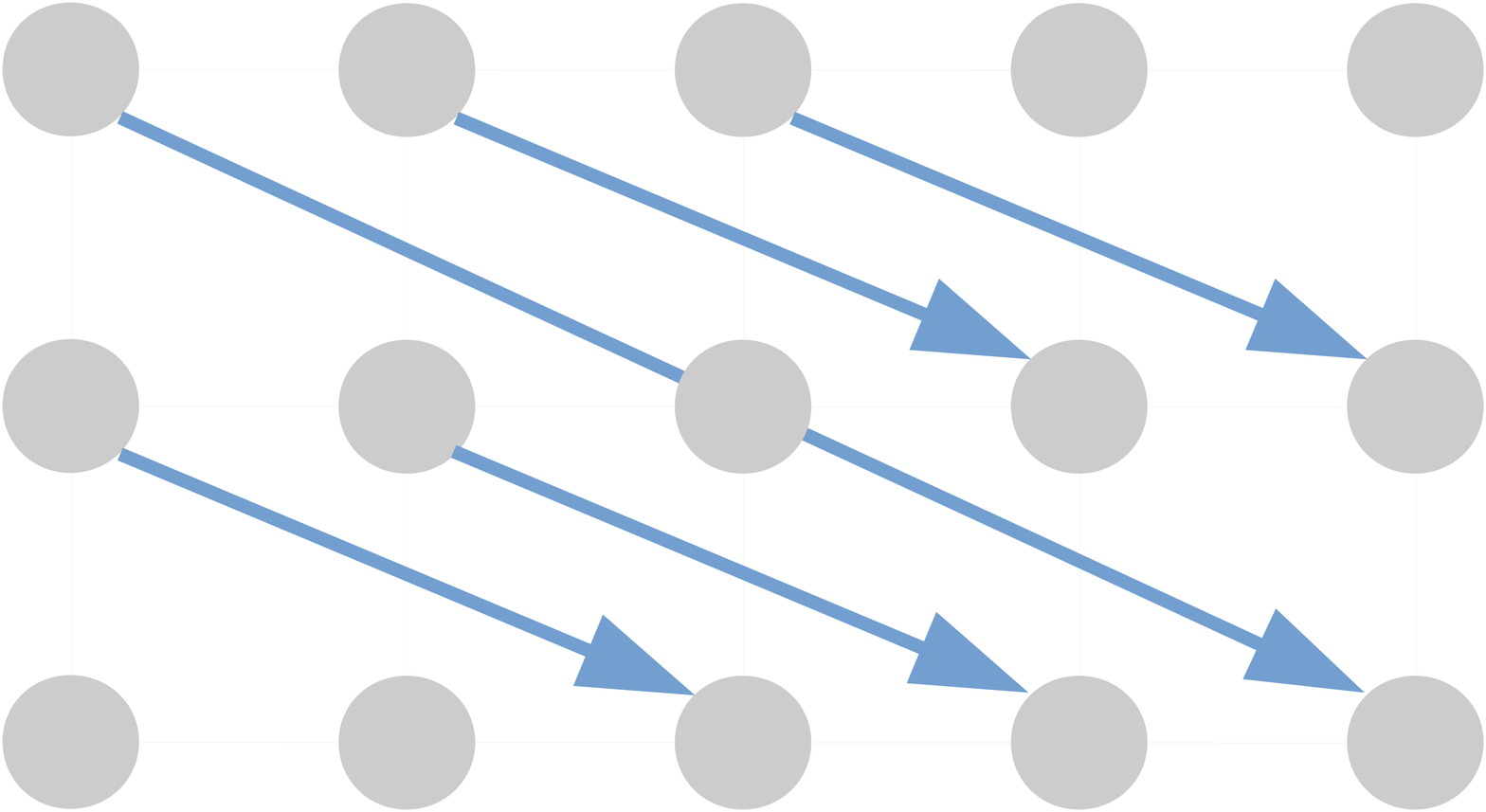}
\caption{}
\end{subfigure}
\caption{Multi-direction message passing(forward passing
in 6 directions).
(a) horizontal trees. (b) vertical trees. (c) symmetric trees
from up-left to down-right.
(d) symmetric trees from up-right to down-left. (e) asymmetric
narrow trees with height
and width steps $S=(S_h, S_w)=(2, 1)$. (f) asymmetric wide trees
with $S=(1, 2)$.}
\label{fig:supp-paths}
\end{figure}

\noindent \textbf{Horizontal and vertical graph trees.} Coordinate
of the first node
of a horizontal and vertical tree, $p=(p_h, p_w)$, can be presented
by ($p_h,0$) and
($0,p_w$) respectively in the forward pass, and ($p_h, W-1$) and
($H-1, p_w$) respectively
in the backward pass.

\noindent \textbf{Symmetric and asymmetric wide graph trees.}
Coordinate of the first
node $p=(p_h, p_w)$ is calculated by
\begin{equation}
\label{eq:supp-wide}
\begin{aligned}
  N &= W + (H-1)*\text{abs}(S_w)\ , \\
  p_w &= \left[0:N-1\right] - (H-1) * \max{(S_w, 0)}\ , \\
  p_h &= \begin{cases} 0 & \text{ if } S_h > 0\ ,\\
           H - 1 & \text{otherwise}\ ,
         \end{cases}
\end{aligned}
\end{equation}
where $N$ is tree number, $\text{abs}(*)$ is absolution, and $T_s$
is shifted indices of trees.

\textbf{Asymmetric narrow graph trees.} Coordinate of the first node $p$ by
interpolation is calculated by

\begin{equation}
\label{eq:supp-narrow}
\begin{aligned}
  c_1 &= \text{mod}(T_s, \text{abs}(S_h))\ , \\
  c_2 &= \frac{\text{float}(T_s)}{\text{float}(\text{abs}(S_h))}\ , \\
  p_h &= \begin{cases}
           \text{mod}(\text{abs}(S_h) - c_1, \text{abs}(S_h)) & \text{ if } S_w > 0\ ,\\
           c_1 & \text{otherwise}\ ,
         \end{cases} \\
  p_h &= H - 1 - p_h \text{ if } S_h < 0\ , \\
  p_w &= \begin{cases}
           \text{ceil}(c_2) & \text{ if } S_w > 0\ ,\\
           \text{floor}(c_2) & \text{otherwise}\ ,
         \end{cases}
\end{aligned}
\end{equation}
where $\text{mod}(*)$ is modulo, $\text{floor}(*)$ and $\text{ceil}(*)$ are two
integer approximations, $\text{float}(*)$ is data conversion for single-precision
floating-point values, and the rest share the same notations in \eqref{eq:supp-wide}.

Although ISGMR and TRWP are parallelized over individual trees, message updates
on a tree are sequential. The interpolation for asymmetric diagonals avoids as many
redundant scanning as possible, shown in \myfig{fig:supp-head}. This is more practical
for realistic stereo image pairs that the width is much larger than the height.

\begin{figure}
\begin{subfigure}[b]{0.4\textwidth}
\centering
\includegraphics[width=0.8\textwidth]{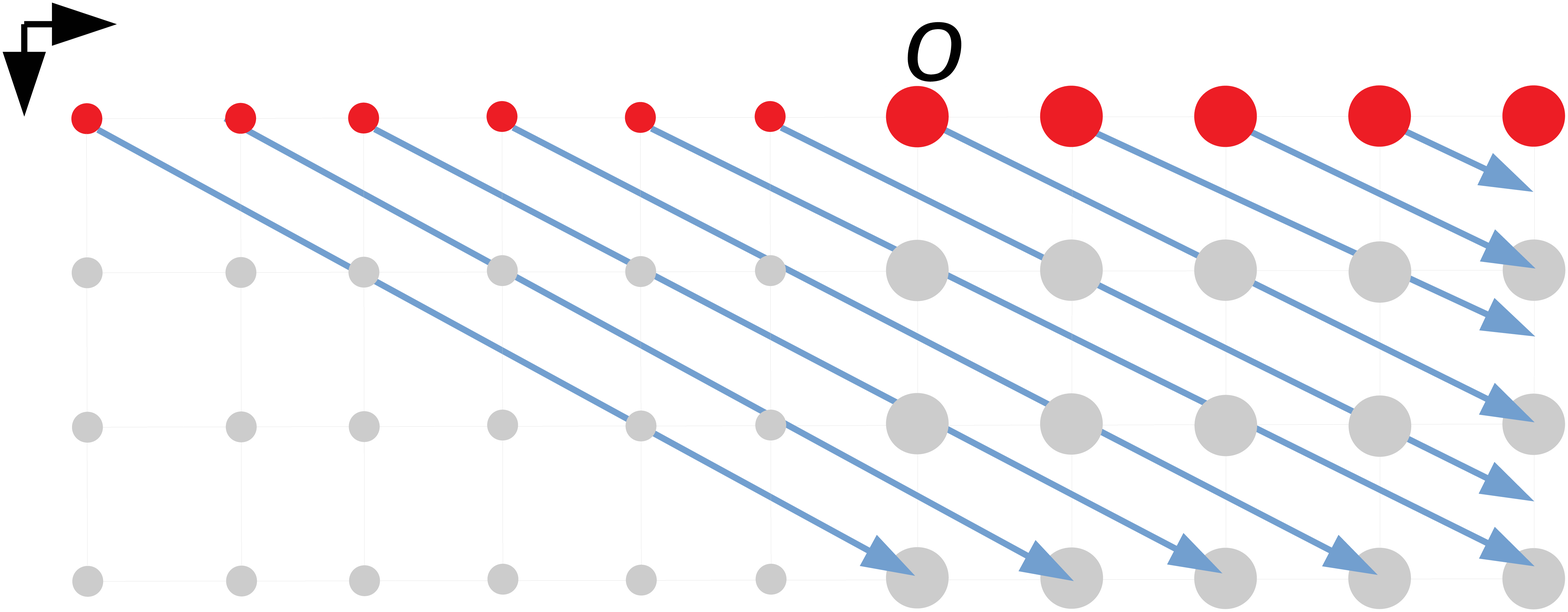}
\caption{}
\end{subfigure}
~
\begin{subfigure}[b]{0.25\textwidth}
\centering
\includegraphics[width=0.75\textwidth]{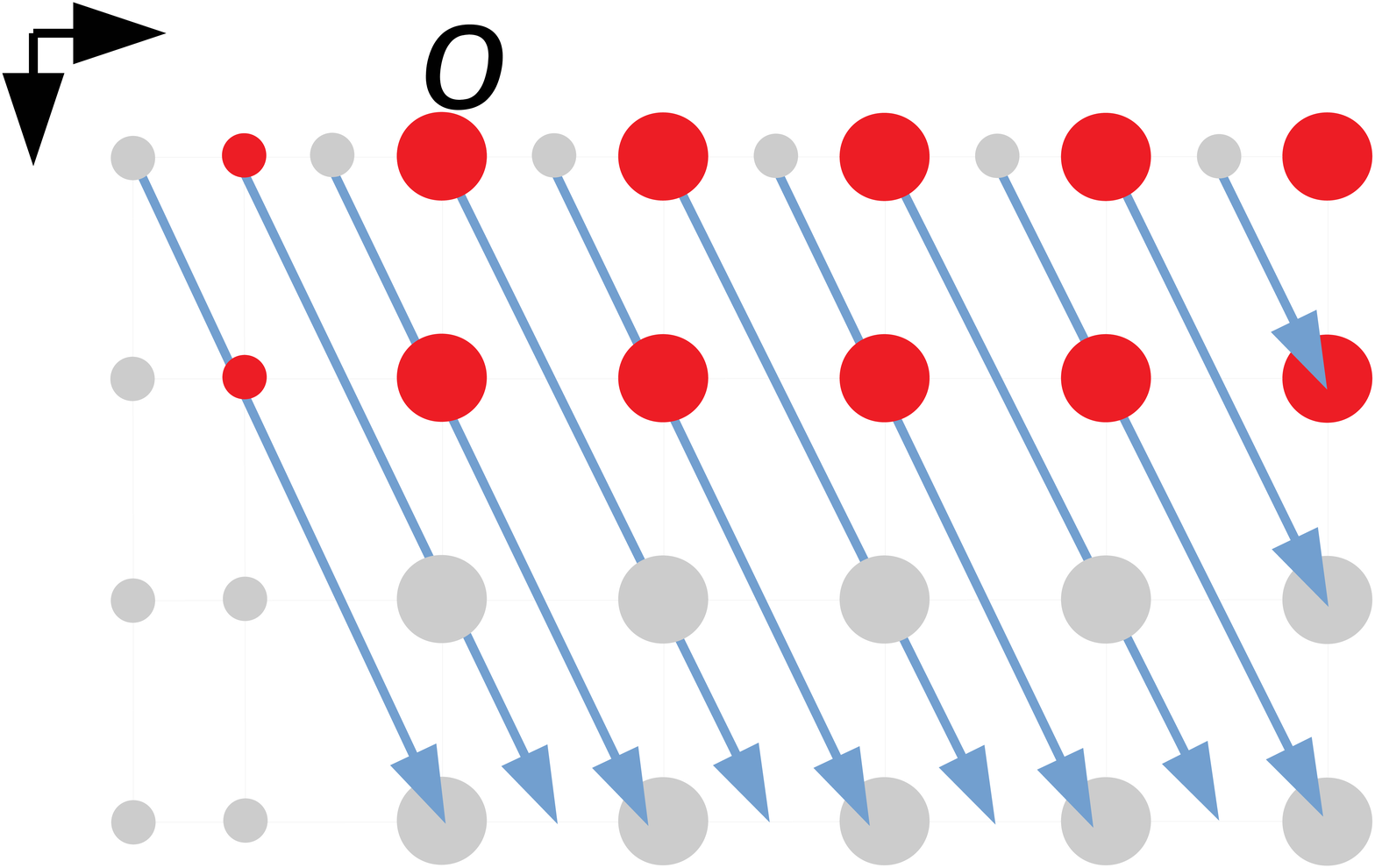}
\caption{}
\end{subfigure}
~
\begin{subfigure}[b]{0.25\textwidth}
\centering
\includegraphics[width=0.75\textwidth]{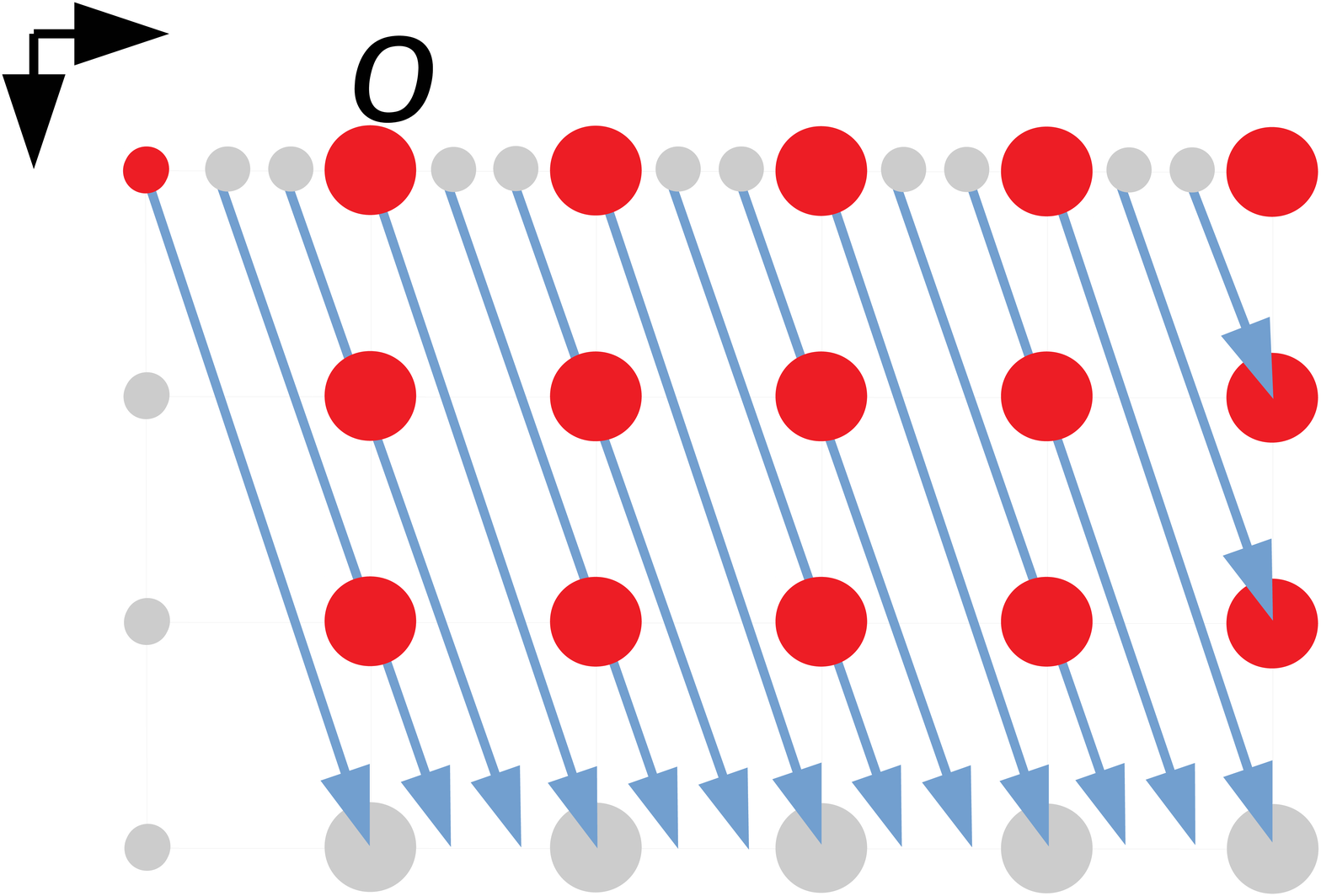}
\caption{}
\end{subfigure}
\caption{Interpolation in asymmetric graph trees in forward passing. (a)
asymmetric wide trees with steps $S=(1, 2)$. (b) asymmetric narrow trees with $S=(2, 1)$.
(c) asymmetric narrow trees with $S=(3, 1)$. Red circles are first nodes of trees;
large circles are within image size; small circles are interpolated; $o$ is axes center.
Coordinates of interpolations in (a) are integral; in (b)-(c) round to the nearest
integers by \eqref{eq:supp-narrow}.}
\label{fig:supp-head}
\end{figure}

%==========================================================================
\section{Differentiability of ISGMR}
Below, we replace $m^{r,k}$ by $m^r$ and $m^{r,k+1}$ by $\hat{m}^r$
for simplicity. This is because from the practical implementation, messages
in direction $r$ should be updated instead of allocating new
memories in each iteration to avoid GPU memory increase. Thus, we only use
two variables $m^r$ and $\hat{m}^r$ for messages before and after an iteration.

%--------------------------------------------------------------------------
\subsection{Explicit Representation of Forward Propagation}
Since message update in ISGMR relies on recursively updated messages $\hat{\mathbf{m}}^r$
in each scanning direction $r$ and messages $\mathbf{m}^r$ from all the other directions
updated in the previous iteration, an explicit ISGMR message update is

\begin{equation}
\label{eq:supp-isgmr-explicit}
\begin{aligned}
\hmesnew{r}{i}{\lambda} &= \min_{\mu \in \mathcal{L}}\, \big(\theta_{i-r}{(\mu)}
+ \theta_{i-r, i}(\mu, \lambda)
+ \hmesnew{r}{i-r}{\mu} 
+ \sum_{d \in \mathcal{R} \setminus \{r, r^-\}} \mesnew{d}{i-r}{\mu} \big), \\
&\quad \quad \forall i \in \mathcal{V}, \forall \lambda \in \mathcal{L},
\forall r \in \mathcal{R}\ .
\end{aligned}
\end{equation}
Applying message reparametrization by

\begin{equation}
\label{eq:supp-isgmr-norm}
\begin{aligned}
\hmesnew{r}{i}{\lambda} = \hmesnew{r}{i}{\lambda} - \min_{k \in \mathcal{L}}
\hmesnew{r}{i}{k}, \quad \forall i \in \mathcal{V}, \forall \lambda \in \mathcal{L},
\forall r \in \mathcal{R}\ .
\end{aligned}
\end{equation}
After updating messages in \textbf{all directions within an iteration},
we assign the updated message $\hat{\mathbf{m}}$ to $\mathbf{m}$ by

\begin{equation}
\label{eq:supp-isgmr-renew}
\begin{aligned}
m_{i}{(\lambda)} = \hat{m}_{i}{(\lambda)}, \quad \forall i \in \mathcal{V},
\forall \lambda \in \mathcal{L}\ .
\end{aligned}
\end{equation}
Eventually, \textbf{after all iterations}, unary potentials and updated messages
from all directions will be aggregated by

\begin{equation}
\label{eq:supp-isgmr-aggregate}
\begin{aligned}
c_{i}{(\lambda)} = \theta_{i}{(\lambda)} + \sum_{d \in \mathcal{R}}
\mesnew{d}{i}{\lambda}, \quad \forall i \in \mathcal{V}, \forall \lambda \in \mathcal{L}\ .
\end{aligned}
\end{equation}

Different from optimization with winner-takes-all for labelling in learning by
$\mathbf{x}_{i} = \text{argmin}_{\lambda \in \mathcal{L}}c_{i}{(\lambda)},
\forall i \in \mathcal{V}$, a regression with disparity confidences calculated
by the final costs is used to fit with the real-valued ground truth
disparities $\mathbf{g}=\{g_i\}, \forall i \in \mathcal{V}$. Generally, the
disparity confidence $f_{i}{(\lambda)}$ with a normalization such as SoftMin() is represented by

\begin{equation}
\label{eq:supp-isgmr-confidence}
\begin{aligned}
f_{i}{(\lambda)} = \text{SoftMin}(c_{i}{(\lambda)}), \quad \forall i \in
\mathcal{V}, \forall \lambda \in \mathcal{L}\ ,
\end{aligned}
\end{equation}
and the regression for real-valued disparity $\mathbf{d}=\{d_i\},
\forall i \in \mathcal{V}$ is

\begin{equation}
\label{eq:supp-isgmr-regression}
\begin{aligned}
d_{i} = \sum_{\lambda \in \mathcal{L}} \lambda f_{i}{(\lambda)},
\forall i \in \mathcal{V}\ .
\end{aligned}
\end{equation}

The loss function $L(\mathbf{d}, \mathbf{g})$ in learning can be standard L1 or smooth L1
loss function.

%--------------------------------------------------------------------------
\subsection{Derivations of Differentiability}

Now we do backpropagation at $k$th iteration for learnable parameters $\{ \theta_i$, $\theta_{i,j} \}$.
With the same notations in Section~\myblue{4} in the main paper, $\{ p^{r}_{k,i}{(\lambda)} \}$
and $\{ q^r_{k,i} \}$ are indices stored in the forward propagation from message minimization and
reparameterization respectively, and $\nabla * = dL / d*$.

\subsubsection{Gradients of unary potentials} \

\noindent \textbf{Proposition:} Gradients of unary potentials $\{\theta_{i}{(\lambda)}\}$ are represented by

\begin{equation}
\label{eq:supp-gradient-unary}
\begin{aligned}
\nabla \theta_{i}{(\lambda)}
&= \nabla c_{i}{(\lambda)} + \sum_{v \in \mathcal{L}} \sum_{r \in \mathcal{R}} \left. \nabla
\hmesnew{r}{i+r}{v} \right \vert_{\lambda = p^r_{k,i+r}{(v)}} \\
&= \nabla c_{i}{(\lambda)} + \sum_{v \in \mathcal{L}} \sum_{r \in \mathcal{R}} \sum_{\mu \in \mathcal{L}}
\Big( \left. \nabla \hmesnew{r}{i+2r}{\mu} \right \vert_{v = p^r_{k,i+2r}{(\mu)}} \\
&\quad \quad + \sum_{d \in \mathcal{R} \setminus \{r, r^{-}\}} \left.
\nabla \mesnew{d}{i+r+d}{\mu} \right \vert_{v = p^d_{k,i+r+d}{(\mu)}} \left. \Big ) \right \vert_{\lambda = p^r_{k,i+r}{(v)}}\ .
\end{aligned}
\end{equation}

\noindent \textit{Derivation:}

The backpropagation from \eqref{eq:supp-isgmr-regression}-\eqref{eq:supp-isgmr-explicit} is

\begin{equation}
\label{eq:supp-isgmr-back-unary}
\begin{aligned}
\nabla \theta_{i}{(\lambda)}
&=\frac{dL}{d\theta_{i}{(\lambda)}} \\
&= \sum_{j \in \mathcal{V}} \sum_{v \in \mathcal{L}} \frac{\partial L}{\partial d_{j}{(v)}}
\frac{\partial d_{j}{(v)}}{\partial f_{j}{(v)}} \frac{\partial f_{j}{(v)}}{\partial c_{j}{(v)}}
\frac{\partial c_{j}{(v)}}{\partial \theta_{i}{(\lambda)}}
& \tcp*[f]{back \eqref{eq:supp-isgmr-regression}-\eqref{eq:supp-isgmr-confidence}} \\
&= \sum_{j \in \mathcal{V}} \sum_{v \in \mathcal{L}} \nabla c_{j}{(v)}
\Big ( \frac{\partial c_{j}{(v)}}{\partial \theta_{j}{(v)}}
\frac{\partial \theta_{j}{(v)}}{\partial \theta_{i}{(\lambda)}} + \sum_{r \in \mathcal{R}}
\frac{\partial c_{j}{(v)}}{\partial m^r_{j}{(v)}} \frac{\partial m^r_{j}{(v)}}{\partial \theta_{i}{(\lambda)}} \Big )
& \tcp*[f]{back \eqref{eq:supp-isgmr-aggregate}} \\
&= \nabla c_{i}{(\lambda)} + \sum_{j \in \mathcal{V}} \sum_{v \in \mathcal{L}}
\sum_{r \in \mathcal{R}} \nabla \mesnew{r}{j}{v} \frac{\partial \mesnew{r}{j}{v}}{\partial \theta_{i}{(\lambda)}}  \\
&= \nabla c_{i}{(\lambda)} + \sum_{j \in \mathcal{V}} \sum_{v \in \mathcal{L}}
\sum_{r \in \mathcal{R}} \nabla \mesnew{r}{j}{v} \frac{\partial \mesnew{r}{j}{v}}{\partial \hmesnew{r}{j}{v}}
\frac{\partial \hmesnew{r}{j}{v}}{\partial \theta_{i}{(\lambda)}} 
& \tcp*[f]{back \eqref{eq:supp-isgmr-renew}} \\
&= \nabla c_{i}{(\lambda)} + \sum_{j \in \mathcal{V}} \sum_{v \in \mathcal{L}}
\sum_{r \in \mathcal{R}} \nabla \hmesnew{r}{j}{v} \frac{\partial \hmesnew{r}{j}{v}}{\partial \theta_{i}{(\lambda)}}\ .
\end{aligned}
\end{equation}

With backpropagation of \eqref{eq:supp-isgmr-norm} using an implicit message reparametrization
with index $v^* = q^{r}_{k,j}$ at $k$th iteration, $\nabla \hmesnew{r}{j}{v}$ in the second term above is updated by

\begin{equation}
\label{eq:supp-isgmr-back-norm}
\begin{aligned}
\nabla \hmesnew{r}{j}{v}
& \gets
  \begin{cases}
    \nabla \hmesnew{r}{j}{v} & \text{ if } v \neq v^*\ , \\
     - \sum_{v^{'}\in \mathcal{L} \setminus v^*} \nabla \hmesnew{r}{j}{v^{'}} & \text{otherwise}\ .
  \end{cases}
\end{aligned}
\end{equation}

\noindent \textit{Derivation of \eqref{eq:supp-isgmr-back-norm}:}

Explicit representation of \eqref{eq:supp-isgmr-norm} is $\tilde{m}^{r}_{i}{(\lambda)} =
\hmesnew{r}{i}{\lambda} - \hmesnew{r}{i}{\lambda^{*}}$, where $\lambda^{*} = q^r_{k,i}$, then we have

\begin{equation}
\label{eq:supp-isgmr-back-norm-explicit}
\centering
\begin{aligned}
\nabla \hmesnew{r}{i}{\lambda}
&= \frac{\partial L}{\partial \hmesnew{r}{i}{\lambda}} \\
&= \sum_{i^{'} \in \mathcal{V}} \sum_{\lambda^{'} \in \mathcal{L}}
\frac{\partial L}{\partial \tilde{m}^r_{i^{'}}{(\lambda^{'})}}
\frac{\partial \tilde{m}^r_{i^{'}}{(\lambda^{'})}}{\partial \hmesnew{r}{i}{\lambda}}  \\
&= \sum_{i^{'} \in \mathcal{V}} \sum_{\lambda^{'} \in \mathcal{L}}
\frac{\partial L}{\partial \tilde{m}^r_{i^{'}}{(\lambda^{'})}}
\Big( \frac{\partial \tilde{m}^r_{i^{'}}{(\lambda^{'})}}{\partial \hmesnew{r}{i^{'}}{\lambda^{'}}}
\frac{\partial \hmesnew{r}{i^{'}}{\lambda^{'}}}{\partial \hmesnew{r}{i}{\lambda}}
+ \frac{\partial \tilde{m}^r_{i^{'}}{(\lambda^{'})}}{\partial \hmesnew{r}{i^{'}}{\lambda^{*}}}
\frac{\partial \hmesnew{r}{i^{'}}{\lambda^{*}}}{\partial \hmesnew{r}{i}{\lambda}} \Big) \\
&= \frac{\partial L}{\partial \tilde{m}^r_{i}{(\lambda)}} - \sum_{\lambda^{'} \in \mathcal{L}} \left.
\frac{\partial L}{\partial \tilde{m}_{i}{(\lambda^{'})}} \right \vert _{\lambda=\lambda^{*}} \\
&= \begin{cases}
     \nabla \tilde{m}^r_{i}{(\lambda)} & \text{ if } \lambda \neq \lambda^* \ ,\\
      - \sum_{\lambda^{'}\in \mathcal{L} \setminus \lambda^*} \nabla \tilde{m}^r_{i}{(\lambda^{'})} & \text{otherwise}\ .
   \end{cases}
\end{aligned}
\end{equation}
Back to the implicit message reparametrization with $\nabla \tilde{m}^r$ replaced
by $\nabla \hat{m}^r$, we have

\begin{equation}
\begin{aligned}
\nabla \hmesnew{r}{i}{\lambda}
&= \begin{cases}
     \nabla \hmesnew{r}{i}{\lambda} & \text{ if } \lambda \neq \lambda^* \ , \\
      - \sum_{\lambda^{'}\in \mathcal{L} \setminus \lambda^*} \nabla \hmesnew{r}{i}{\lambda^{'}} & \text{otherwise}\ .
   \end{cases}
\end{aligned}
\end{equation}

\noindent \textit{End of the derivation of \eqref{eq:supp-isgmr-back-norm}}. 

Next, we continue the backpropagation through \eqref{eq:supp-isgmr-explicit} for unary potentials as

\begin{equation}
\label{eq:supp-isgmr-back-message-unary}
\begin{aligned}
\nabla \theta_{i}{(\lambda)}
&= \nabla c_{i}{(\lambda)} + \sum_{j \in \mathcal{V}} \sum_{v \in \mathcal{L}}
\sum_{r \in \mathcal{R}} \nabla \hmesnew{r}{j}{v} \frac{\partial \hmesnew{r}{j}{v}}{\partial \theta_{i}{(\lambda)}} 
& \tcp*[f]{from \eqref{eq:supp-isgmr-back-unary}} \\
&= \nabla c_{i}{(\lambda)} + \sum_{v \in \mathcal{L}} \sum_{r \in \mathcal{R}}
\nabla \hmesnew{r}{i+r}{v} \frac{\partial \hmesnew{r}{i+r}{v}}{\partial \theta_{i}{(\lambda)}} 
& \tcp*[f]{back \eqref{eq:supp-isgmr-explicit} without recursion} \\
&= \nabla c_{i}{(\lambda)} + \sum_{v \in \mathcal{L}} \sum_{r \in \mathcal{R}} \left.
\nabla \hmesnew{r}{i+r}{v} \right \vert_{\lambda = p^r_{k,i+r}{(v)}}. \ & \tcp*[f]{satisfy argmin()
rule in \eqref{eq:supp-isgmr-explicit}}
\end{aligned}
\end{equation}

Derivation of $\nabla c_{i}{(\lambda)}$ by backpropagation from the loss function,
disparity regression, and SoftMin(), can be obtained by PyTorch autograd directly.
For the readability of derivations by avoiding using
$\{m^r_{i+r}(m^{r}_i(\theta_{i-r}{(\lambda)})), m^r_{i+2r}(m^r_{i+r}(m^{r}_i(\theta_{i-r}{(\lambda)}))), ...\}$, 
we do not write the recursion of gradients in the derivations. Below, we derive
$\nabla \hmesnew{r}{i+r}{v}$ in the backpropagation.

\subsubsection{Gradients of Messages} \

For notation readability, we first derive message gradient $\nabla \hmesnew{r}{i}{\lambda}$
instead of $\nabla \hmesnew{r}{i+r}{v}$.

\noindent \textbf{Proposition:} Gradients of messages $\{ \hmesnew{r}{i}{\lambda} \}$ are
represented by

\begin{equation}
\label{eq:supp-gradient-message}
\begin{aligned}
\nabla \hmesnew{r}{i}{\lambda}
= \sum_{v \in \mathcal{L}} \Big( \left. \nabla \hmesnew{r}{i+r}{v} \right \vert_{\lambda
= p^r_{k,i+r}{(v)}} + \sum_{d \in \mathcal{R} \setminus \{r, r^{-}\}} \left. \nabla m^d_{i+d}{(v)} \right \vert_{\lambda = p^d_{k,i+d}{(v)}} \Big )\ .
\end{aligned}
\end{equation}

\noindent \textit{Derivation:}

\begin{equation}
\label{eq:supp-isgmr-back-message}
\begin{aligned}
\nabla \hmesnew{r}{i}{\lambda}
&= \frac{dL}{d\hmesnew{r}{i}{\lambda}} \\
&= \sum_{j \in \mathcal{V}} \sum_{v \in \mathcal{L}} \nabla c_{j}{(v)}
\frac{\partial c_{j}{(v)}}{\partial \hmesnew{r}{i}{\lambda}} 
& \tcp*[f]{back \eqref{eq:supp-isgmr-regression}-\eqref{eq:supp-isgmr-confidence}} \\
&= \sum_{j \in \mathcal{V}} \sum_{v \in \mathcal{L}}  \nabla c_{j}{(v)}
\sum_{d \in \mathcal{R}} \frac{\partial c_{j}{(v)}}{\partial m^d_{j}{(v)}}
\frac{\partial m^d_{j}{(v)}}{\partial \hmesnew{r}{i}{\lambda}}
& \tcp*[f]{back \eqref{eq:supp-isgmr-aggregate}} \\
&= \sum_{j \in \mathcal{V}} \sum_{v \in \mathcal{L}} \nabla c_{j}{(v)}
\sum_{d \in \mathcal{R}} \frac{\partial c_{j}{(v)}}{\partial m^d_{j}{(v)}}
\frac{\partial m^d_{j}{(v)}}{\partial \hmesnew{d}{j}{v}} \frac{\partial \hmesnew{d}{j}{v}}{\partial \hmesnew{r}{i}{\lambda}} 
& \tcp*[f]{back \eqref{eq:supp-isgmr-renew}} \\
&= \sum_{j \in \mathcal{V}} \sum_{v \in \mathcal{L}} \sum_{d \in \mathcal{R}}
\nabla \hmesnew{d}{j}{v} \frac{\partial \hmesnew{d}{j}{v}}{\partial \hmesnew{r}{i}{\lambda}}\ ,
\end{aligned}
\end{equation}
then we update $\nabla \hmesnew{d}{j}{v}$ by \eqref{eq:supp-isgmr-back-norm} and continue as follows,

\begin{equation}
\label{eq:supp-isgmr-back-message-2}
\begin{aligned}
\nabla \hmesnew{r}{i}{\lambda}
&= \sum_{j \in \mathcal{V}} \sum_{v \in \mathcal{L}} \sum_{d \in \mathcal{R}}
\nabla \hmesnew{d}{j}{v} \frac{\partial \hmesnew{d}{j}{v}}{\partial \hmesnew{r}{i}{\lambda}} 
& \tcp*[f]{from \eqref{eq:supp-isgmr-back-message}} \\
&= \sum_{j \in \mathcal{V}} \sum_{v \in \mathcal{L}} \sum_{d \in \mathcal{R}}
\nabla \hmesnew{d}{j}{v} \Big ( \sum_{\lambda^{'} \in \mathcal{L}}
\frac{\partial \hmesnew{d}{j}{v}}{\partial \hmesnew{d}{j-d}{\lambda^{'}}}
\frac{\partial \hmesnew{d}{j-d}{\lambda^{'}}}{\partial \hmesnew{r}{i}{\lambda}} \\
& \quad + \sum_{d^{'} \in \mathcal{R} \setminus \{d, d^{-}\}}
\sum_{\lambda^{'} \in \mathcal{L}} \frac{\partial \hmesnew{d}{j}{v}}{
\partial m^{d^{'}}_{j-d}{(\lambda^{'})}} \frac{\partial m^{d^{'}}_{j-d}{(\lambda^{'})}}{\partial \hmesnew{r}{i}{\lambda}} \Big )
& \tcp*[f]{back \eqref{eq:supp-isgmr-explicit}} \\
&= \sum_{v \in \mathcal{L}} \Big( \left. \nabla \hmesnew{r}{i+r}{v}
\right \vert_{\lambda = p^{r}_{k,i+r}{(v)}} \\
&\quad + \sum_{d \in \mathcal{R}} \sum_{d^{'} \in \mathcal{R} \setminus \{d, d^{-}\}}
\sum_{\lambda^{'} \in \mathcal{L}} \nabla \hmesnew{d}{j}{v}
\frac{\partial \hmesnew{d}{j}{v}}{\partial m^{d^{'}}_{j-d}{(\lambda^{'})}}
\frac{\partial m^{d^{'}}_{j-d}{(\lambda^{'})}}{\partial \hmesnew{r}{i}{\lambda}} \Big ) \ .
\end{aligned}
\end{equation}
Since $m^{d^{'}}_{j-d}{(\lambda^{'})}$ is differentiable by $\hmesnew{r}{i}{\lambda}$
due to \eqref{eq:supp-isgmr-renew} and, for ISGMR, message gradients in directions
except the current direction $r$ come from the next iteration (since in the forward
propagation these messages come from the previous iteration), we have

\begin{equation}
\label{eq:supp-isgmr-back-message-3}
\begin{aligned}
\nabla \hmesnew{r}{i}{\lambda}
&= \sum_{v \in \mathcal{L}} \Big( \left. \nabla \hmesnew{r}{i+r}{v}
\right \vert_{\lambda = p^{r}_{k,i+r}{(v)}} \\
&\quad + \sum_{d \in \mathcal{R}} \sum_{d^{'} \in \mathcal{R} \setminus \{d, d^{-}\}}
\sum_{\lambda^{'} \in \mathcal{L}} \nabla \hmesnew{d}{j}{v}
\frac{\partial \hmesnew{d}{j}{v}}{\partial m^{d^{'}}_{j-d}{(\lambda^{'})}}
\frac{\partial m^{d^{'}}_{j-d}{(\lambda^{'})}}{\partial \hat{m}^r_{i}{(\lambda)}} \Big ) 
&\tcp*[f]{from \eqref{eq:supp-isgmr-back-message-2}} \\
&= \sum_{v \in \mathcal{L}} \Big( \left. \nabla \hmesnew{r}{i+r}{v}
\right \vert_{\lambda = p^{r}_{k,i+r}{(v)}} \\
&\quad + \sum_{d \in \mathcal{R}} \left. \nabla \hmesnew{d}{i+d}{v}
\frac{\partial \hmesnew{d}{i+d}{v}}{\partial m^{r}_{i}{(\lambda)}}
\frac{\partial m^{r}_{i}{(\lambda)}}{\partial \hmesnew{r}{i}{\lambda}}
\right \vert_{r \not \in \{d, d^{-}\}} \Big ) 
& \tcp*[f]{due to \eqref{eq:supp-isgmr-renew}} \\
&= \sum_{v \in \mathcal{L}} \Big( \left. \nabla \hmesnew{r}{i+r}{v}
\right \vert_{\lambda = p^r_{k,i+r}{(v)}} \\
&\quad + \sum_{d \in \mathcal{R} \setminus \{r, r^{-}\}} \left.
\nabla m^d_{i+d}{(v)} \right \vert_{\lambda = p^d_{k,i+d}{(v)}} \Big )\ .
\end{aligned}
\end{equation}
Here, updating the message gradient at node $i$ depends on its next node
$i + r$ along the scanning direction $r$; this scanning direction is opposite
to the forward scanning direction, and thus, it depends on node $i + r$ instead
of $i - r$. Gradient of message $m^r_{i}{(\lambda)}$ can be derived in the same way.

Now one can derive $\nabla \hmesnew{r}{i+r}{v}$ in the same manner of
$\nabla \hmesnew{r}{i}{\lambda}$ and apply it to \eqref{eq:supp-isgmr-back-message-unary}
to obtain \eqref{eq:supp-gradient-unary}.

\subsubsection{Gradient of Pairwise Potentials} \

\noindent \textbf{Proposition:} Gradients of pairwise potentials
$\{ \theta_{i-r,i}(\mu,\lambda) \}$ are represented by

\begin{equation}
\label{eq:supp-gradient-pairwise}
\begin{aligned}
\nabla \theta_{i-r,i}(\mu,\lambda) = \left. \nabla \hmesnew{r}{i}{\lambda}
\right \vert_{\mu=p^{r}_{k,i}{(\lambda)}}, \quad \forall i \in \mathcal{V},
\forall r \in \mathcal{R}, \forall \lambda,\mu \in \mathcal{L}\ .
\end{aligned}
\end{equation}

\noindent \textit{Derivation:}

\begin{equation}
\label{eq:supp-isgmr-back-pairwise}
\begin{aligned}
\nabla \theta_{i-r,i}(\mu,\lambda)
&= \frac{dL}{d\theta_{i-r, i}(\mu,\lambda)} \\
&= \sum_{j \in \mathcal{V}} \sum_{v \in \mathcal{L}} \nabla c_{j}{(v)}
\frac{\partial c_{j}{(v)}}{\partial \theta_{i-r, i}(\mu,\lambda)} 
& \tcp*[f]{back \eqref{eq:supp-isgmr-regression}-\eqref{eq:supp-isgmr-confidence}} \\
&= \sum_{j \in \mathcal{V}} \sum_{v \in \mathcal{L}} \nabla c_{j}{(v)}
\sum_{d \in \mathcal{R}} \frac{\partial c_{j}{(v)}}{\partial m^d_{j}{(v)}} 
\frac{\partial m^d_{j}{(v)}}{\partial \theta_{i-r, i}(\mu,\lambda)} 
& \tcp*[f]{back \eqref{eq:supp-isgmr-aggregate}} \\
&= \sum_{j \in \mathcal{V}} \sum_{v \in \mathcal{L}} \sum_{d \in \mathcal{R}}
\nabla c_{j}{(v)} \frac{\partial c_{j}{(v)}}{\partial m^d_{j}{(v)}}
\frac{\partial m^d_{j}{(v)}}{\partial \hmesnew{d}{j}{v}} \frac{\partial \hmesnew{d}{j}{v}}{\partial \theta_{i-r, i}(\mu,\lambda)} 
& \tcp*[f]{back \eqref{eq:supp-isgmr-renew}} \\
&= \sum_{j \in \mathcal{V}} \sum_{v \in \mathcal{L}} \sum_{d \in \mathcal{R}}
\nabla \hmesnew{d}{j}{v} \frac{\partial \hmesnew{d}{j}{v}}{\partial \theta_{i-r, i}(\mu,\lambda)}\ .
\end{aligned}
\end{equation}

Now we update $\nabla \hmesnew{d}{j}{v}$ by \eqref{eq:supp-isgmr-back-norm}. Then

\begin{equation}
\label{eq:supp-isgmr-back-pairwise-2}
\begin{aligned}
\nabla \theta_{i-r,i}(\mu,\lambda)
&= \sum_{j \in \mathcal{V}} \sum_{v \in \mathcal{L}} \sum_{d \in \mathcal{R}}
\nabla \hmesnew{d}{j}{v} \frac{\partial \hmesnew{d}{j}{v}}{\partial \theta_{i-r, i}(\mu,\lambda)} 
& \tcp*[f]{from \eqref{eq:supp-isgmr-back-pairwise}} \\
&= \left. \nabla \hmesnew{r}{i}{\lambda} \right \vert_{\mu=p^{r}_{k,i}{(\lambda)}} \ .
& \tcp*[f]{back \eqref{eq:supp-isgmr-explicit} without recursion}
\end{aligned}
\end{equation}

One can note that the memory requirement of $\{\theta_{i-r,i}(\mu, \lambda)\}$ is
$4 \sum_{r\in\mathcal{R}} |\mathcal{E}^r| |\mathcal{L}| |\mathcal{L}|$ bytes
using single-precision floating-point values. This will be high when the number
of disparities $|\mathcal{L}|$ is large. In practical, since the pairwise potentials
can be decomposed by $\theta_{i,j}(\lambda, \mu) = \theta_{i,j} V(\lambda, \mu),
\forall (i,j) \in \mathcal{E}, \forall \lambda, \mu \in \mathcal{L}$ with edge weights
$\theta_{i,j}$ and a pairwise function $V(\cdot, \cdot)$, it takes up
$4 ( \sum_{r\in\mathcal{R}} |\mathcal{E}^r| + |\mathcal{L}| |\mathcal{L}|)$ bytes
in total, which is much less than $4 \sum_{r\in\mathcal{R}} |\mathcal{E}^r| |\mathcal{L}| |\mathcal{L}|$ above.
Therefore, we additionally provide the gradient derivations of these two terms,
edge weights and pairwise functions, for practical implementations of the backpropagation.

\subsubsection{Gradient of Edge Weights} \

\noindent \textbf{Proposition:} Gradients of edge weights $\{ \theta_{i-r,i} \}$ are represented by

\begin{equation}
\label{eq:supp-gradient-edge}
\begin{aligned}
\nabla \theta_{i-r,i} = \sum_{v \in \mathcal{L}} \nabla \hmesnew{r}{i}{v}
V(p^r_{k,i}{(v)},v), \quad \forall i \in \mathcal{V}, \forall r \in \mathcal{R}\ .
\end{aligned}
\end{equation}

\noindent \textit{Derivation:}

\begin{equation}
\label{eq:supp-isgmr-back-edge}
\begin{aligned}
\nabla \theta_{i-r,i}
&= \frac{dL}{d\theta_{i-r, i}} \\
&= \sum_{j \in \mathcal{V}} \sum_{v \in \mathcal{L}} \nabla c_{j}{(v)}
\frac{\partial c_{j}{(v)}}{\partial \theta_{i-r, i}} 
& \tcp*[f]{back \eqref{eq:supp-isgmr-regression}-\eqref{eq:supp-isgmr-confidence}} \\
&= \sum_{j \in \mathcal{V}} \sum_{v \in \mathcal{L}} \nabla c_{j}{(v)}
\sum_{d \in \mathcal{R}} \frac{\partial c_{j}{(v)}}{\partial m^d_{j}{(v)}}
\frac{\partial m^d_{j}{(v)}}{\partial \theta_{i-r, i}} 
& \tcp*[f]{back \eqref{eq:supp-isgmr-aggregate}} \\
&= \sum_{j \in \mathcal{V}} \sum_{v \in \mathcal{L}} \sum_{d \in \mathcal{R}}
\nabla c_{j}{(v)} \frac{\partial c_{j}{(v)}}{\partial m^d_{j}{(v)}}
\frac{\partial m^d_{j}{(v)}}{\partial \hmesnew{d}{j}{v}} \frac{\partial \hmesnew{d}{j}{v}}{\partial \theta_{i-r, i}}
& \tcp*[f]{back \eqref{eq:supp-isgmr-renew}} \\
&= \sum_{j \in \mathcal{V}} \sum_{v \in \mathcal{L}} \sum_{d \in \mathcal{R}}
\nabla \hmesnew{d}{j}{v} \frac{\partial \hmesnew{d}{j}{v}}{\partial \theta_{i-r, i}}\ .
\end{aligned}
\end{equation}
Again, before updating gradients of edge weights by \eqref{eq:supp-isgmr-explicit},
$\nabla \hmesnew{d}{j}{v}$ is updated by \eqref{eq:supp-isgmr-back-norm}. Then

\begin{equation}
\label{eq:supp-isgmr-back-edge-2}
\begin{aligned}
\nabla \theta_{i-r,i}
&= \sum_{j \in \mathcal{V}} \sum_{v \in \mathcal{L}} \sum_{d \in \mathcal{R}}
\nabla \hmesnew{d}{j}{v} \frac{\partial \hmesnew{d}{j}{v}}{\partial \theta_{i-r, i}}
& \tcp*[f]{from \eqref{eq:supp-isgmr-back-edge}} \\
&= \sum_{j \in \mathcal{V}} \sum_{v \in \mathcal{L}} \sum_{d \in \mathcal{R}}
\nabla \hmesnew{d}{j}{v} \frac{\partial \hmesnew{d}{j}{v}}{\partial \theta_{j-d, j}} V(p^d_{k,j}{(v)},v)
\frac{\partial \theta_{j-d, j}}{\partial \theta_{i-r, i}}
& \tcp*[f]{back \eqref{eq:supp-isgmr-explicit}, no recursion} \\
&= \sum_{v \in \mathcal{L}} \nabla \hmesnew{r}{i}{v} V(p^r_{k,i}{(v)},v)\ .
\end{aligned}
\end{equation}

In the case that when edge weights are undirected, \ie, $\theta_{i,j} = \theta_{j,i}$,
the derivations above still hold, and if $\theta_{i,j} = \theta_{j,i}$ are stored in
the same tensor, $\nabla \theta_{i,j}$ will be accumulated by adding $\nabla \theta_{j,i}$
for storing the gradient of this edge weight. This is also applied to the gradient of
pairwise potentials in \eqref{eq:supp-gradient-pairwise} above.

\subsubsection{Gradients of Pairwise Functions} \

\noindent \textbf{Proposition:} Gradients of a pairwise function $V(\cdot, \cdot)$ are

\begin{equation}
\label{eq:supp-gradient-smoothness}
\begin{aligned}
\nabla V(\lambda, \mu) = \sum_{j \in \mathcal{V}} \sum_{r \in \mathcal{R}} \left.
\theta_{j-r, j} \nabla \hmesnew{r}{j}{\mu} \right \vert_{\lambda = p^r_{k,j}{(\mu)}}, \quad \forall \lambda, \mu \in \mathcal{L}\ .
\end{aligned}
\end{equation}

\noindent \textit{Derivation:}

\begin{equation}
\label{eq:supp-isgmr-back-smoothness}
\begin{aligned}
\nabla V(\lambda, \mu)
&= \frac{dL}{dV(\lambda, \mu)} \\
&= \sum_{j \in \mathcal{V}} \sum_{v \in \mathcal{L}} \nabla c_{j}{(v)}
\frac{\partial c_{j}{(v)}}{\partial V(\lambda, \mu)}
& \tcp*[f]{back \eqref{eq:supp-isgmr-regression}-\eqref{eq:supp-isgmr-confidence}} \\
&= \sum_{j \in \mathcal{V}} \sum_{v \in \mathcal{L}} \nabla c_{j}{(v)}
\sum_{r \in \mathcal{R}} \frac{\partial c_{j}{(v)}}{\partial m^r_{j}{(v)}}
\frac{\partial m^r_{j}{(v)}}{\partial V(\lambda, \mu)}
& \tcp*[f]{back \eqref{eq:supp-isgmr-aggregate}} \\
&= \sum_{j \in \mathcal{V}} \sum_{v \in \mathcal{L}} \sum_{r \in \mathcal{R}}
\nabla c_{j}{(v)} \frac{\partial c_{j}{(v)}}{\partial m^r_{j}{(v)}}
\frac{\partial m^r_{j}{(v)}}{\partial \hmesnew{r}{j}{v}} \frac{\partial \hmesnew{r}{j}{v}}{\partial V(\lambda, \mu)}
& \tcp*[f]{back \eqref{eq:supp-isgmr-renew}} \\
&= \sum_{j \in \mathcal{V}} \sum_{v \in \mathcal{L}} \sum_{r \in \mathcal{R}}
\nabla \hmesnew{r}{j}{v} \frac{\partial \hmesnew{r}{j}{v}}{\partial V(\lambda, \mu)}\ .
\end{aligned}
\end{equation}

$\nabla \hmesnew{r}{j}{v}$ is updated by \eqref{eq:supp-isgmr-back-norm}. Then

\begin{equation}
\label{eq:supp-isgmr-back-smoothness-2}
\begin{aligned}
\nabla V(\lambda, \mu)
&= \sum_{j \in \mathcal{V}} \sum_{v \in \mathcal{L}} \sum_{r \in \mathcal{R}}
\nabla \hmesnew{r}{j}{v} \frac{\partial \hmesnew{r}{j}{v}}{\partial V(\lambda, \mu)}
& \tcp*[f]{from \eqref{eq:supp-isgmr-back-smoothness}} \\
&= \sum_{j \in \mathcal{V}} \sum_{v \in \mathcal{L}} \sum_{r \in \mathcal{R}}
\nabla \hmesnew{r}{j}{v} \sum_{\lambda^{'} \in \mathcal{L}}
\frac{\partial \hmesnew{r}{j}{v}}{\partial V(\lambda^{'}, v)} \frac{\partial V(\lambda^{'}, v)}{\partial V(\lambda, \mu)}
& \tcp*[f]{from \eqref{eq:supp-isgmr-explicit}} \\
&= \sum_{j \in \mathcal{V}} \sum_{r \in \mathcal{R}} \left. \theta_{j-r, j}
\nabla \hmesnew{r}{j}{\mu} \right \vert_{\lambda = p^r_{k,j}{(\mu)}}\ .
\end{aligned}
\end{equation}

%--------------------------------------------------------------------------
\subsection{Characteristics of Backpropagation}
\textbf{1. Accumulation.}
Since a message update usually has several components, its gradient is therefore
accumulated when backpropagating through every component.
For instance, in \eqref{eq:supp-isgmr-back-message-unary}, the gradient of unary
potential $\nabla \theta_{i}{(\lambda)}$ has $\nabla c_{i}{(\lambda)}$ and $\nabla \hmesnew{r}{i+r}{v}, \forall r \in \mathcal{R}$
and $\forall v$ satisfying $\lambda = p^r_{k, i+r}{(v)}$ at $k$th iteration.
It is calculated recursively but not at once due to multiple nodes on a tree,
multiple directions, and multiple iterations. In \eqref{eq:supp-isgmr-back-message-3},
the message gradient of a node relies on the gradient of all nodes after it in the
forward propagation since this message will be used to all the message updates after this node.

\noindent \textbf{2. Zero Out Gradients.}
Message gradients are not accumulated throughout the backpropagation but should be
zeroed out in some cases. In more details, in the forward propagation, the repeated
usage of $\mathbf{m}^\mathbf{r}$ and $\hat{\mathbf{m}}^\mathbf{r}$ is for all iterations
but the messages are, in fact, new variables whenever they are updated. Since the
gradient of a new message must be initialized to 0, zeroing out the gradients of
the new messages is important. Specifically, in ISGMR that within an iteration
$\mathbf{m^r} \gets \hat{\mathbf{m}}^{\mathbf{r}}$ is executed only when message
updates in all directions are done. Thus, $\nabla \mathbf{m^r}$ must be zeroed
out after $\nabla \hat{\mathbf{m}}^\mathbf{r} \gets \nabla \mathbf{m^r}$. Similarly,
after using $\nabla \hat{\mathbf{m}}^\mathbf{r}$ to update the gradients of learnable
parameters and messages, $\nabla \hat{\mathbf{m}}^\mathbf{r} \gets 0, \forall r \in \mathcal{R}$.

%--------------------------------------------------------------------------
\subsection{PyTorch GPU version \textit{vs.} our CUDA version}
For the compared PyTorch GPU version, we highly paralleled individual trees
in each direction while sequential message updates in each tree (equally
scanline) are iterative.
As Pytorch auto-grad is not customized for our min-sum message passing
algorithms, these iterative message updates require to allocate new GPU
memory for each updated message, which makes it very inefficient and
memory-consuming.
Its backpropagation is slower since extra memory is needed to unroll the
forward message passing to compute gradients of messages and all
intermediate variables that require gradients.

In contrast, our implementation is specific to the min-sum message passing.
This min-sum form greatly accelerates our backpropagation by updating
gradients only related to the indices which are stored in pre-allocated
GPU memory during forward pass (line 10 in Alg. 1).
For example, from node $i$ to $i + r$ in Fig. 2(a), forward pass needs messages
over 9 edges (grey lines); but only one (1 of 3 blue lines) from $i + r$ to
$i$ requires gradient updates in the backpropagation.
This makes our CUDA implementation much faster than the PyTorch GPU
version, especially the backpropagation with at least $700\times$ speed-up.

%==========================================================================
\section{Computational Complexity of Min-Sum \& Sum-Product TRW}
Given a graph with parameters $\{\theta_{i}, \theta_{i,j}\}$, maximum iteration $K$,
set of edges $\{\mathcal{E}^r\}$, disparities $\mathcal{L}$, directions $\mathcal{R}$,
computational complexities of min-sum and sum-product TRW are shown below.
For the efficient implementation, let $\theta_{i,j}(\lambda, \mu) = \theta_{i,j} V(\lambda, \mu)$.

%--------------------------------------------------------------------------
\subsection{Computational Complexity of Min-Sum TRW}
Representation of a message update in min-sum TRW is

\begin{equation}
\label{eq:supp-cc-min-sum}
m^r_{i}{(\lambda)} = \min_{\mu \in \mathcal{L}} \Big( \rho_{i-r, i}
\big(\theta_{i-r}{(\mu)} + \sum_{d \in \mathcal{R}} m^d_{i-r}{(\mu)} \big) - m^{r^-}_{i-r}{(\mu)}
+ \theta_{i-r,i} V(\mu, \lambda) \Big) \ .
\end{equation}

In our case where the maximum disparity is less than 256, memory for the backpropagatio
of the min-sum TRW above is only for indices $\mu^* = p^r_{k,i}{(\lambda)} \in \mathcal{L}$
from message minimization with $K\sum_{r\in\mathcal{R}}|\mathcal{E}^r| |\mathcal{L}|$
bytes 8-bit unsigned integer values, as well as for indices from message reparametrization
with $K\sum_{r\in\mathcal{R}}|\mathcal{E}^r|$ bytes. In total, the min-sum TRW
needs $K\sum_{r\in\mathcal{R}}|\mathcal{E}^r| \left( |\mathcal{L}| + 1 \right)$ bytes for the backpropagation.

%--------------------------------------------------------------------------
\subsection{Computational Complexity of Sum-Product TRW}
Representation of a message update in sum-product TRW is
\begin{equation}
\label{eq:supp-cc-sum-product}
\begin{aligned}
\exp(-m^r_{i}{(\lambda)})
&= \sum_{\mu \in \mathcal{L}} \exp \Big( -\rho_{i-r, i} \big( \theta_{i-r}{(\mu)}
+ \sum_{d \in \mathcal{R}} m^d_{i-r}{(\mu)} \big) + m^{r^-}_{i-r}{(\mu)} \\
&\quad - \theta_{i-r, i} V(\mu, \lambda) \Big) \\
&= \sum_{\mu \in \mathcal{L}} \Big( \exp \big(-\rho_{i-r, i} \theta_{i-r}{(\mu)} \big)
\prod_{d \in \mathcal{R}} \exp \big(- \rho_{i-r, i} m^d_{i-r}{(\mu)} \big) \\
&\quad \exp( m^{r^-}_{i-r}{(\mu)}) \exp \big(- \theta_{i-r, i} V(\mu, \lambda) \big) \Big)\ .
\end{aligned}
\end{equation}
Usually, it can be represented as
\begin{equation}
\label{eq:supp-cc-sum-product-2}
\begin{aligned}
\tilde{m}^r_{i}{(\lambda)}
&= \sum_{\mu \in \mathcal{L}} \Big( \exp^{-\rho_{i-r, i} \underline{\theta_{i-r}{(\mu)}}_{1}}
\prod_{d \in \mathcal{R}} \big( \underline{\tilde{m}^d_{i-r}{(\mu)}}_{2} \big)^{\rho_{i-r,i}}
\frac{1}{ \underline{\tilde{m}^{r^-}_{i-r}{(\mu)}}}_{3} \exp^{- \underline{\theta_{i-r,i}}_{4}
\underline{V(\mu, \lambda)}_{5}} \Big) \ .
\end{aligned}
\end{equation}

\noindent \textbf{Problem 1: Numerical Overflow:}
For single-precision floating-point data, a valid numerical range of $x$ in $\exp(x)$ is
less than around 88.7229; otherwise, it will be infinite.
Therefore, for the exponential index in \eqref{eq:supp-cc-sum-product}, a numerical overflow
will happen quite easily. One solution is to reparametrize these messages to a small range,
such as $[0, 1]$, in the same manner as SoftMax(), which requires logarithm to find the
maximum index, followed by exponential operations.

\noindent \textbf{Problem 2: Low efficiency OR high memory requirement in backpropagation:}
In the backpropagation, due to the factorization in \eqref{eq:supp-cc-sum-product-2}, it
needs to rerun the forward propagation to calculate intermediate values OR store all these
values in the forward propagation. However, the former makes the backpropagation at least
as slow as the forward propagation while the later requires a large memory, \\
$K\sum_{r\in\mathcal{R}}|\mathcal{E}^r| |\mathcal{L}| \left( 8|\mathcal{L}|
+ 4 |\mathcal{R}| |\mathcal{L}| + 4 \right)$ bytes single-precision floating-point values.

\noindent \textit{Derivation:}

For one message update in \eqref{eq:supp-cc-sum-product-2}, the gradient calculation
of terms 1,2-3,4,5 (underlined) requires $4 \times \{|\mathcal{L}|,
|\mathcal{R}| |\mathcal{L}|, 1, |\mathcal{L}|\}$ bytes respectively.
For $K$ iterations, set of directions $\mathcal{R}$, edges $\{\mathcal{E}^r\}$,
$\forall r \in \mathcal{R}$, it requires $K\sum_{r\in\mathcal{R}}|\mathcal{E}^r|
|\mathcal{L}| \left( 8|\mathcal{L}| + 4 |\mathcal{R}| |\mathcal{L}| + 4 \right)$ bytes in total.
This is in $\mathcal{O} (|\mathcal{R}| |\mathcal{L}|)$ order higher than the memory
requirement in the min-sum TRW memory requirement, $K\sum_{r\in\mathcal{R}}|\mathcal{E}^r| \left( |\mathcal{L}| + 1 \right)$ bytes.

%==========================================================================
\section{Additional Evaluations}

%--------------------------------------------------------------------------
\subsection{More Evaluations with Constant Edge Weights}
More results from the main experiments are given in Tables~\myblue{\ref{tb:supp-middlebury}}-\myblue{\ref{tb:supp-kitti}}.

\begin{table*}
\centering
\caption{Energy minimization on Middlebury with constant edge weights. For Map,
ISGMR-4 has the lowest energy among ISGMR-related methods; for others, ISGMR-8 and TRWP-4
have the lowest energies in ISGMR-related and TRWP-related methods respectively. ISGMR is
more effective than SGM in optimization, and TRWP-4 outperforms MF and SGM.}
\resizebox{\textwidth}{!}{\begin{tabular}{l||r|r|r|r|r|r|r|r|r|r}
  \hline
  \multicolumn{1}{c||}{\multirow{2}{*}{\textbf{Method}}} & \multicolumn{2}{c|}{\textbf{Tsukuba}} &
  \multicolumn{2}{c|}{\textbf{Teddy}} & \multicolumn{2}{c|}{\textbf{Venus}} &
  \multicolumn{2}{c|}{\textbf{Cones}} & \multicolumn{2}{c}{\textbf{Map}} \\
  \cline{2-11}
  & \multicolumn{1}{c|}{\textbf{1 iter}}
  & \multicolumn{1}{c|}{\textbf{50 iter}}
  & \multicolumn{1}{c|}{\textbf{1 iter}}
  & \multicolumn{1}{c|}{\textbf{50 iter}}
  & \multicolumn{1}{c|}{\textbf{1 iter}}
  & \multicolumn{1}{c|}{\textbf{50 iter}}
  & \multicolumn{1}{c|}{\textbf{1 iter}}
  & \multicolumn{1}{c|}{\textbf{50 iter}}
  & \multicolumn{1}{c|}{\textbf{1 iter}}
  & \multicolumn{1}{c}{\textbf{50 iter}} \\
  \hline \hline
  MF-4 & 3121704 & 1620524 & 3206347 & 2583784 & 108494928 & 14618819 & 9686122 &
  6379392 & 1116641 & 363221 \\
  SGM-4 & 873777 & 644840 & 2825535 & 2559016 & 5119933 & 2637164 & 3697880 &
  3170715 & 255054 & 216713 \\
  TRWS-4 & 352178 & \underline{314393} & 1855625 & \underline{1807423} & 1325651
  & \lgray 1219774 & 2415087 & \lgray 2329324 & 150853 & \lgray 143197 \\
  ISGMR-4 (ours) & 824694 & 637996 & 2626648 & 1898641 & 4595032 & 1964032 & 3296594
  & 2473646 & 215875 & 148049 \\
  TRWP-4 (ours) & 869363 & \lgray 314037 & 2234163 & \lgray 1806990 & 32896024 &
  \underline{1292619} & 3284868 & \underline{2329343} & 192200 & \underline{143364} \\
  \hline \hline
  MF-8 & 2322139 & 504815 & 3244710 & 2545226 & 68718520 & 2920117 & 7762269 &
  3553975 & 840615 & 213827 \\
  SGM-8 & 776706 & 574758 & 2868131 & 2728682 & 4651016 & 2559933 & 3631020 &
  3309643 & 243058 & 222678 \\
  ISGMR-8 (ours) & 684185 & \lgray 340347 & 2532071 & \lgray 1847833 & 4062167 &
  \lgray 1285330 & 3039638 & \lgray 2398060 & 195718 & \lgray 149857 \\
  TRWP-8 (ours) & 496727 & \underline{348447} & 1981582 & \underline{1849287} &
  8736569 & \underline{1347060} & 2654033 & \underline{2396257} & 162432 & \underline{151970} \\
  \hline
  MF-16 & 1979155 & 404404 & 3315900 & 2622047 & 43077872 & 1981096 & 6741127 & 3062965 & 638753 & 204737 \\
  SGM-16 & 710727 & 587376 & 2907051 & 2846133 & 4081905 & 2720669 & 3564423 & 3413752 & 242932 & 232875 \\
  ISGMR-16 (ours) & 591554 & \lgray 377427 & 2453592 & \lgray 1956343 & 3222851 &
  \lgray 1396914 & 2866149 & \lgray 2595487 & 190847 & \lgray 165249 \\
  TRWP-16 (ours) & 402033 & \underline{396036} & 1935791 & \underline{1976839} &
  2636413 & \underline{1486880} & 2524566 & \underline{2660964} & 162655 & \underline{164704} \\
  \hline
\end{tabular}}
\label{tb:supp-middlebury}
\end{table*}

\begin{table*}
\centering
\caption{Energy minimization on 3 image pairs of KITTI2015 and 2 of ETH-3D with
constant edge weights. ISGMR is more effective than SGM in optimization in both single
and multiple iterations, and TRWP-4 outperforms MF and SGM.}
\label{tb:supp-kitti}
\resizebox{\textwidth}{!}{\begin{tabular}{l||r|r|r|r|r|r|r|r|r|r}
  \hline
  \multicolumn{1}{c||}{\multirow{2}{*}{\textbf{Method}}}
  & \multicolumn{2}{c|}{\textbf{000002\_11}}
  & \multicolumn{2}{c|}{\textbf{000041\_10}}
  & \multicolumn{2}{c|}{\textbf{000119\_10}}
  & \multicolumn{2}{c|}{\textbf{delivery\_area\_1l}}
  & \multicolumn{2}{c}{\textbf{facade\_1s}} \\
  \cline{2-11}
  & \multicolumn{1}{c|}{\textbf{1 iter}}
  & \multicolumn{1}{c|}{\textbf{50 iter}}
  & \multicolumn{1}{c|}{\textbf{1 iter}}
  & \multicolumn{1}{c|}{\textbf{50 iter}}
  & \multicolumn{1}{c|}{\textbf{1 iter}}
  & \multicolumn{1}{c|}{\textbf{50 iter}}
  & \multicolumn{1}{c|}{\textbf{1 iter}}
  & \multicolumn{1}{c|}{\textbf{50 iter}}
  & \multicolumn{1}{c|}{\textbf{1 iter}}
  & \multicolumn{1}{c}{\textbf{50 iter}} \\
  \hline \hline
  MF-4 & 82523536 & 44410056 & 69894016 & 36163508 & 72659040 & 42392548 & 19945352 &
  9013862 & 13299859 & 6681882 \\
  SGM-4 & 24343250 & 18060026 & 15926416 & 12141643 & 24999424 & 18595020 & 5851489 &
  4267990 & 1797314 & 1429254 \\
  TRWS-4 & 9109976 & \lgray 8322635 & 6876291 & \lgray 6491169 & 10811576 &
  \lgray 9669367 & 1628879 & \lgray 1534961 & 891282 & \lgray 851273 \\
  ISGMR-4 (ours) & 22259606 & 12659612 & 14434318 & 9984545 & 23180608 & 18541970
  & 5282024 & 2212106 & 1572377 & 980151 \\
  TRWP-4 (ours) & 40473776 & \underline{8385450} & 30399548 & \underline{6528642}
  & 36873904 & \underline{9765540} & 9899787 & \underline{1546795} & 2851700 & \underline{854552} \\
  \hline
  MF-8 & 61157072 & 18416536 & 53302252 & 16473121 & 57201868 & 21320892 & 16581587
  & 4510834 & 10978978 & 3422296 \\
  SGM-8 & 20324684 & 16406781 & 13740635 & 11671740 & 20771096 & 16652122 & 5396353
  & 4428411 & 1717285 & 1464208 \\
  ISGMR-8 (ours) & 17489158 & \lgray 8753990 & 11802603 & \lgray 6639570 & 18411930
  & \lgray 10173513 & 4474404 & \lgray 1571528 & 1438210 & \lgray 884241 \\
  TRWP-8 (ours) & 18424062 & \underline{8860552} & 13319964 & \underline{6678844}
  & 20581640 & \underline{10445172} & 4443931 & \underline{1587917} & 1358270 & \underline{889907} \\
  \hline
  MF-16 & 46614232 & 14192750 & 40838292 & 12974839 & 44706364 & 16708809 & 13223338
  & 3229021 & 9189592 & 2631006 \\
  SGM-16 & 18893122 & 16791762 & 13252150 & 12162330 & 19284684 & 16936852 & 5092094
  & 4611821 & 1670997 & 1535778 \\
  ISGMR-16 (ours) & 15455787 & \lgray 9556611 & 10731068 & \lgray 6806150 & 16608803
  & \lgray 11037483 & 3689863 & \lgray 1594877 & 1324235 & \lgray 937102 \\
  TRWP-16 (ours) & 11239113 & \underline{9736704} & 8187380 & \underline{6895937}
  & 13602307 & \underline{11309673} & 2261402 & \underline{1630973} & 1000985 & \underline{950607} \\
  \hline
\end{tabular}}
\end{table*}

%--------------------------------------------------------------------------
\subsection{More Visualizations for Image Denoising}
We provide more visualizations of image denoising on ``Penguin" and ``House" in
Figures~\myblue{\ref{fig:denoise_penguin}}-\myblue{\ref{fig:denoise_house}}
corresponding to Table~\myblue{2} in the main paper.

\begin{figure}[]
\centering
\begin{subfigure}[b]{0.15\textwidth}
\includegraphics[width=\textwidth]{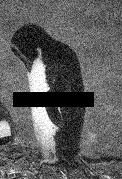}
\caption{noisy}
\end{subfigure}
\begin{subfigure}[b]{0.15\textwidth}
\includegraphics[width=\textwidth]{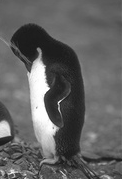}
\caption{GT}
\end{subfigure}
\begin{subfigure}[b]{0.15\textwidth}
\includegraphics[width=\textwidth]{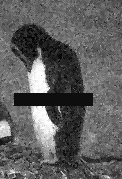}
\caption{MF-4}
\end{subfigure}
\begin{subfigure}[b]{0.15\textwidth}
\includegraphics[width=\textwidth]{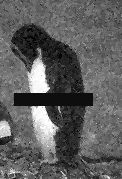}
\caption{MF-8}
\end{subfigure}
\begin{subfigure}[b]{0.15\textwidth}
\includegraphics[width=\textwidth]{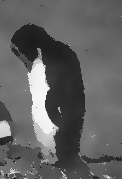}
\caption{MF-16}
\end{subfigure}
\\
\begin{subfigure}[b]{0.15\textwidth}
\includegraphics[width=\textwidth]{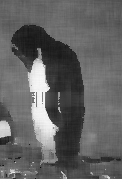}
\caption{SGM-4}
\end{subfigure}
\begin{subfigure}[b]{0.15\textwidth}
\includegraphics[width=\textwidth]{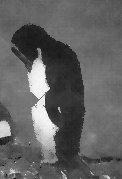}
\caption{SGM-8}
\end{subfigure}
\begin{subfigure}[b]{0.15\textwidth}
\includegraphics[width=\textwidth]{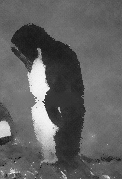}
\caption{SGM-16}
\end{subfigure}
\begin{subfigure}[b]{0.15\textwidth}
\includegraphics[width=\textwidth]{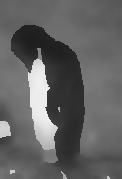}
\caption{TRWS-4}
\end{subfigure}
\begin{subfigure}[b]{0.15\textwidth}
\includegraphics[width=\textwidth]{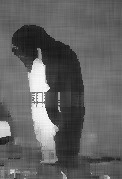}
\caption{ISGMR-4}
\end{subfigure}
\\%
\begin{subfigure}[b]{0.15\textwidth}
\includegraphics[width=\textwidth]{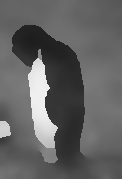}
\caption{ISGMR-8}
\end{subfigure}
\begin{subfigure}[b]{0.15\textwidth}
\includegraphics[width=\textwidth]{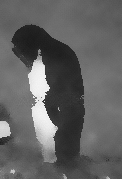}
\caption{\scriptsize ISGMR-16}
\end{subfigure}
\begin{subfigure}[b]{0.15\textwidth}
\includegraphics[width=\textwidth]{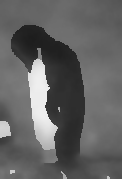}
\caption{TRWP-4}
\end{subfigure}
\begin{subfigure}[b]{0.15\textwidth}
\includegraphics[width=\textwidth]{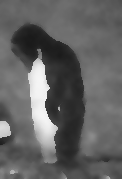}
\caption{TRWP-8}
\end{subfigure}
\begin{subfigure}[b]{0.15\textwidth}
\includegraphics[width=\textwidth]{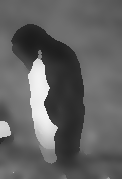}
\caption{TRWP-16}
\end{subfigure}
\caption{Visualization of MRF inferences for image denoising on ``Penguin".}
\label{fig:denoise_penguin}
\end{figure}

\begin{figure}[]
\centering
\begin{subfigure}[b]{0.15\textwidth}
\includegraphics[width=\textwidth]{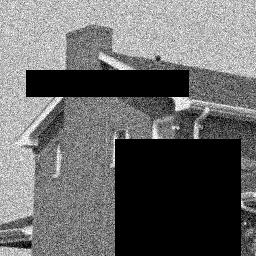}
\caption{noisy}
\end{subfigure}
\begin{subfigure}[b]{0.15\textwidth}
\includegraphics[width=\textwidth]{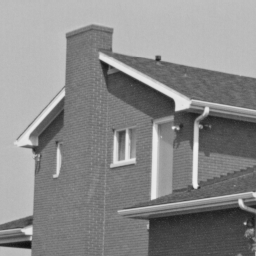}
\caption{GT}
\end{subfigure}
\begin{subfigure}[b]{0.15\textwidth}
\includegraphics[width=\textwidth]{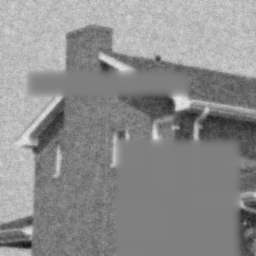}
\caption{MF-4}
\end{subfigure}
\begin{subfigure}[b]{0.15\textwidth}
\includegraphics[width=\textwidth]{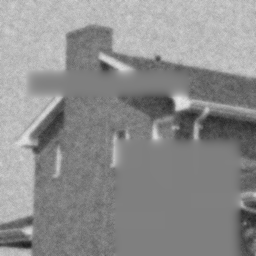}
\caption{MF-8}
\end{subfigure}
\begin{subfigure}[b]{0.15\textwidth}
\includegraphics[width=\textwidth]{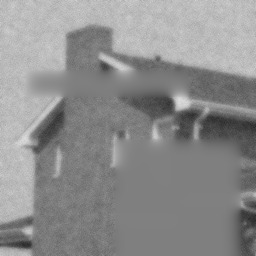}
\caption{MF-16}
\end{subfigure}
\\
\begin{subfigure}[b]{0.15\textwidth}
\includegraphics[width=\textwidth]{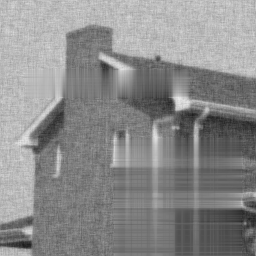}
\caption{SGM-4}
\end{subfigure}
\begin{subfigure}[b]{0.15\textwidth}
\includegraphics[width=\textwidth]{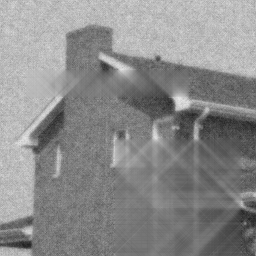}
\caption{SGM-8}
\end{subfigure}
\begin{subfigure}[b]{0.15\textwidth}
\includegraphics[width=\textwidth]{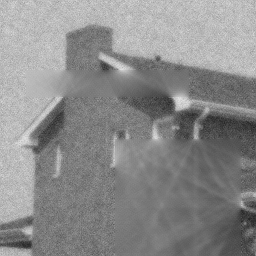}
\caption{SGM-16}
\end{subfigure}
\begin{subfigure}[b]{0.15\textwidth}
\includegraphics[width=\textwidth]{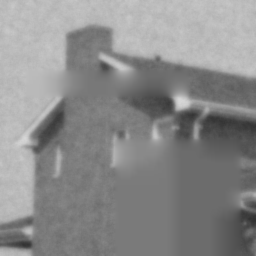}
\caption{TRWS-4}
\end{subfigure}
\begin{subfigure}[b]{0.15\textwidth}
\includegraphics[width=\textwidth]{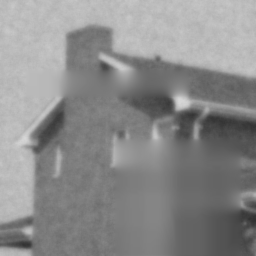}
\caption{ISGMR-4}
\end{subfigure}
\\%
\begin{subfigure}[b]{0.15\textwidth}
\includegraphics[width=\textwidth]{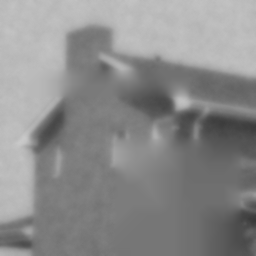}
\caption{ISGMR-8}
\end{subfigure}
\begin{subfigure}[b]{0.15\textwidth}
\includegraphics[width=\textwidth]{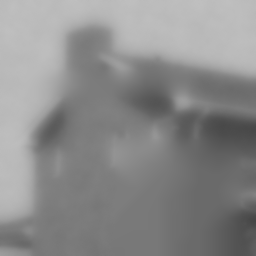}
\caption{\scriptsize ISGMR-16}
\end{subfigure}
\begin{subfigure}[b]{0.15\textwidth}
\includegraphics[width=\textwidth]{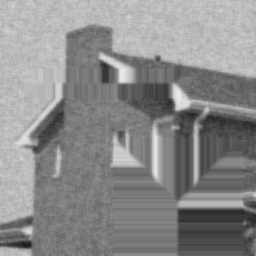}
\caption{TRWP-4}
\end{subfigure}
\begin{subfigure}[b]{0.15\textwidth}
\includegraphics[width=\textwidth]{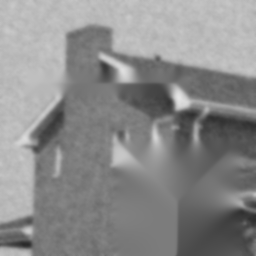}
\caption{TRWP-8}
\end{subfigure}
\begin{subfigure}[b]{0.15\textwidth}
\includegraphics[width=\textwidth]{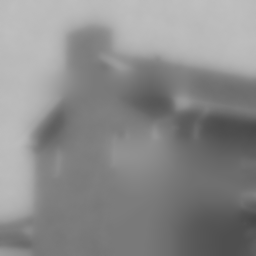}
\caption{TRWP-16}
\end{subfigure}
\caption{Visualization of MRF inferences for image denoising on ``House".}
\label{fig:denoise_house}
\end{figure}
\else
    %==========================================================================
\begin{abstract}
Despite the availability of many Markov Random Field (MRF) optimization
algorithms, their widespread usage is currently limited due to imperfect
MRF modelling arising from hand-crafted model parameters and the selection
of inferior inference algorithm.
In addition to differentiability, the two main aspects that enable
learning these model parameters are the forward and backward propagation
time of the MRF optimization algorithm and its inference capabilities.
In this work, we introduce two fast and differentiable message passing
algorithms, namely, Iterative Semi-Global Matching Revised (ISGMR) and
Parallel Tree-Reweighted Message Passing (TRWP) which are greatly sped
up on a GPU by exploiting massive parallelism.
Specifically, ISGMR is an iterative and revised version of the standard
SGM for general pairwise MRFs with improved optimization effectiveness,
and TRWP is a highly parallel version of Sequential TRW (TRWS) for faster
optimization.
Our experiments on the standard stereo and denoising benchmarks
demonstrated that ISGMR and TRWP achieve much lower energies than SGM and
Mean-Field (MF), and TRWP is two orders of magnitude faster than TRWS
without losing effectiveness in optimization.
We further demonstrated the effectiveness of our algorithms on end-to-end
learning for semantic segmentation.
Notably, our CUDA implementations are at least $7$ and $700$ times faster
than PyTorch GPU implementations for forward and backward propagation
respectively, enabling efficient end-to-end learning with message passing.
\end{abstract}

%==========================================================================
\section{Introduction}
Optimization of Markov Random Fields (MRFs) has been a well-studied problem
for decades
with a significant impact on many computer vision applications such as
stereo vision \cite{sgm}, image segmentation \cite{grab_cut}, texture
modeling \cite{texture_modeling}.
The widespread use of these MRF optimization algorithms is currently
limited due to imperfect MRF modelling \cite{mrf_study} because of
hand-crafted model parameters, the usage
of inferior inference methods, and non-differentiability
for parameter learning.
Thus, better inference capability and computing efficiency are essential 
to improve its performance on optimization and modelling, such as
energy optimization and end-to-end learning.

Even though parameter and structural learning with MRFs has been employed
successfully in certain cases, well-known algorithms such as Mean-Field
(MF)~\cite{crfasrnn,densecrf} and Semi-Glocal Matching
(SGM)~\cite{sgm-net}, are suboptimal in terms of optimization capability.
Specifically, the choice of an MRF algorithm for optimization is driven
by its inference ability, and for learning capability through efficient
forward and backward propagation and parallelization capabilities.

In this work, we consider message passing algorithms due to their
generality, high inference ability, and differentiability, and provide
efficient CUDA implementations of their forward and backward propagation
by exploiting massive parallelism.
In particular, we revise the popular SGM method \cite{sgm} and derive an
iterative version noting its relation to traditional message passing
algorithms~\cite{derivation}.
In addition, we introduce a highly parallelizable version of the
state-of-the-art Sequential Tree-Reweighted Message Passing (TRWS)
algorithm~\cite{trws}, which is more efficient than TRWS and has similar
minimum energies.
For both these methods, we derive efficient backpropagation by unrolling
their message updates and cost aggregation and discuss massively parallel
CUDA implementations which enable their feasibility in end-to-end learning.

Our experiments on the standard stereo and denoising benchmarks demonstrate
that our Iterative and Revised SGM method (ISGMR) obtains much lower
energies compared to the standard SGM and our Parallel TRW method (TRWP)
is two orders of magnitude faster than TRWS with virtually the same minimum
energies and that both outperform the popular MF and SGM inferences.
Their performance is further evaluated by end-to-end learning for semantic
segmentation on PASCAL VOC 2012 dataset.

Furthermore, we empirically evaluate various implementations of the forward
and backward propagation of these algorithms and demonstrate that our CUDA
implementation is the fastest, with {\em at least 700 times speed-up} in
backpropagation compared to a PyTorch GPU version.
Code is available at \textit{\url{https://github.com/zwxu064/MPLayers.git}}.

Contributions of this paper can be summarised as:

\begin{itemize}
  \item[$\bullet$] We introduce two message passing algorithms, ISGMR and TRWP, where
ISGMR has higher optimization effectiveness than SGM and TRWP is much faster
than TRWS.
Both of them outperform the popular SGM and MF inferences.
  \item[$\bullet$] Our ISGMR and TRWP are massively parallelized on GPU and can support
any pairwise potentials.
The CUDA implementation of the backpropagation is at least $700$ times
faster than the PyTorch auto-gradient version on GPU.
  \item[$\bullet$] The differentiability of ISGMR and TRWP is presented with gradient
derivations, with effectiveness validated by end-to-end learning for
semantic segmentation.
\end{itemize}

%==========================================================================
\section{Related Work}
In MRF optimization, estimating the optimal latent variables can be regarded as
minimizing a particular energy function with given model parameters. Even if the
minimum energy is obtained, high accuracy cannot be guaranteed since the model
parameters of these MRFs are usually handcrafted and imperfect. To tackle this
problem, learning-based methods were proposed. However, most of these methods
rely greatly on finetuning the network architecture or adding learnable
parameters to increase the fitting ability with ground truth. This may not be
effective and usually requires high GPU memory.

Nevertheless, considering the highly effective MRF optimization algorithms, the
field of exploiting their optimization capability with parameter learning to
alleviate each other's drawbacks is rarely explored. A few works provide this
capability in certain cases, such as CRFasRNN in semantic
segmentation~\cite{crfasrnn} and SGMNet in stereo vision~\cite{sgm-net}, with
less effective MRF algorithms, that is MF and SGM respectively. Thus, it is
important to adopt highly effective and efficient MRF inference algorithms for
optimization and end-to-end learning.

\textbf{MRF Optimization.} Determining an effective MRF optimization algorithm
needs a thorough study of the possibility of their optimization capability,
differentiability, and time efficiency. In the two main categories of MRF
optimization algorithms, namely move-making algorithms (known as graph
cuts)~\cite{memf,irgc,expansion,multi-swap,gen-range-move,range-move} and
message passing algorithms~\cite{sgm,jordan-book,accv_2010,trws,lbp,lbp-0,jordan-report,accv_2014_best},
the state-of-the-art methods are $\alpha$-expansion \cite{expansion} and
Sequential Tree-Reweighted Message Passing (TRWS)~\cite{trws} respectively. The
move-making algorithms, however, cannot easily be used for parameter learning as
they are not differentiable and are usually limited to certain types of energy
functions.

In contrast, message passing algorithms adapt better to any energy
functions and can be made differentiable and fast if well designed. Some works in
probabilistic graphical models indeed demonstrate the learning ability of TRW
algorithms with sum-product and max-product~\cite{jordan-book,jordan-report}
message passing.
A comprehensive study and comparison of these methods can be found in
Middlebury \cite{mrf_study} and OpenGM \cite{openGM}.
Although SGM \cite{sgm} is not in the benchmark, it was proved to have a high
running efficiency due to the fast one-dimensional Dynamic Programming (DP) that
is independent in each scanline and scanning direction \cite{sgm}.

\textbf{End-to-End Learning.} Sum-product
TRW~\cite{justin,max-margin,large-margin} and
mean-field~\cite{crfasrnn,dpn,piecewise} have been used for end-to-end learning for
semantic segmentation, which presents their highly effective learning ability.
Meanwhile, for stereo vision, several MRF/CRF based methods
~\cite{sgm-net,ganet,cnn_crf_stereo}, such as SGM-related, have been proposed.
These further indicate the high efficiency of selected MRF optimization
algorithms in end-to-end learning.

In our work, we improve optimization effectiveness and time efficiency based on
classical SGM and TRWS. In particular, we revise the standard SGM and make it
iterative in order to improve its optimization capability.  
We denote the resulting algorithm as ISGMR. Our other algorithm, TRWP,
is a massively parallelizable version of TRWS, which
greatly increases running speed without losing the optimization
effectiveness.

%==========================================================================
\section{Message Passing Algorithms}  \label{sec:algorithm}
We first briefly review the typical form of a pairwise MRF energy function and
discuss two highly parallelizable message passing approaches, ISGMR and TRWP.
Such a parallelization capability is essential for fast implementation on GPU
and enables relatively straightforward integration to existing deep learning
models.

%-------------------------------------------------------------------------
\subsection{Pairwise MRF Energy Function}
Let $X_i$ be a random variable taking label $x_i \in \mathcal{L}$. 
A pairwise MRF energy function defined over a set of such variables,
parametrized by $\mathbf{\Theta}= \{\theta_i, \theta_{i,j}\}$, is written as
\begin{equation}
\label{eq:energy-function}
  E(\mathbf{x} ~|~ \mathbf{\Theta}) = \sum_{i \in \mathcal{V}} \theta_{i}(x_i) +
\sum_{(i, j) \in \mathcal{E}} \theta_{i,j}(x_i, x_j)\ ,
\end{equation}
where $\theta_{i}$ and $\theta_{i,j}$ denote unary potentials and pairwise
potentials respectively,
$\mathcal{V}$ is the set of vertices (corresponding, for instance, to image pixels or
superpixels), and $\mathcal{E}$ is the set of edges in the MRF (usually encoding a
4-connected or 8-connected grid).

%-------------------------------------------------------------------------
\subsection{Iterative Semi-Global Matching Revised}
We first introduce the standard SGM for stereo vision supporting only a single
iteration. With its connection to message passing, we then revise its message
update equation and introduce an iterative version.
\myfig{fig:directions} shows a 4-connected SGM on a grid MRF.

%-------------------------------------------------------------------------
\subsubsection{Revised Semi-Global Matching.}
We cast the popular SGM algorithm~\cite{sgm} as an optimization method for a
particular MRF and discuss its relation to message passing as noted
in~\cite{derivation}. 
In SGM, pairwise potentials are simplified for all edges $ (i,j) \in
\mathcal{E}$ as
\begin{equation}\label{eq:vsgm}
\theta_{i,j}(\lambda,\mu) = \theta_{i, j}(|\lambda- \mu|) = \left\{
\begin{array}{ll}
0  & \mbox{if $\lambda=\mu$}\ ,\\ 
 P_1  & \mbox{if $|\lambda-\mu|=1$}\ ,\\
 P_2  & \mbox{if $|\lambda-\mu|\ge2$}\ ,
 \end{array}
 \right.
\end{equation}
where $0 < P_1 \le P_2$.
The idea of SGM relies on cost aggregation in multiple directions (each
direction having multiple one-dimensional scanlines) using Dynamic Programming
(DP).
The main observation made by~\cite{derivation} is that, in SGM the unary
potentials are over-counted $|\mathcal{R}|-1$ times (where $\mathcal{R}$ denotes
the set of directions) compared to the standard message passing
and this over-counting corrected SGM is shown to perform slightly better
in~\cite{mgm}.
Noting this, we use symbol $\mesnew{r}{i}{\lambda}$ to denote the message-vector 
passed {\bf to}
node $i$, along a scan-line in the direction $r$, {\bf from} the previous
node, denoted $i-r$. This is a vector indexed by $\lambda \in \mathcal{L}$.
Now, the SGM update is \textit{revised} from
\begin{align}
\label{eq:sgm_old}
  \mesnew{r}{i}{\lambda} &= \min_{\mu \in \mathcal{L}}\, \big(\theta_{i}{(\lambda)} +
\mesnew{r}{i-r}{\mu} + \theta_{i-r,i}(\mu,\lambda)\big) \ ,
\end{align}
which is the form given in \cite{sgm}, to
\begin{align}
\label{eq:sgm}
  \mesnew{r}{i}{\lambda} &= \min_{\mu \in \mathcal{L}}\, \big(\theta_{i-r}{(\mu)} +
\mesnew{r}{i-r}{\mu} + \theta_{i-r, i}(\mu,\lambda)\big) \ .
\end{align}
The $\mesnew{r}{i}{\lambda}$ represents the minimum cost due to possible assignments to
all nodes previous to node $i$ along the scanline in direction $r$, and 
assigning label $\lambda$ to node $i$.
It does not include the cost $\theta_i(\lambda)$ associated with node $i$ itself.

Since subtracting a fixed value for all $\lambda$ from
messages preserves minima, the
message $\mesnew{r}{i}{\lambda}$ can be reparametrized as
\begin{equation}
\label{eq:isgmr-norm}
\mesnew{r}{i}{\lambda} = \mesnew{r}{i}{\lambda} - \min_{\mu \in \mathcal{L}} \mesnew{r}{i}{\mu}\ ,
\end{equation}
which does not alter the minimum energy.
Since the values of $\theta_i(\lambda)$ are not included
in the messages, the final cost at a particular node $i$ at label $\lambda$ is
\textit{revised} from

\begin{equation}
\label{eq:isgmr-aggregation-old}
  c_{i}{(\lambda)} = \sum_{r \in \mathcal{R}} \mesnew{r}{i}{\lambda}
\end{equation}
to
\begin{equation}
\label{eq:isgmr-aggregation}
  c_{i}{(\lambda)} = \theta_{i}{(\lambda)} + \sum_{r \in \mathcal{R}}
\mesnew{r}{i}{\lambda}\ ,
\end{equation}
which is the sum of messages over all the directions plus the unary term.
The final labelling is then obtained by
\begin{equation}
\label{eq:disp}
\begin{aligned}
  x_i^* = \argmin_{\lambda \in \mathcal{L}} c_{i}{(\lambda)}\ ,\quad \forall\, i
\in \mathcal{V}\ .
\end{aligned}
\end{equation}
Here, the message update in the revised SGM, \ie, \eqref{eq:sgm}, is performed in
parallel for all scanlines for all directions. 
This massive parallelization makes it suitable for real-time
applications~\cite{real-time} and end-to-end learning for
stereo vision~\cite{sgm-net}.

\begin{figure}[t]
\centering
\begin{subfigure}[b]{0.25\linewidth}
\includegraphics[width=\linewidth]{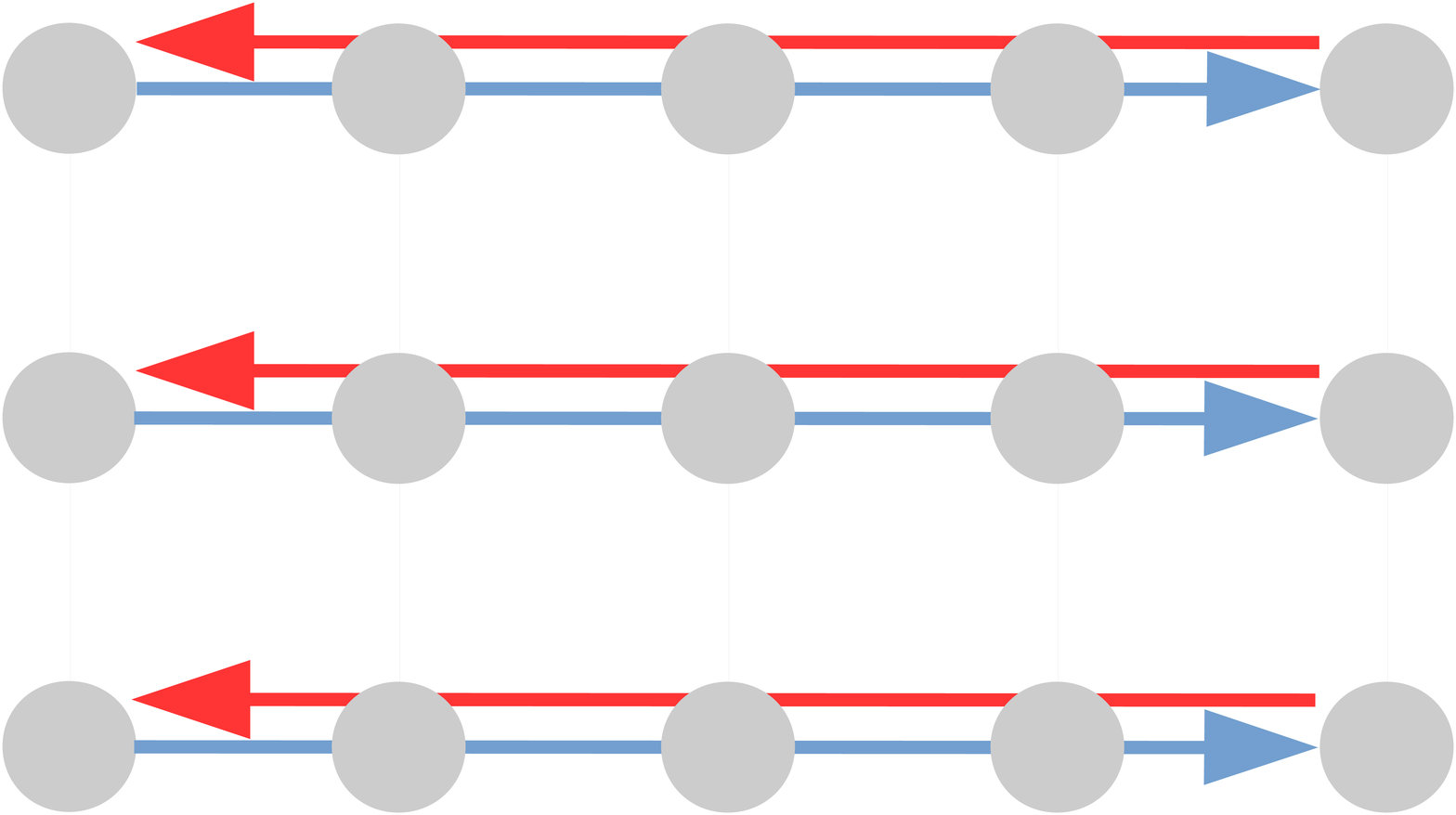}
\end{subfigure}
\qquad \qquad
\begin{subfigure}[b]{0.25\linewidth}
\includegraphics[width=\linewidth]{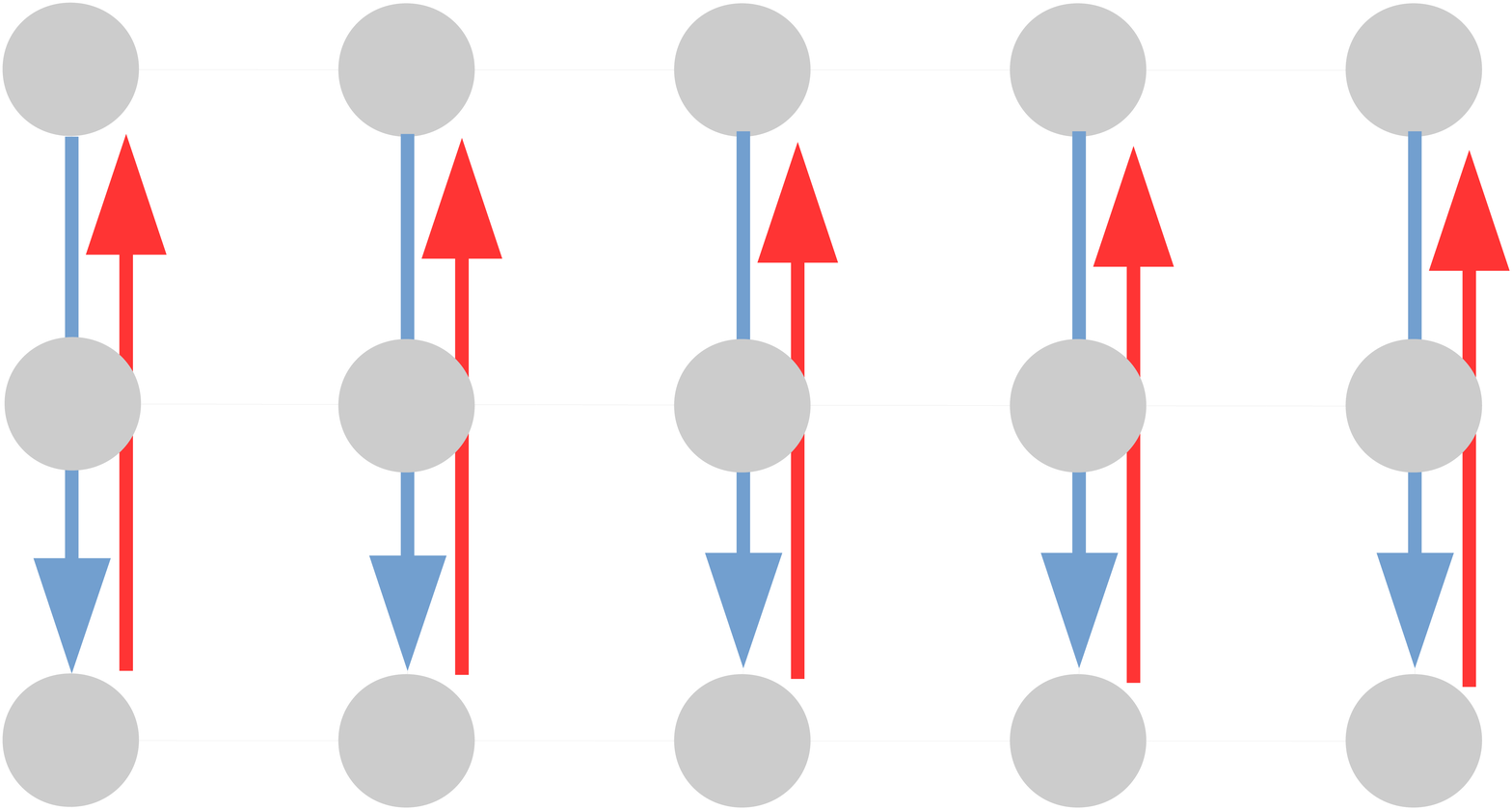}
\end{subfigure}
\caption{An example of 4-connected SGM on a grid MRF: left-right,
right-left, up-down, down-up. Message passing along all these scanlines
can be accomplished in parllel.}
\label{fig:directions}
\end{figure}

%--------------------------------------------------------------------------
\subsubsection{Iteration of Revised Semi-Global Matching.}
\label{sec:isgmr}
In spite of the revision for the over-counting problem, the 3-penalty pairwise
potential in \eqref{eq:vsgm} is insufficient to obtain dominant penalties under
a large range of disparities in different camera settings.
To this end, we consider more general pairwise potentials
$\theta_{i,j}(\lambda,\mu)$ and introduce an iterative version of the revised
SGM.
The message update for the iterative version is

\begin{equation}
\label{eq:isgmr}
 \mesnew{r,k+1}{i}{\lambda} = \min_{\mu \in \mathcal{L}} \big(\theta_{i-r}{(\mu)} +
 \theta_{i-r,i}(\mu,\lambda) + \mesnew{r,k+1}{i-r}{\mu} + \sum_{d \in \mathcal{R} \setminus \{r,r^-\}}
\mesnew{d,k}{i-r}{\mu}\big),
\end{equation}
where $r^-$ denotes the opposite direction of $r$ and $\mesnew{r,k+1}{i-r}{\mu}$
denotes the updated message in $k$th iteration while $\mesnew{r,k}{i-r}{\mu}$ is
updated in $(k-1)$th iteration.
The exclusion of the messages from direction $r^{-}$ is important to ensure that
the update is analogous to the standard message passing and the same energy
function is minimized at each iteration. A simple combination of several
standard SGMs does not satisfy this rule and performs worse than our iterative
version, as reported in~Tables~\myblue{\ref{tb:middlebury}-\ref{tb:denoise}}.
Usually, $\mathbf{m}^r{}$ for all $r\in\mathcal{R}$ are initialized to
$0$, the exclusion of $r^-$ from $\mathcal{R}$ is thus redundant for a single iteration
but not multiple iterations.
Even so, messages can be reparametrized by \eqref{eq:isgmr-norm}.

After multiple iterations, the final cost for node $i\in\mathcal{V}$ is
calculated by \eqref{eq:isgmr-aggregation},
and the final labelling is calculated in the same manner as \eqref{eq:disp}.
We denote this iterative and revised SGM as ISGMR, summarized in
\myalg{alg:isgmr}.

In sum, the improvement of ISGMR from SGM lies in the exclusion of over-counted unary
terms by \eqref{eq:sgm} to increase the effects of pairwise terms
as well as the iterative energy minimization by \eqref{eq:isgmr} to further
decrease the energy with updated messages.

\SetInd{0.5em}{0.5em}
\newlength{\textfloatsepsave}
\setlength{\textfloatsepsave}{\textfloatsep} 
\setlength{\textfloatsep}{0.5ex}
\begin{algorithm}[t]
\KwIn{Energy parameters $\mathbf{\Theta} = \{\theta_i, \theta_{i,j}(\cdot,
\cdot)\}$, set of nodes $\mathcal{V}$, edges $\mathcal{E}$, directions
$\mathcal{R}$, iteration number $K$.
We replace $m^{r,k}$ by $m^r$ and $m^{r,k+1}$ by $\hat{m}^{r}$ for simplicity.}
\KwOut{Labelling $\mathbf{x}^*$ for optimization, costs $\{c_{i}{(\lambda)}\}$ for
learning, indices $\{p^r_{k,i}{(\lambda)}\}$ and $\{q^r_{k,i}\}$ for
backpropagation.}
\caption{Forward Propagation of ISGMR} \label{alg:isgmr}
$\mathbf{\hmes{}{}}\gets 0$ and $\mathbf{\mes{}{}}\gets 0$ \tcp*[f]{initialize
all messages}\\
\For{\textup{iteration $k \in \{1,\ldots,K\}$}}{
  \ForAll(\tcp*[f]{\textbf{parallel}}){\textup{directions $r \in
\mathcal{R}$}}{
    \ForAll(\tcp*[f]{\textbf{parallel}}){\textup{scanlines $t$ in direction
$r$}}{
      \For(\tcp*[f]{\textbf{sequential}}){\textup{node $i$ in scanline $t$}}{
        \For{\textup{label $\lambda \in \mathcal{L}$}}{
          $ \setlength\belowdisplayskip{0pt}
            \begin{aligned}
              \Delta{(\lambda,\mu)} \gets &\theta_{i-r}{(\mu)}
              + \theta_{i-r,i}(\mu,\lambda)
              + \hmesnew{r}{i-r}{\mu}
+\sum_{d \in \mathcal{R} \setminus \{r, r^-\}}
\mesnew{d}{i-r}{\mu}
            \end{aligned} $ \\
          $p^r_{k,i}{(\lambda)} \gets \mu^* \gets
\argmin_{\mu\in\mathcal{L}}\Delta{(\lambda,\mu)}$ \tcp*[f]{store index} \\
          $\hmesnew{r}{i}{\lambda} \gets \Delta{(\lambda,\mu^*)}$ \tcp*[f]{message
update~\plaineqref{eq:isgmr}}
        }
        $q^r_{k,i} \gets \lambda^* \gets \argmin_{\lambda \in \mathcal{L}}
\hmesnew{r}{i}{\lambda}$ \tcp*[f]{store index} \\
        $\hmesnew{r}{i}{\lambda} \gets \hmesnew{r}{i}{\lambda} - \hmesnew{r}{i}{\lambda^*}$
\tcp*[f]{reparametrization~\plaineqref{eq:isgmr-norm}} \\
      }
    }
  }
  $\mathbf{\mes{}{}} \gets \mathbf{\hmes{}{}}$ \tcp*[f]{update messages after
iteration}
}
$c_{i}{(\lambda)} \gets \theta_{i}{(\lambda)} + \sum_{r \in \mathcal{R}}
\mesnew{r}{i}{\lambda}, \forall i \in \mathcal{V}, \lambda \in \mathcal{L}$
\tcp*[f]{\eqref{eq:isgmr-aggregation}} \\
$x_{i}^* \gets \argmin_{\lambda \in \mathcal{L}} c_{i}{(\lambda)}, \forall i \in
\mathcal{V}$ \tcp*[f]{\eqref{eq:disp}} \\
\end{algorithm}

%-------------------------------------------------------------------------
\subsection{Parallel Tree-Reweighted Message Passing}
TRWS~\cite{trws} is another state-of-the-art message passing
algorithm that optimizes the Linear Programming (LP) relaxation of a general
pairwise MRF energy given in~\eqref{eq:energy-function}.
The main idea of the family of TRW algorithms~\cite{trw-t} is to decompose the
underlying graph ${\mathcal{G}=(\mathcal{V}, \mathcal{E})}$ of the MRF with
parameters $\mathbf{\Theta}$ into a combination of trees where the sum of
parameters of all the trees is equal to that of the MRF, \ie,
$\sum_{T\in\mathcal{T}} \mathbf{\Theta}_T = \mathbf{\Theta}$.
Then, at each iteration message passing is performed in each of these trees
independently, followed by an averaging operation.
Even though any combinations of trees would theoretically result in the same
final labelling, the best performance is achieved by choosing a monotonic chain
decomposition and a sequential message passing update rule, which is TRWS.
Interested readers please refer to~\cite{trws} for more details.

Since we intend to enable fast message passing by exploiting parallelism, our
idea is to choose a tree decomposition that can be massively parallelized,
denoted as TRWP.
In the literature, edge-based or tree-based parallel TRW algorithms have been
considered, namely, TRWE and TRWT in the probability space (specifically
sum-product message passing) rather than for minimizing the
energy~\cite{trw-t}.
Optimizing in the probability domain involves exponential calculations which are
prone to numerical instability, and the sum-product version requires
$\mathcal{O}(|\mathcal{R}||\mathcal{L}|)$ times more memory compared to the
min-sum message passing in backpropagation. More details are in
Appendix~\myref{E}.

\setlength{\textfloatsep}{0.5ex}
\begin{algorithm}[t]
\KwIn{Energy parameters $\mathbf{\Theta} = \{\theta_i, \theta_{i,j}(\cdot,
\cdot)\}$, set of nodes $\mathcal{V}$, edges $\mathcal{E}$, directions
$\mathcal{R}$, tree decomposition coefficients $\{\rho_{i,j}\}$, iteration number $K$.}
\KwOut{Labelling $\mathbf{x}^*$ for optimization, costs $\{c_{i}{(\lambda})\}$ for
learning, indices $\{p^r_{k,i}{(\lambda)}\}$ and $\{q^r_{k,i}\}$ for
backpropagation.}
\caption{Forward Propagation of TRWP} \label{alg:trwp}
$\mathbf{\mes{}{}}\gets 0$ \tcp*[f]{initialize all messages} \\
\For{\textup{iteration $k \in \{1,\ldots,K\}$}}{
  \For(\tcp*[f]{\textbf{sequential}}){\textup{direction $r \in \mathcal{R}$}}{
    \ForAll(\tcp*[f]{\textbf{parallel}}){\textup{scanlines $t$ in direction
$r$}}{
      \For(\tcp*[f]{\textbf{sequential}}){\textup{node $i$ in scanline $t$}}{
        \For{\textup{label $\lambda \in \mathcal{L}$}}{
          $ \setlength\belowdisplayskip{0pt}
            \begin{aligned}
              \Delta{(\lambda,\mu)} \gets &\rho_{i-r,i} \big(\theta_{i-r}{(\mu)} +
\sum_{d \in \mathcal{R}}\mesnew{d}{i-r}{\mu}\big) 
              - \mesnew{r^-}{i-r}{\mu} + \theta_{i-r,i}(\mu,\lambda)
            \end{aligned} $ \\
            $p^r_{k,i}{(\lambda)} \gets \mu^* \gets
\argmin_{\mu\in\mathcal{L}}\Delta{(\lambda,\mu)}$ \tcp*[f]{store index} \\
            $\mesnew{r}{i}{\lambda} \gets \Delta{(\lambda,\mu^*)}$ \tcp*[f]{message
update~\plaineqref{eq:trwp}} \\
          }
          $q^r_{k,i} \gets \lambda^* \gets \argmin_{\lambda \in \mathcal{L}}
\mesnew{r}{i}{\lambda}$  \tcp*[f]{store index} \\
          $\mesnew{r}{i}{\lambda} \gets \mesnew{r}{i}{\lambda} - \mesnew{r}{i}{\lambda^*}$
\tcp*[f]{reparametrization~\plaineqref{eq:isgmr-norm}} \\
      }
    }
  }
}
$c_{i}{(\lambda)} \gets \theta_{i}{(\lambda)} + \sum_{r \in \mathcal{R}}
\mesnew{r}{i}{\lambda}, \forall i \in \mathcal{V}, \lambda \in \mathcal{L}$
\tcp*[f]{\eqref{eq:isgmr-aggregation}} \\
$x_{i}^* \gets \argmin_{\lambda \in \mathcal{L}} c_{i}{(\lambda)}, \forall i \in
\mathcal{V}$ \tcp*[f]{\eqref{eq:disp}} \\
\end{algorithm}

Correspondingly, our TRWP directly minimizes the energy in the min-sum message
passing fashion similar to TRWS, and thus, its update can be written as

\begin{equation}
\label{eq:trwp}
  \mesnew{r}{i}{\lambda} = \min_{\mu \in \mathcal{L}} \ \big(\rho_{i-r,i}
(\theta_{i-r}{(\mu)} + \sum_{d \in \mathcal{R}}\mesnew{d}{i-r}{\mu}) -
\mesnew{r^-}{i-r}{\mu} + \theta_{i-r,i}(\mu,\lambda)\big).
\end{equation}
Here, the coefficient $\rho_{i-r,i} = \gamma_{i-r,i}/\gamma_{i-r}$, where
$\gamma_{i-r,i}$ and $\gamma_{i-r}$ are 
the number of trees containing the edge $(i-r,i)$ and the node $i-r$
respectively in the considered tree decomposition.
For loopy belief propagation, since there is no tree decomposition,
${\rho_{i-r,i} = 1}$.
For a 4-connected graph decomposed into all horizontal and vertical
one-dimensional trees, we have $\rho_{i-r,i}=0.5$ for all edges.

Note that, similar to ISGMR, we use the scanline to denote a tree.
The above update can be performed in parallel for all scanlines in a single
direction; however, the message updates over a scanline are sequential.
The same reparametrization~\eqref{eq:isgmr-norm} is applied.
While TRWP cannot guarantee the non-decreasing monotonicity of the lower bound
of energy, it dramatically improves the forward propagation speed and yields
virtually similar minimum energies to those of TRWS.
The procedure is in \myalg{alg:trwp}.

In sum, our TRWP benefits from a high speed-up without losing optimization
capability by the massive GPU parallelism over individual trees that are
decomposed from the single-chain tree in TRWS.
All trees in each direction $r$ are paralleled by \eqref{eq:trwp}.

%-------------------------------------------------------------------------
\subsection{Relation between ISGMR and TRWP} \label{subsec:mpvs}
Both ISGMR and TRWP use messages from neighbouring nodes to perform recursive and
iterative message updates via dynamic programming.
Comparison of~\eqref{eq:isgmr} and~\eqref{eq:trwp} indicates the introduction of
the coefficients $\{\rho_{i-r,i}\}$.
This is due to the tree decomposition, which is analogous to the difference
between loopy belief propagation and TRW algorithms.
The most important difference, however, is the way message updates are
defined.
Specifically, within an iteration, ISGMR can be parallelized over all directions
since the most updated messages $\hat{\mathbf{m}}^r$ are used only for the
current scanning direction $r$ and previous messages are used for the other
directions (refer~\eqref{eq:isgmr}).
In contrast, aggregated messages in TRWP are up-to-date {\em direction-by-direction},
which largely contributes to the improved
effectiveness of TRWP over ISGMR.

%-------------------------------------------------------------------------
\subsection{Fast Implementation by Tree Parallelization}
Independent trees make the parallelization possible.
We implemented on CPU and GPU, where for the C++ multi-thread versions (CPU), 8
threads on Open Multi-Processing (OpenMP) \cite{openmp} are used while for
the CUDA versions (GPU), 512 threads per block are used. Each tree is headed
by its first node by interpolation.
The node indexing details for efficient parallelism are provided in
Appendix~\myref{C}.
In the next section, we derive efficient backpropagation through each of these
algorithms for parameter learning.

%-------------------------------------------------------------------------
\section{Differentiability of Message Passing} \label{sec:differentiability}

Effective and differentiable MRF optimization algorithms can greatly improve the
performance of end-to-end learning. 
Typical methods such as CRFasRNN for semantic segmentation~\cite{crfasrnn} by MF
and SGMNet for stereo vision~\cite{sgm-net} by SGM use inferior inferences
in the optimization capability compared to ISGMR and TRWP.

In order to embed ISGMR and TRWP into end-to-end learning, differentiability of
them is required and essential. Below, we describe the gradient updates for the
learnable MRF model parameters, and detailed derivations are given in
Appendix~\myref{D}.
The backpropagation pseudocodes are in
% Algorithms~\myblue{\ref{alg:isgmr-back}}-\myblue{\ref{alg:trwp-back}}
Algorithms 3-4 in Appendix A.

Since ISGMR and TRWP use
min-sum message passing, no exponent and logarithm
are required. Only indices in message minimization and reparametrization are
stored in two unsigned 8-bit integer tensors, denoted as $\{p^r_{k,i}{(\lambda})\}$
and $\{q^r_{k,i}\}$ with indices of direction $r$, iteration $k$, node $i$, and
label $\lambda$. This makes the backpropagation time less than 50\% of the
forward propagation time. In \myfig{fig:flow}, the gradient updates in
backpropagation are performed along edges that have the minimum messages in the
forward direction.
In \myfig{fig:accumulation}, a message gradient at node $i$ is accumulated from
all following nodes after $i$ from all backpropagation directions.
Below, we denote the gradient of a variable $*$ from loss $L$ as $\nabla * = dL
/ d*$.

For ISGMR at $k$th iteration, the gradients of the model parameters in \eqref{eq:isgmr}
are
\begin{equation}
\begin{aligned}
\label{eq:isgmr-back-unary}
\nabla \theta_{i}{(\lambda)}
=& \nabla c_{i}{(\lambda)} + \sum_{v \in \mathcal{L}} \sum_{r \in \mathcal{R}}
\sum_{\mu \in \mathcal{L}} \Big( \left. \nabla \mesnew{r,k+1}{i+2r}{\mu} \right
\vert_{v = p^r_{k,i+2r}{(\mu)}} \\
 &+ \left. \sum_{d \in \mathcal{R} \setminus \{r, r^{-}\}} \left.
\nabla \mesnew{d,k}{i+r+d}{\mu} \right \vert_{v = p^d_{k,i+r+d}{(\mu)}} \Big) \right
\vert_{\lambda = p^r_{k,i+r}{(v)}}\ ,
\end{aligned}
\end{equation}
 
\begin{equation}
\begin{aligned}
\label{eq:isgmr-back-pairwise}
\nabla \theta_{i-r,i}(\mu, \lambda) = \left. \nabla \mesnew{r,k+1}{i}{\lambda} \right
\vert_{\mu = p^r_{k,i}{(\lambda)}}\ .
\end{aligned}
\end{equation}
Importantly, within an iteration in ISGMR, $\nabla \mathbf{m}^{\mathbf{r},k}$ are updated but
do not affect $\nabla {\mathbf{m}}^{\mathbf{r},k+1}$ until the backpropagation
along all directions $\mathbf{r}$ is executed (line \myblue{18} in
% \myalg{alg:isgmr-back}
Algorithm 3 in Appendix A). This is because within $k$th iteration, independently
updated ${\mathbf{m}}^{r,k+1}$ in $r$ will not affect $\mathbf{m}^{d,k}, \forall d \in
\mathcal{R} \setminus \{r, r^-\}$, until the next iteration (line
\myblue{12} in \myalg{alg:isgmr}).

In contrast, message gradients in TRWP from a direction will affect messages
from other directions since, within an iteration in the forward propagation,
message updates are \textit{direction-by-direction}. For TRWP at $k$th
iteration, $\nabla \theta_{i}{(\lambda)}$ related to \eqref{eq:trwp} is

\begin{equation}
\label{eq:trwp-back-unary}
\begin{aligned}
\nabla &\theta_{i}{(\lambda)}
= \nabla c_{i}{(\lambda)} + \sum_{v \in \mathcal{L}} \sum_{r \in \mathcal{R}}
\sum_{\mu \in \mathcal{L}} \Big( \left. -\nabla \mesnew{r^-}{i}{\mu} \right \vert_{v
= p^{r^-}_{k,i}{(\mu)}} \\
  &+ \left. \sum_{d \in \mathcal{R}} \left.
  \rho_{i+r,i+r+d}
\nabla \mesnew{d}{i+r+d}{\mu} \right \vert_{v = p^d_{k,i+r+d}{(\mu)}} \Big) \right
\vert_{\lambda = p^r_{k,i+r}{(v)}},
\end{aligned}
\end{equation}
where coefficient $\rho_{i+r,i+r+d}$ is for the edge connecting node $i+r$ and
its next one in direction $d$ which is denoted as node $i+r+d$, and the
calculation of $\nabla \theta_{i-r,i}(\lambda, \mu)$ is in the same manner as
\eqref{eq:isgmr-back-pairwise} by replacing $m^{r,k+1}$ with $m^r$.

The backpropagation of TRWP can be derived similarly to ISGMR. 
We must know that gradients of the unary potentials and the pairwise potentials
are accumulated along the opposite direction of the forward scanning direction.
Therefore, an updated message is, in fact, a new variable, and its gradient
should not be accumulated by its previous value but set to 0. This is extremely
important, especially in ISGMR.
It requires the message gradients to be accumulated and assigned in every
iteration (lines \myblue{17}-\myblue{18} in
Algorithm 3 in Appendix A) and be
zero-out (lines \myblue{4} and \myblue{16} in
Algorithm 3
and line
\myblue{14} in
Algorithm 4 in Appendix A).
Meanwhile, gradient derivations of ISGMR and characteristics are provided in
Appendix~\myref{D}.

\begin{figure}[t]
\centering
\begin{subfigure}[b]{0.4\textwidth}
\centering
\includegraphics[width=\textwidth]{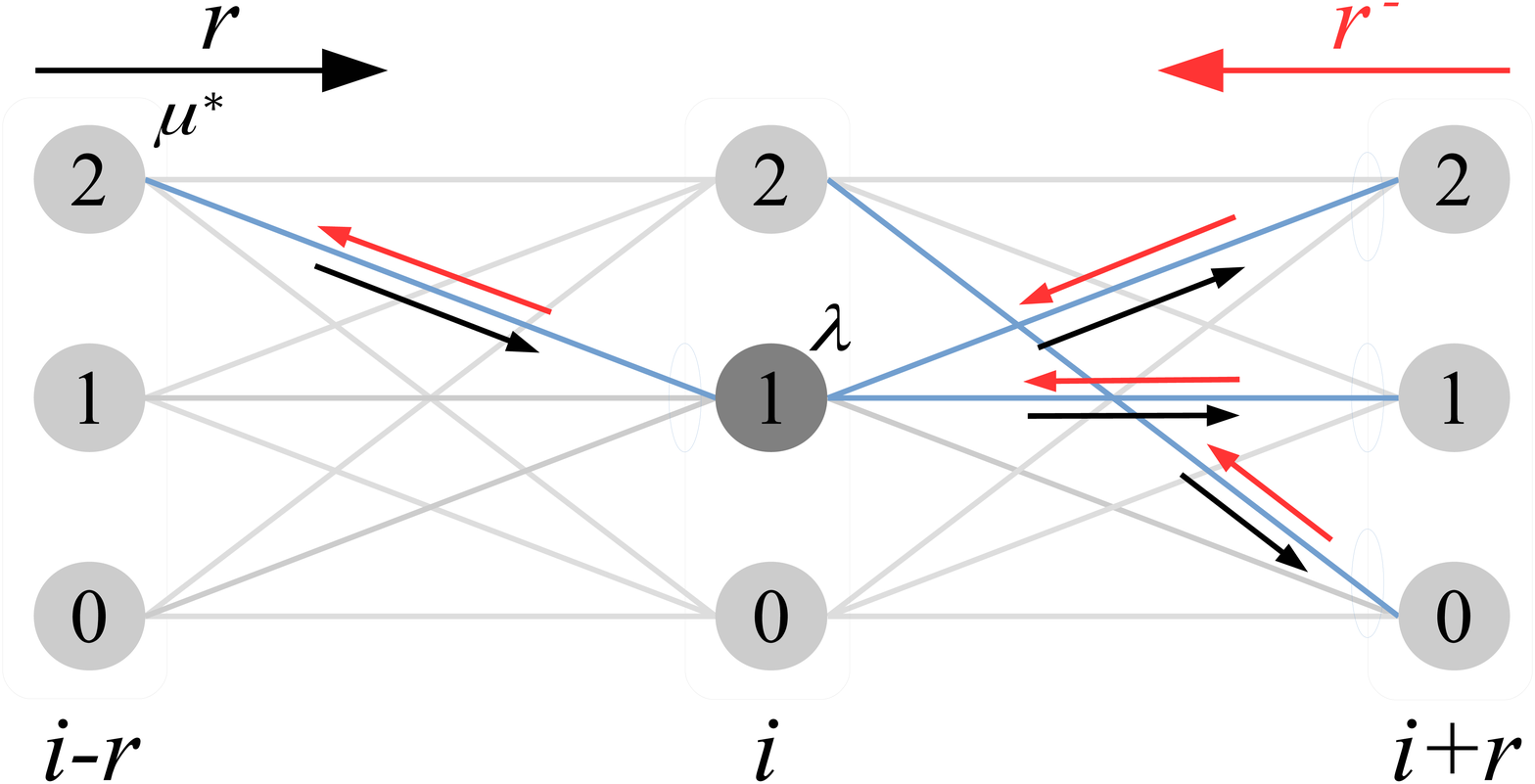}
\caption{message passing}
\label{fig:flow}
\end{subfigure}
\begin{subfigure}[b]{0.4\textwidth}
\centering
\includegraphics[width=0.6\textwidth]{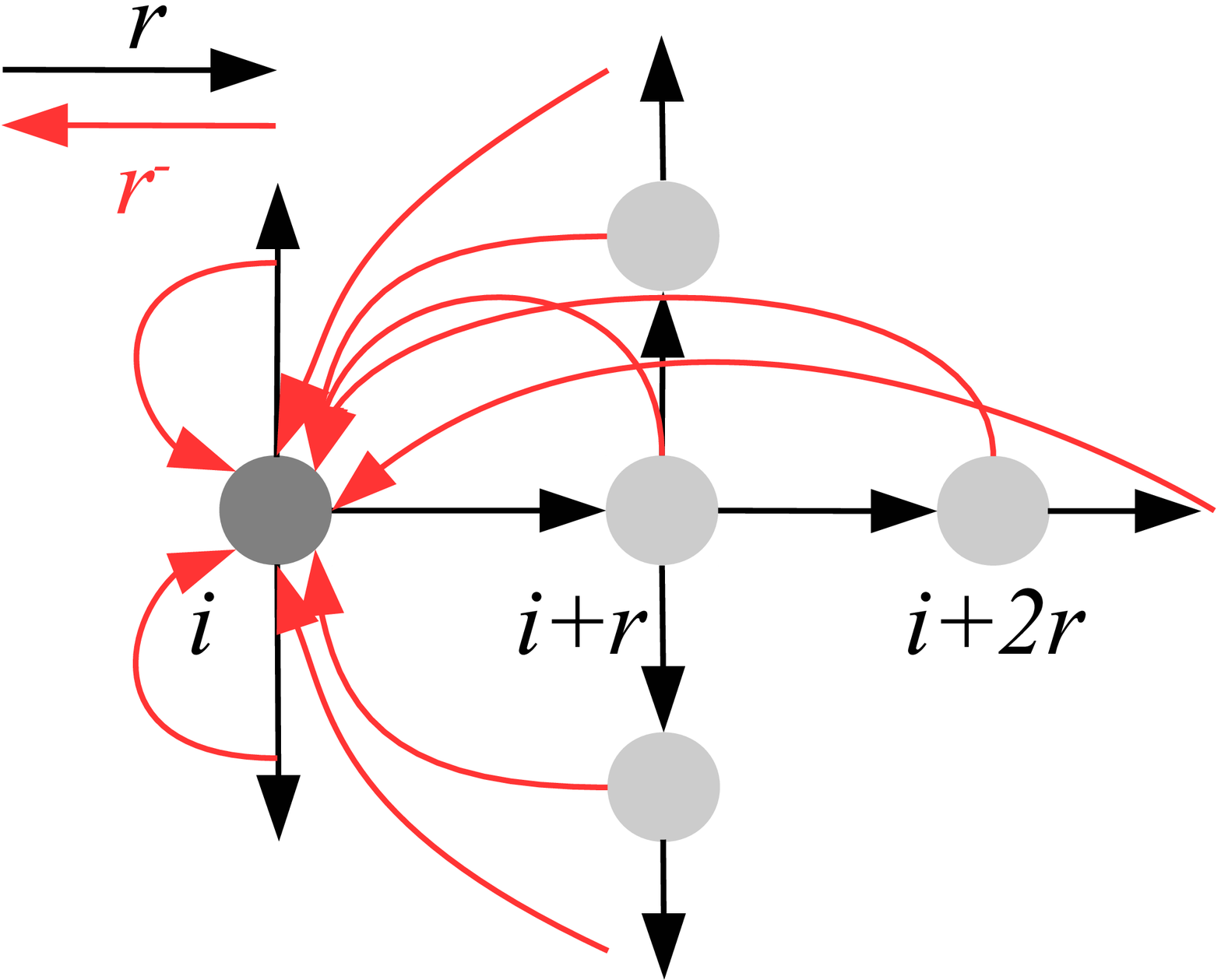}
\caption{gradient accumulation}
\label{fig:accumulation}
\end{subfigure}
\caption{Forward and backward propagation, a target node is in dark gray,
$r$: forward direction, $r^-$: backpropagation direction. (a) blue ellipse: min
operation as MAP, blue line: an edge having the minimum message. (b) a message
gradient at node $i$ accumulated from nodes in $r^{-}$.}
\end{figure}

%==========================================================================
\section{Experiments}
Below, we evaluated the optimization capability of message passing algorithms on
stereo vision and image denoising with fixed yet qualified data terms from
benchmark settings.
In addition, differentiability was evaluated by end-to-end learning for 21-class semantic segmentation.
The experiments include effectiveness and efficiency studies of the message
passing algorithms.
Additional experiments are in Appendix~\myblue{F}.

We implemented SGM, ISGMR, TRWP in C++ with single and multiple threads,
PyTorch, and CUDA from scratch. PyTorch versions are for time comparison and
gradient checking.
For a fair comparison, we adopted benchmark code of TRWS from
\cite{mrf-benchmark} with general pairwise functions; MF followed
Eq.~\myblue{(4)} in \cite{densecrf}.
For iterative SGM, unary potentials were reparametrized by
\eqref{eq:isgmr-aggregation-old}.
OpenGM \cite{openGM} can be used for more comparisons in optimization noting TRWS as one of the most effective inference methods.

Our experiments were on 3.6GHz i7-7700 Intel(R) Core(TM) and Tesla P100 SXM2.

%--------------------------------------------------------------------------
\subsection{Optimization for Stereo Vision and Image Denoising}
\label{sec:energy-min}
The capability of minimizing an energy function determines the significance of
selected algorithms.
We compared our ISGMR and TRWP with MF, SGM with single and multiple iterations, and TRWS.
The evaluated energies are calculated with 4 connections.

\textbf{Datasets.} For stereo vision, we used Tsukuba, Teddy, Venus, Map,
and Cones from Middlebury \cite{middlebury-1,middlebury-2}, 000041\_10 and
000119\_10 from KITTI2015 \cite{kitti2015-1,kitti2015-2}, and delivery\_area\_1l
and facade\_1 from ETH3D two-view \cite{eth3d} for different types of stereo views.
For image denoising, Penguin and House from Middlebury dataset
\footnote{\url{http://vision.middlebury.edu/MRF/results}} were used.

\textbf{MRF model parameters.} Model parameters include unary and pairwise
potentials. In practice, the pairwise potentials consist of a pairwise function
and edge weights, as $\theta_{i,j}(\lambda,\mu) = \theta_{i,j}V(\lambda, \mu)$.
For the pairwise function $V(\cdot,\cdot)$, one can adopt
(truncated) linear, (truncated) quadratic, Cauchy, Huber, \etc,
\cite{multiview}. For the edge weights $\theta_{i,j}$, some methods apply a
higher penalty on edge gradients under a given threshold. We set it as a
constant for the comparison with SGM. Moreover, we adopted edge weights in
\cite{mrf-benchmark} and pairwise functions for Tsukuba, Teddy, and Venus, and
\cite{irgc} for Cones and Map; for the others, the pairwise function was linear
and edges weights were 10.
More evaluations with constant edge weights are given in Appendix~\myref{F}.

\textbf{Number of directions matters.} In \myfig{fig:energy-curve}, ISGMR-8
and TRWP-4 outperform the others in ISGMR-related and TRWP-related methods in
most cases.
From the experiments, 4 directions are sufficient for TRWP, but for ISGMR
energies with 8 directions are lower than those with 4 directions. This is
because messages from 4 directions in ISGMR are insufficient to gather local
information due to independent message updates in each direction. In contrast,
messages from 4 directions in TRWP are highly updated in each direction and
affected by those from the other directions. Note that in
\eqref{eq:isgmr-aggregation} messages from all directions are summed equally,
this makes the labels by TRWP over-smooth within the connection area, for example,
the camera is oversmooth in \myfig{fig:tsukuba-label-TRWP-8}.
Overall, TRWP-4 and ISGMR-8 are the best.

\textbf{ISGMR vs SGM.} \cite{mgm} demonstrates the decrease in energy
of
the over-count corrected SGM compared with the standard SGM. The result shows the
improved optimization results achieved by subtracting unary potentials $(|\mathcal{R}|-1)$ times.
For experimental completion, we show both the decreased energies and improved disparity
maps produced by ISGMR. From Tables~\myblue{\ref{tb:middlebury}-\ref{tb:denoise}},
SGM-related energies are much higher than ISGMR's because of
the over-counted unary potentials. Moreover, ISGMR at the 50th iteration has much
a lower energy value than the 1st iteration, indicating the importance of iterations, and is also much lower than those for MF and SGM at the 50th iteration.

\textbf{TRWP vs TRWS.} TRWP and TRWS have the same manner of updating messages
and could have similar minimum energies. Generally, TRWS has the lowest energy;
at the 50th iteration, however, TRWP-4 has lower energies, for instance, Tsukuba and
Teddy in \mytb{tb:middlebury} and Penguin and House in \mytb{tb:denoise}. For
TRWP, 50 iterations are sufficient to show its high optimization capability, as shown in \myfig{fig:energy-curve}.
More visualizations of Penguin and House denoising are in
Appendix~\myblue{F}.

\begin{figure}[t]
\centering
\begin{subfigure}[b]{0.245\textwidth}
\includegraphics[width=\textwidth]{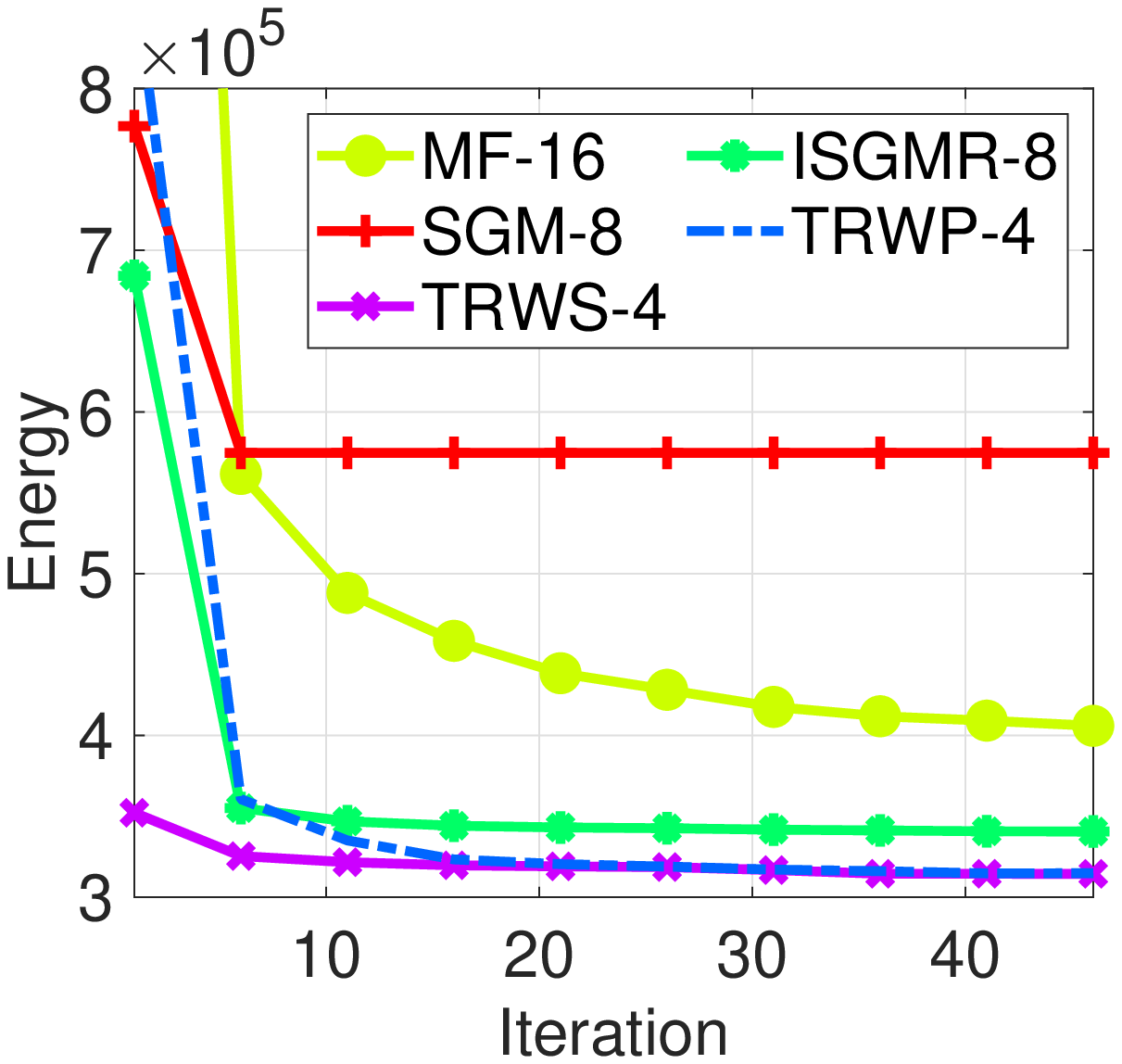}
\caption{Tsukuba}
\end{subfigure}
% ~
\begin{subfigure}[b]{0.245\textwidth}
\includegraphics[width=\textwidth]{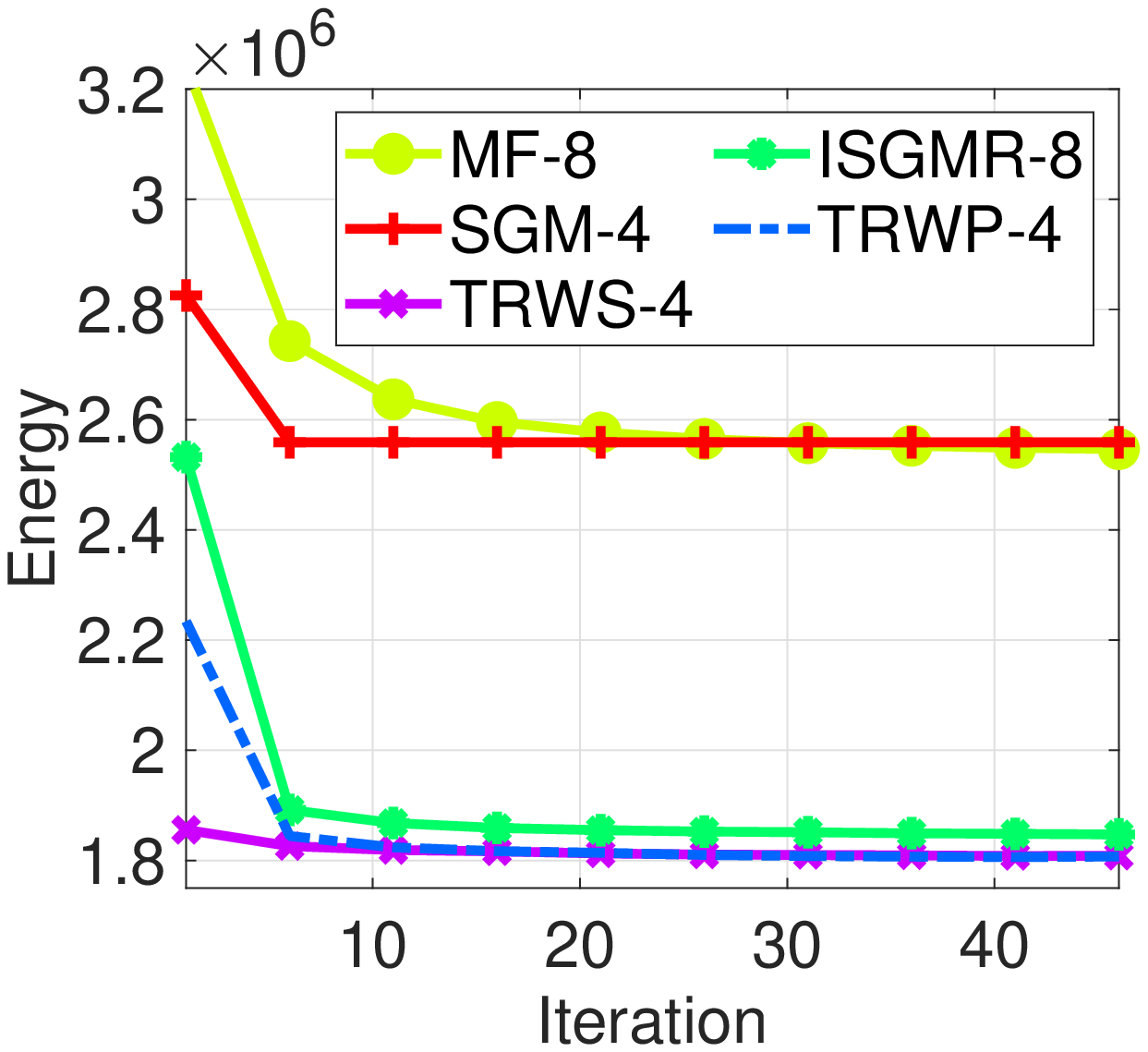}
\caption{Teddy}
\end{subfigure}
% ~
\begin{subfigure}[b]{0.245\textwidth}
\includegraphics[width=\textwidth]{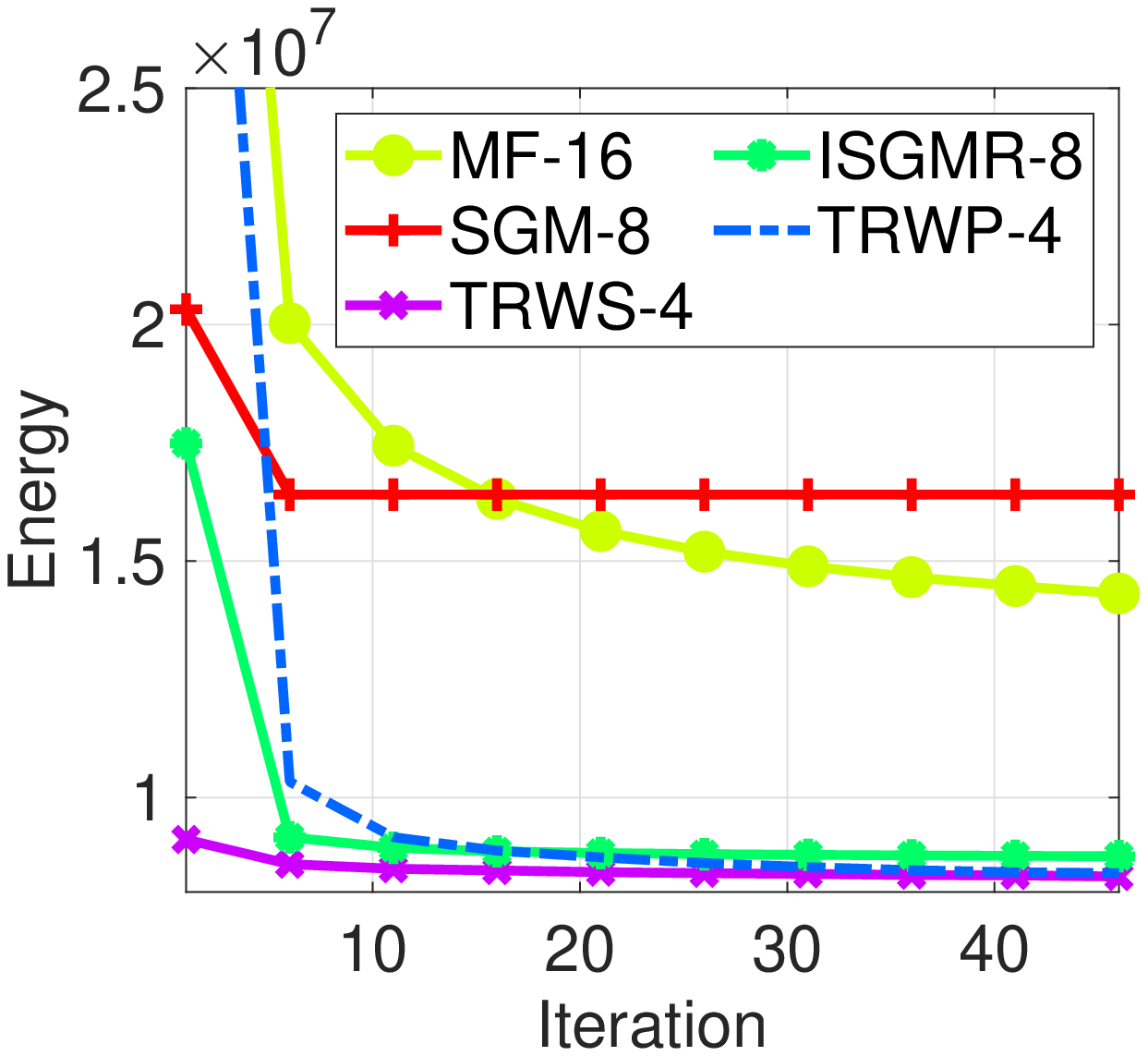}
\caption{000002\_11}
\end{subfigure}
% ~
\begin{subfigure}[b]{0.245\textwidth}
\includegraphics[width=\textwidth]{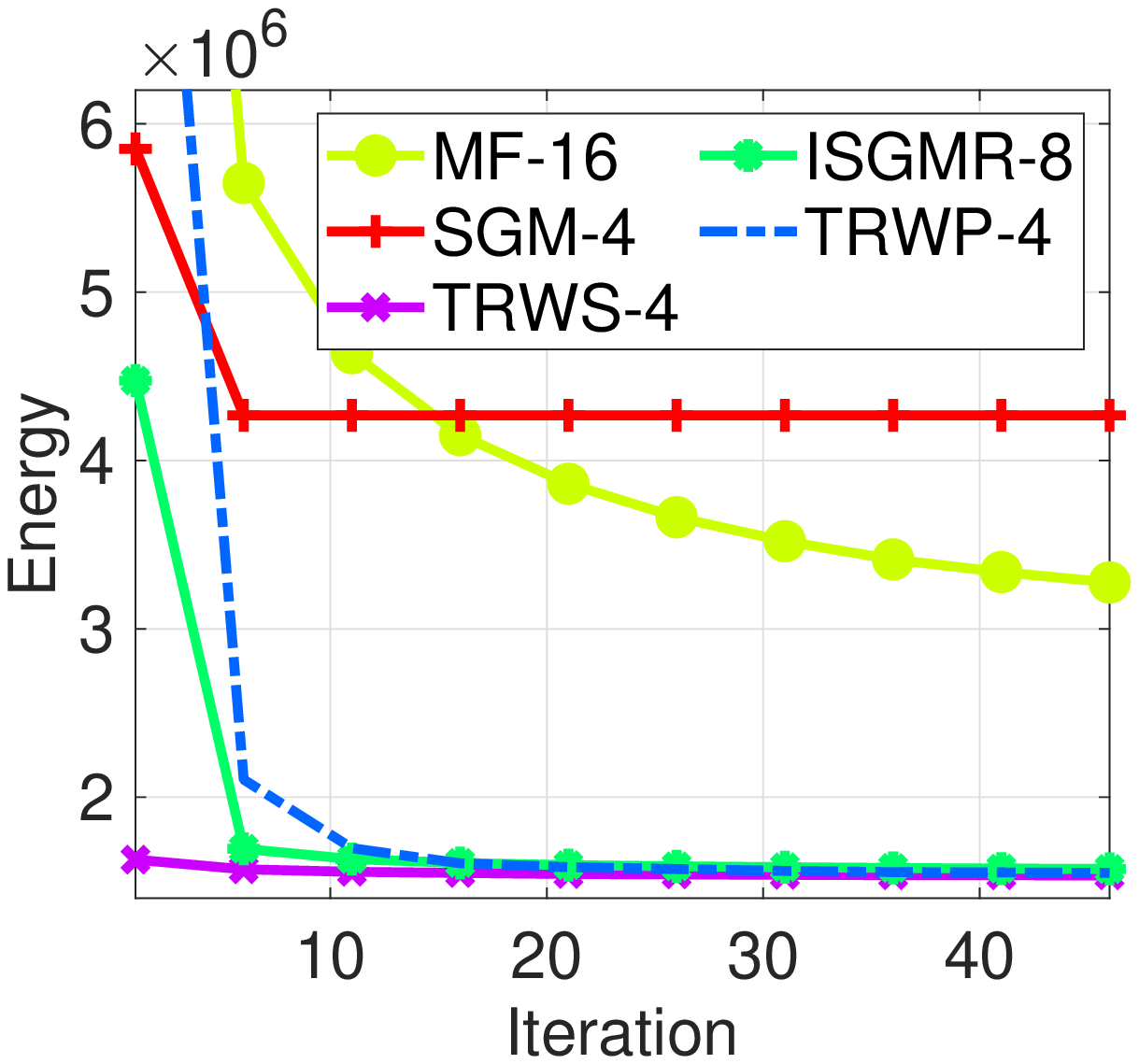}
\caption{delivery\_area\_1l}
\end{subfigure}
\caption{Convergence with the connections having the minimum energy in Table~\ref{tb:middlebury}.}
\label{fig:energy-curve}
\end{figure}

\begin{table*}[t]
\centering
\caption{Energy minimization for stereo vision.
ISGMR is better than SGM and TRWP obtains similar energies as TRWS.
ISGMR and TRWP outperform MF and SGM.}
\resizebox{0.95\textwidth}{!}{
\begin{tabular}{l||r|r|r|r|r|r|r|r}
  \hline
  \multicolumn{1}{c||}{\multirow{2}{*}{\textbf{Method}}} & \multicolumn{2}{c|}{\textbf{Tsukuba}} & \multicolumn{2}{c|}{\textbf{Teddy}} & \multicolumn{2}{c|}{\textbf{000002\_11}} & \multicolumn{2}{c}{\textbf{delivery\_area\_1l}} \\
  \cline{2-9}
   & \multicolumn{1}{c|}{\textbf{1 iter}}
   & \multicolumn{1}{c|}{\textbf{50 iter}}
   & \multicolumn{1}{c|}{\textbf{1 iter}}
   & \multicolumn{1}{c|}{\textbf{50 iter}}
   & \multicolumn{1}{c|}{\textbf{1 iter}}
   & \multicolumn{1}{c|}{\textbf{50 iter}}
  & \multicolumn{1}{c|}{\textbf{1 iter}}
  & \multicolumn{1}{c}{\textbf{50 iter}} \\
  \hline \hline
  MF-4 & 3121704 & 1620524 & 3206347 & 2583784 & 82523536 & 44410056 & 19945352 & 9013862 \\
  SGM-4 & 873777 & 644840 & 2825535 & 2559016 & 24343250 & 18060026 & 5851489 & 4267990 \\
  TRWS-4 & 352178 & \underline{314393} & 1855625 & \underline{1807423} & 9109976 & \lgray 8322635 & 1628879 & \lgray 1534961 \\
  ISGMR-4 (ours) & 824694 & 637996 & 2626648 & 1898641 & 22259606 & 12659612 & 5282024 & 2212106 \\
  TRWP-4 (ours) & 869363 & \lgray 314037 & 2234163 & \lgray 1806990 & 40473776 & \underline{8385450} & 9899787 & \underline{1546795} \\
  \hline
  MF-8 & 2322139 & 504815 & 3244710 & 2545226 & 61157072 & 18416536 & 16581587 & 4510834 \\
  SGM-8 & 776706 & 574758 & 2868131 & 2728682 & 20324684 & 16406781 & 5396353 & 4428411 \\
  ISGMR-8 (ours) & 684185 & \lgray 340347 & 2532071 & \lgray 1847833 & 17489158 & \lgray 8753990 & 4474404 & \lgray 1571528 \\
  TRWP-8 (ours) & 496727 & \underline{348447} & 1981582 & \underline{1849287} & 18424062 & \underline{8860552} & 4443931 & \underline{1587917} \\
  \hline
  MF-16 & 1979155 & 404404 & 3315900 & 2622047 & 46614232 & 14192750 & 13223338 & 3229021 \\
  SGM-16 & 710727 & 587376 & 2907051 & 2846133 & 18893122 & 16791762 & 5092094 & 4611821 \\
  ISGMR-16 (ours) & 591554 & \lgray 377427 & 2453592 & \lgray 1956343 & 15455787 & \lgray 9556611 & 3689863 & \lgray 1594877 \\
  TRWP-16 (ours) & 402033 & \underline{396036} & 1935791 & \underline{1976839} & 11239113 & \underline{9736704} & 2261402 & \underline{1630973} \\
  \hline
\end{tabular}}
\label{tb:middlebury}
\end{table*}

\begin{figure}[ht!]
\centering
\begin{subfigure}[b]{0.18\textwidth}
\includegraphics[width=\textwidth]{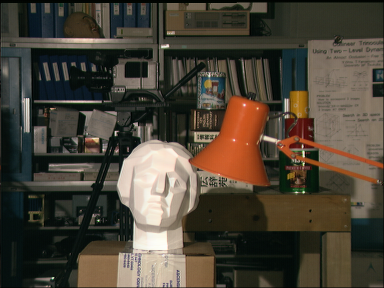}
\caption{left image}
\end{subfigure}
% ~
\begin{subfigure}[b]{0.18\textwidth}
\includegraphics[width=\textwidth]{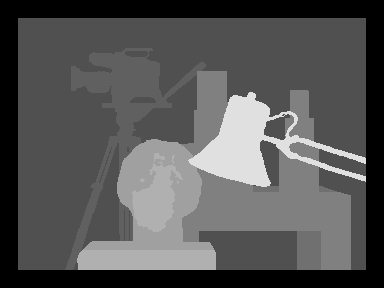}
\caption{GT}
\end{subfigure}
% ~
\begin{subfigure}[b]{0.18\textwidth}
\includegraphics[width=\textwidth]{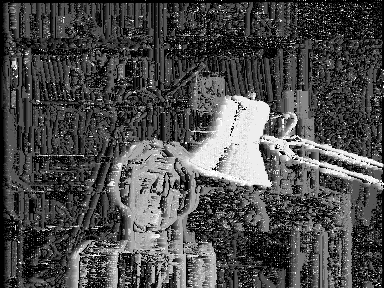}
\caption{unary}
\end{subfigure}
% ~
\begin{subfigure}[b]{0.18\textwidth}
\includegraphics[width=\textwidth]{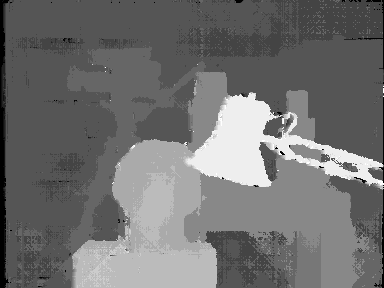}
\caption{1 SGM-8}
\end{subfigure}
% ~
\begin{subfigure}[b]{0.18\textwidth}
\includegraphics[width=\textwidth]{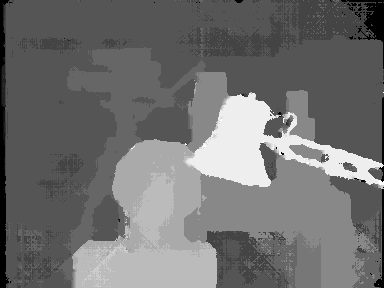}
\caption{1 ISGMR-8}
\end{subfigure}
\\% ~
\begin{subfigure}[b]{0.18\textwidth}
\includegraphics[width=\textwidth]{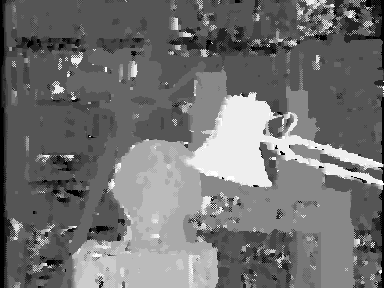}
\caption{MF-4}
\end{subfigure}
% ~
\begin{subfigure}[b]{0.18\textwidth}
\includegraphics[width=\textwidth]{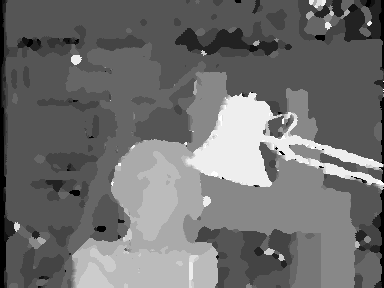}
\caption{MF-8}
\end{subfigure}
% ~
\begin{subfigure}[b]{0.18\textwidth}
\includegraphics[width=\textwidth]{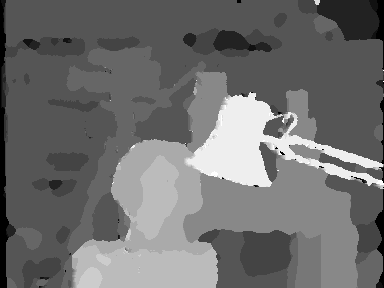}
\caption{MF-16}
\end{subfigure}
% ~
\begin{subfigure}[b]{0.18\textwidth}
\includegraphics[width=\textwidth]{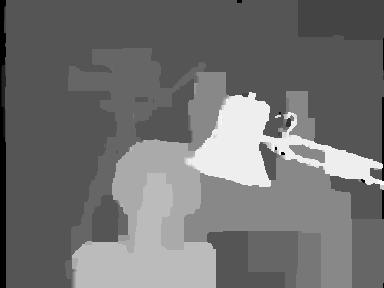}
\caption{TRWS-4}
\end{subfigure}
% ~
\begin{subfigure}[b]{0.18\textwidth}
\includegraphics[width=\textwidth]{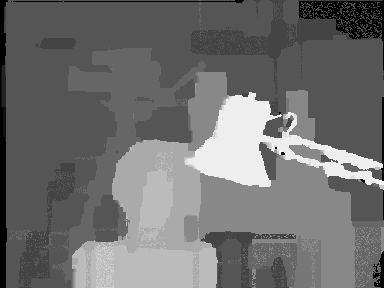}
\caption{ISGMR-4}
\end{subfigure}
\\% ~
\begin{subfigure}[b]{0.18\textwidth}
\includegraphics[width=\textwidth]{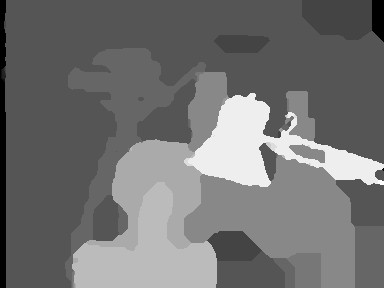}
\caption{ISGMR-8}
\end{subfigure}
% ~
\begin{subfigure}[b]{0.18\textwidth}
\includegraphics[width=\textwidth]{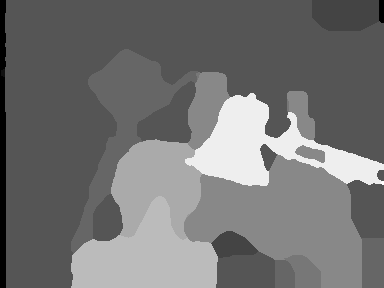}
\caption{ISGMR-16}
\end{subfigure}
% ~
\begin{subfigure}[b]{0.18\textwidth}
\includegraphics[width=\textwidth]{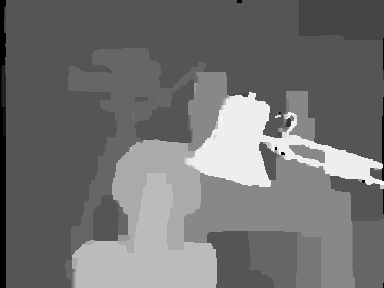}
\caption{TRWP-4}
\end{subfigure}
% ~
\begin{subfigure}[b]{0.18\textwidth}
\includegraphics[width=\textwidth]{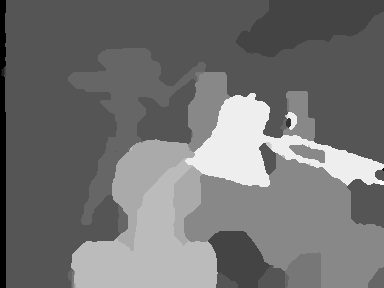}
\caption{TRWP-8}
\label{fig:tsukuba-label-TRWP-8}
\end{subfigure}
% ~
\begin{subfigure}[b]{0.18\textwidth}
\includegraphics[width=\textwidth]{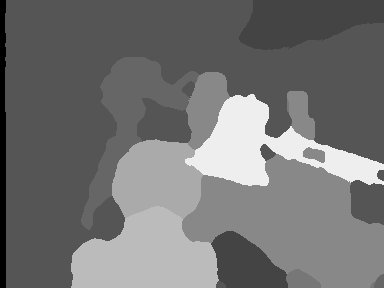}
\caption{TRWP-16}
\end{subfigure}
\caption{Disparities of Tsukuba. (d)-(e) are at 1st iteration. (f)-(o) are at 50th iteration. (j) and (l) have the lowest energies in ISGMR-related and TRWP-related methods respectively. TRWP-4 and TRWS-4 have similar disparities for the most parts.}
\label{fig:tsukuba-label}
\end{figure}

\begin{figure}[ht!]
\centering
\begin{minipage}[]{0.46\textwidth}
\centering
    \captionof{table}{\small{Energy minimization for image denoising at 50th
    iteration with $4$, $8$, $16$ connections
    (all numbers divided by $10^3$). Our ISGMR or TRWP performs best.
    }}
    \begin{tabular}{l||c|c}
      \hline
      \multicolumn{1}{c||}{\multirow{1}{*}{\textbf{Method}}}
      & \multicolumn{1}{c|}{\textbf{Penguin}}
      & \multicolumn{1}{c}{\textbf{House}} \\
      \hline \hline
      MF-4 & 46808 & 50503 \\
      SGM-4 & 31204 & 66324 \\
      TRWS-4 & \underline{15361} & \underline{37572} \\
      ISGMR-4 (ours) & 16514 & 37603 \\
      TRWP-4 (ours) & \lgray 15358 & \lgray 37552 \\
      \hline
      MF-8 & 21956 & 47831 \\
      SGM-8 & 37520 & 76079 \\
      ISGMR-8 (ours) & \lgray 15899 & \lgray 39975 \\
      TRWP-8 (ours) & \underline{16130} & \underline{40209} \\
      \hline
      MF-16 & 20742 & 55513 \\
      SGM-16 & 47028 & 87457 \\
      ISGMR-16 (ours) & \lgray 17035 & \lgray 46997 \\
      TRWP-16 (ours) & \underline{17516} & \underline{47825} \\
      \hline
    \end{tabular}
    \label{tb:denoise}
\end{minipage}
~
\begin{minipage}[]{0.52\textwidth}
\centering
    \begin{subfigure}[b]{0.22\textwidth}
    \includegraphics[width=\textwidth]{denoise/penguin/penguin_input.png}
    \minisubcaption{a}{noisy}
    \end{subfigure}
    \begin{subfigure}[b]{0.22\textwidth}
    \includegraphics[width=\textwidth]{denoise/penguin/penguin_gt.png}
    \minisubcaption{b}{GT}
    \end{subfigure}
    \begin{subfigure}[b]{0.22\textwidth}
    \includegraphics[width=\textwidth]{denoise/penguin/penguin_MeanField_iter_50_TQ_trunc_200_dir_16_rho_1.png}
    \minisubcaption{c}{MF-16}
    \end{subfigure}
    \begin{subfigure}[b]{0.22\textwidth}
    \includegraphics[width=\textwidth]{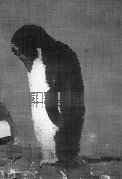}
    \minisubcaption{d}{1 SGM-4}
    \end{subfigure}
    \\
    \begin{subfigure}[b]{0.22\textwidth}
    \includegraphics[width=\textwidth]{denoise/penguin/penguin_SGM_iter_50_TQ_trunc_200_dir_4_rho_1.png}
    \minisubcaption{e}{SGM-4}
    \end{subfigure}
    \begin{subfigure}[b]{0.22\textwidth}
    \includegraphics[width=\textwidth]{denoise/penguin/penguin_TRWS.png}
    \minisubcaption{f}{TRWS-4}
    \end{subfigure}
    \begin{subfigure}[b]{0.22\textwidth}
    \includegraphics[width=\textwidth]{denoise/penguin/penguin_ISGMR_iter_50_TQ_trunc_200_dir_8_rho_1.png}
    \minisubcaption{g}{ISGMR-8}
    \end{subfigure}
    \begin{subfigure}[b]{0.22\textwidth}
    \includegraphics[width=\textwidth]{denoise/penguin/penguin_TRWP_iter_50_TQ_trunc_200_dir_4_rho_0_5.png}
    \minisubcaption{h}{TRWP-4}
    \end{subfigure}
    \captionof{figure}{\small{Penguin denoising
    corresponding to the minimum energies marked with gray color in \mytb{tb:denoise}.
    ISGMR-8 and TRWP-4 are our proposals.
    }}
    \label{fig:denoise}
\end{minipage}
\end{figure}

%--------------------------------------------------------------------------
\subsection{End-to-End Learning for Semantic Segmentation}
Although deep network and multi-scale strategy on CNN make semantic segmentation
smooth and continuous on object regions, effective message passing inference on
pairwise MRFs is beneficial for fine results with auxiliary edge information.
The popular denseCRF \cite{densecrf} demonstrated the effectiveness of
using MF inference and the so-called dense connections; our experiments,
however, illustrated that with local connections, superior inferences, such
as TRWS, ISGMR, and TRWP, have a better convergence ability than MF and SGM to improve
the performance.

Below, we adopted TRWP-4 and ISGMR-8 as our inference methods and
negative logits from
DeepLabV3+ \cite{DeepLabV3+} as unary terms. Edge weights from
Canny edges are in the form of $\theta_{ij} = 1-|e_{i} - e_{j}|$,
where $e_i$ is a binary Canny edge value at node $i$.
Potts model was used for pairwise function $V(\lambda,\mu)$.
Since MF required much larger GPU memory than others due to its dense gradients,
for practical purposes we used MF-4 for learning
with the same batch size 12 within our GPU memory capacity.

\textbf{Datasets.} We used PASCAL VOC 2012 \cite{voc2012} and Berkeley
benchmark \cite{berkeley}, with 1449 samples of the PASCAL VOC 2012 val set for
validation and the other 10582 for training.
These datasets identify 21 classes with 20 objects and 1 background.

\textbf{CNN learning parameters.} We trained the state-of-the-art DeepLabV3+
(ResNet101 as the backbone) with initial learning rate 0.007, ``poly" learning
rate decay scheduler, and image size 512$\times$512.
Negative logits from DeepLabV3+ served as unary terms, the learning rate was
decreased for learning message passing inference with 5 iterations, \ie, 1e-4
for TRWP and SGM and 1e-6 for ISGMR and MF.
Note that we experimented with all of these learning rates for involved
inferences and selected the best for demonstration, for instance, for MF the accuracy by
1e-6 is much higher than the one by 1e-4.

\begin{table}[t]
\centering
\caption{Learning for semantic segmentation with mIoU on PASCAL VOC2012 val
set.}
\label{tb:voc2012_val}
\begin{subtable}{0.45\textwidth}
\centering
\caption{term weight for TRWP-4}
\resizebox{0.6\textwidth}{!}{\begin{tabular}{l||c|c}
  \hline
  \multicolumn{1}{c||}{\multirow{1}{*}{\textbf{Method}}}
  & \multicolumn{1}{c|}{$\boldsymbol \lambda$}
  & \textbf{mIoU (\%)} \\
  \hline \hline
  +TRWP-4 & 1 & 79.27 \\
  \hline
  +TRWP-4 & 10 & 79.53 \\
  \hline
  +TRWP-4 & 20 & \lgray 79.65 \\
  \hline
  +TRWP-4 & 30 & 79.44 \\
  \hline
  +TRWP-4 & 40 & 79.60 \\
  \hline
\end{tabular}}
\end{subtable}
~
\begin{subtable}{0.44\textwidth}
\centering
\caption{full comparison}
\resizebox{0.75\textwidth}{!}{\begin{tabular}{l||c|c}
  \hline
  \multicolumn{1}{c||}{\multirow{1}{*}{\textbf{Method}}}
  & \multicolumn{1}{c|}{$\boldsymbol \lambda$}
  & \textbf{mIoU (\%)} \\
  \hline \hline
  DeepLabV3+ \cite{DeepLabV3+} & - & 78.52 \\
  \hline
  +SGM-8 \cite{sgm} & 5 & 78.94 \\
  \hline
  +MF-4 \cite{densecrf} & 5 & 77.89 \\
  \hline \hline
  +ISGMR-8 (ours) & 5 & 78.95 \\
  \hline
  +TRWP-4 (ours) & 20 & \lgray 79.65 \\
  \hline
\end{tabular}}
\end{subtable}
\end{table}

\begin{figure}[!ht]
\centering
\begin{subfigure}[b]{0.11\textwidth}
\includegraphics[width=\textwidth]{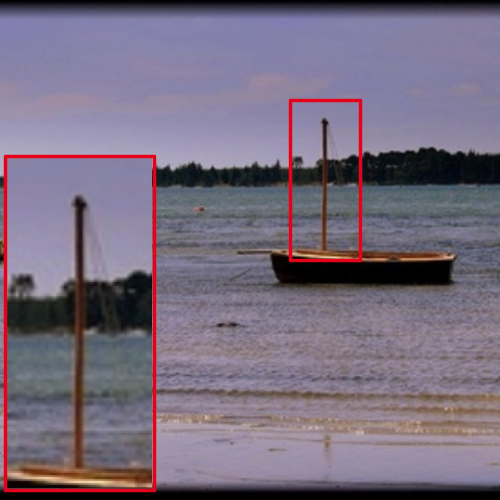}
\end{subfigure}
\begin{subfigure}[b]{0.11\textwidth}
\includegraphics[width=\textwidth]{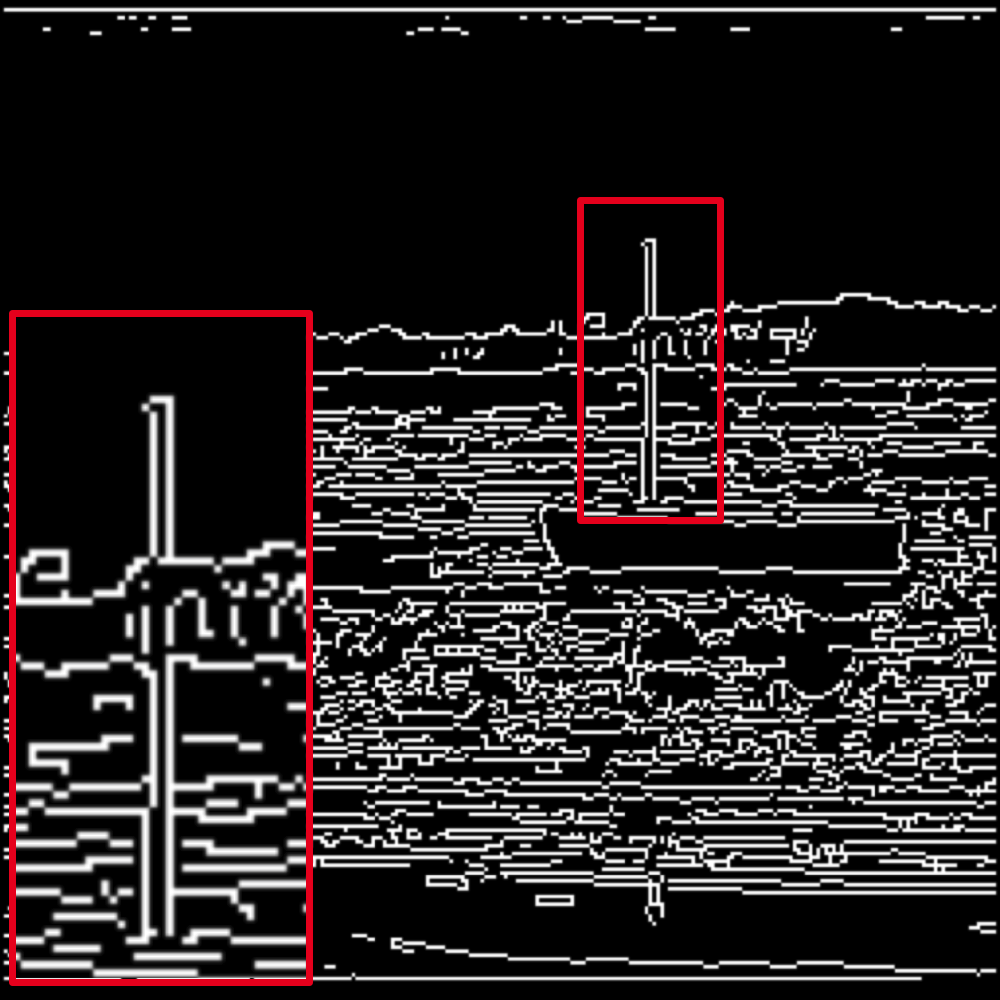}
\end{subfigure}
\begin{subfigure}[b]{0.11\textwidth}
\includegraphics[width=\textwidth]{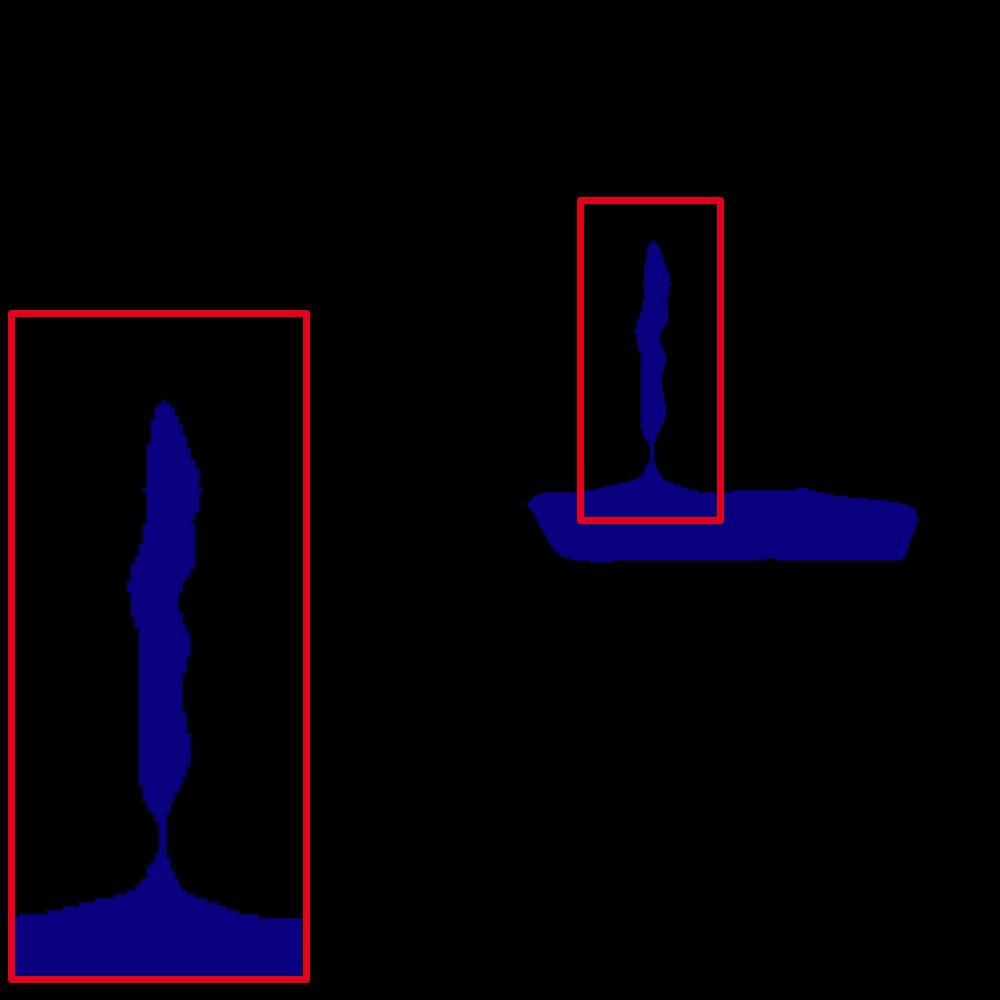}
\end{subfigure}
\begin{subfigure}[b]{0.11\textwidth}
\includegraphics[width=\textwidth]{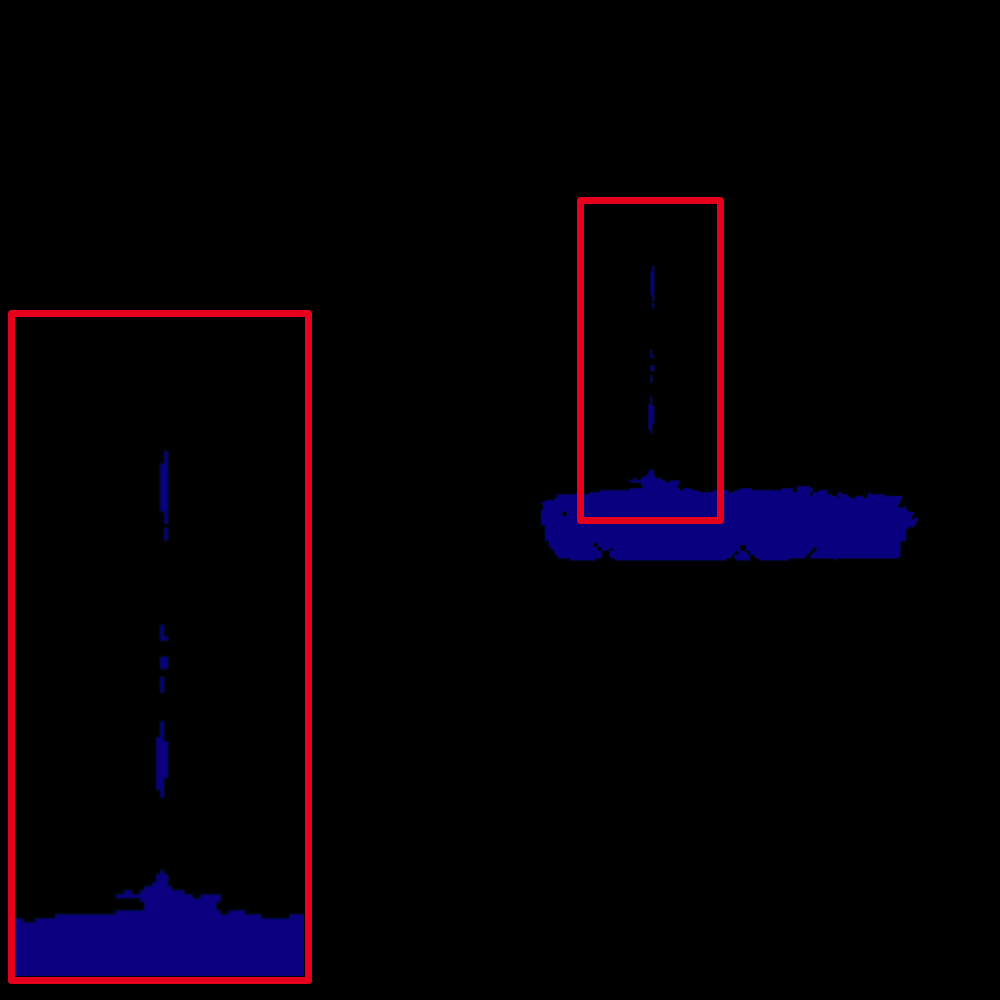}
\end{subfigure}
\begin{subfigure}[b]{0.11\textwidth}
\includegraphics[width=\textwidth]{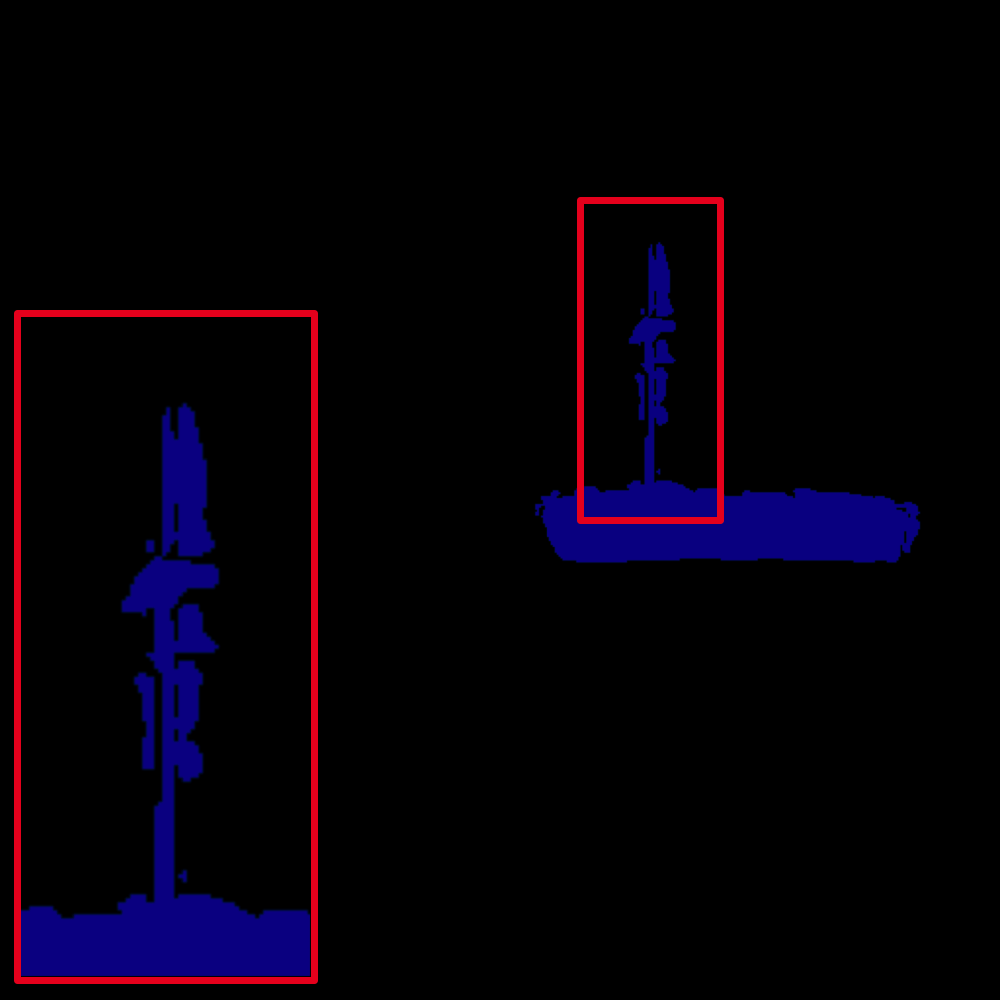}
\end{subfigure}
\begin{subfigure}[b]{0.11\textwidth}
\includegraphics[width=\textwidth]{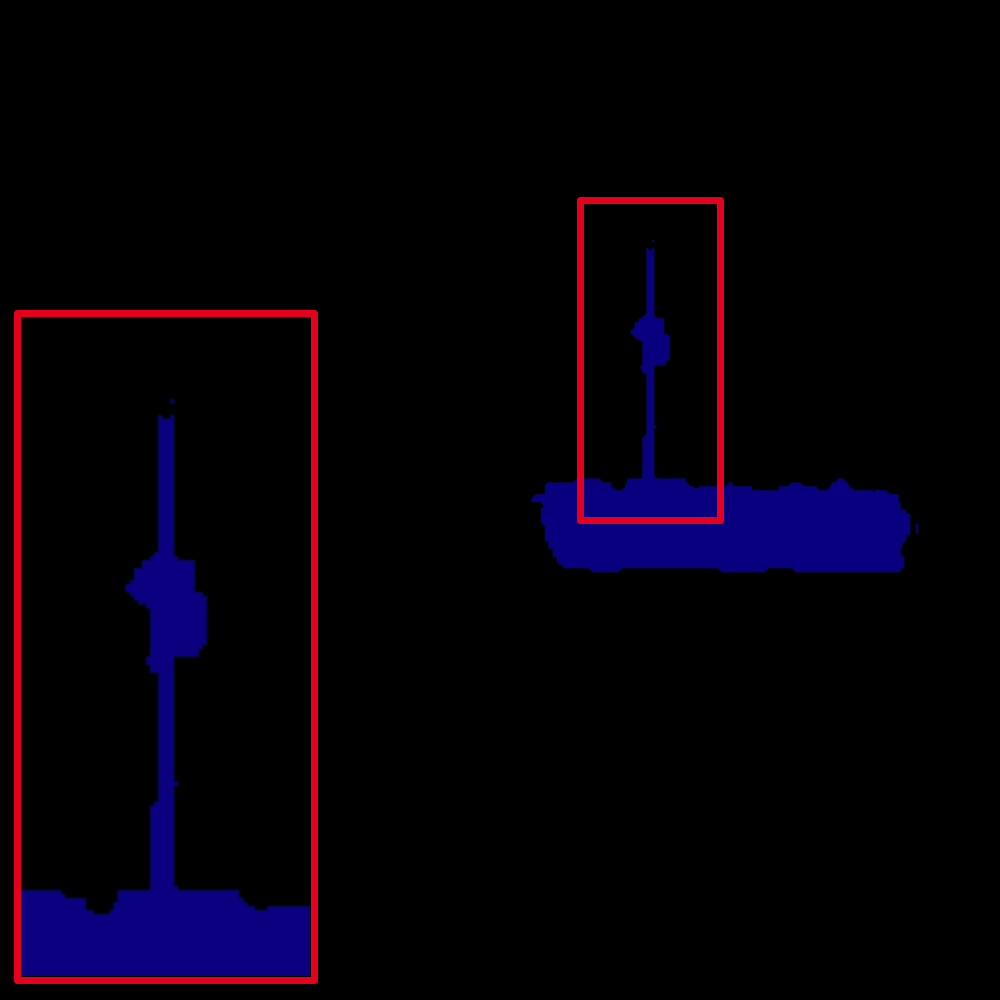}
\end{subfigure}
\begin{subfigure}[b]{0.11\textwidth}
\includegraphics[width=\textwidth]{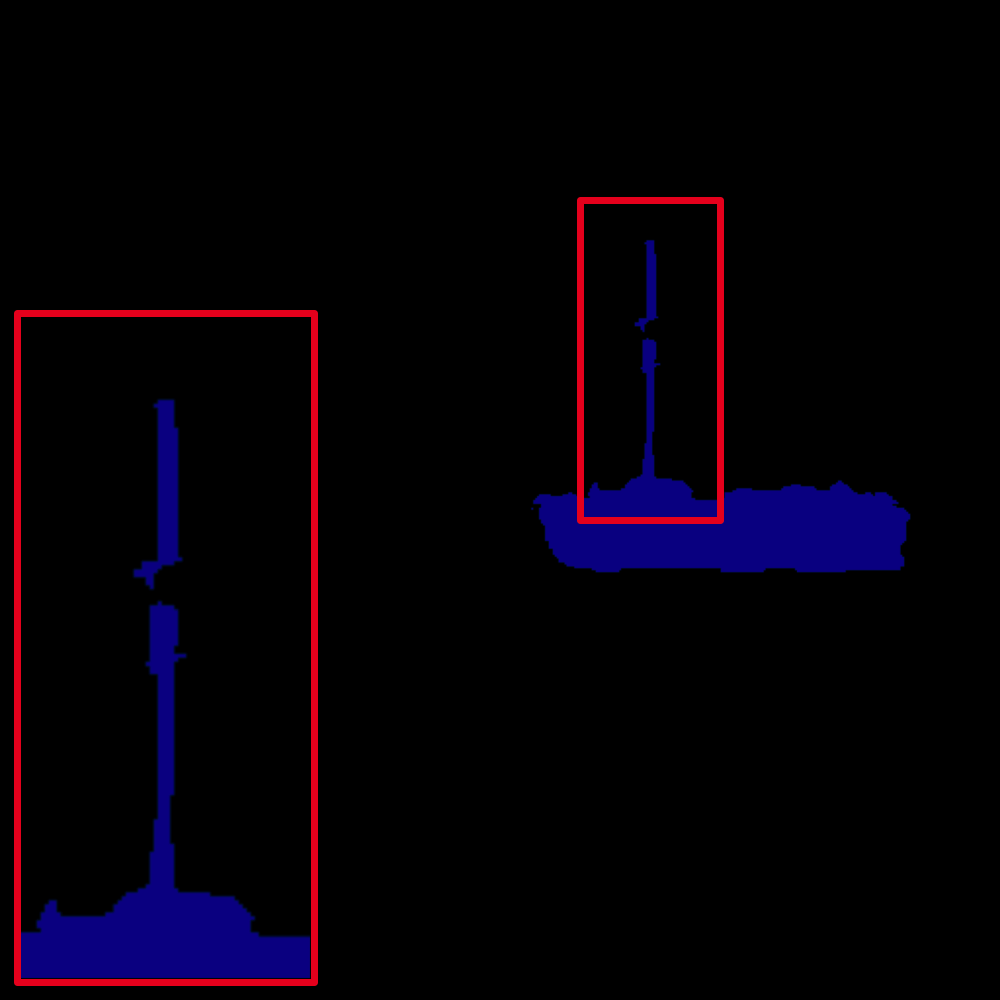}
\end{subfigure}
\begin{subfigure}[b]{0.11\textwidth}
\includegraphics[width=\textwidth]{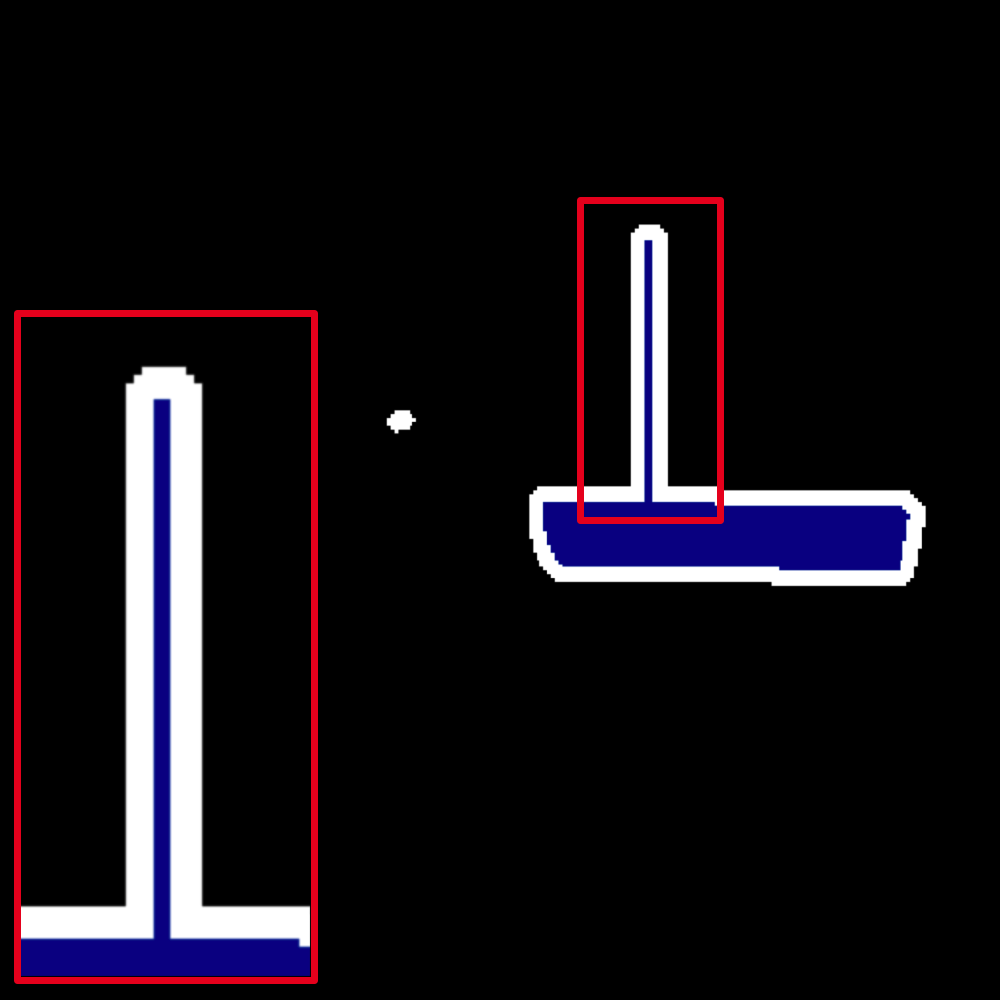}
\end{subfigure}
\\
\begin{subfigure}[b]{0.11\textwidth}
\includegraphics[width=\textwidth]{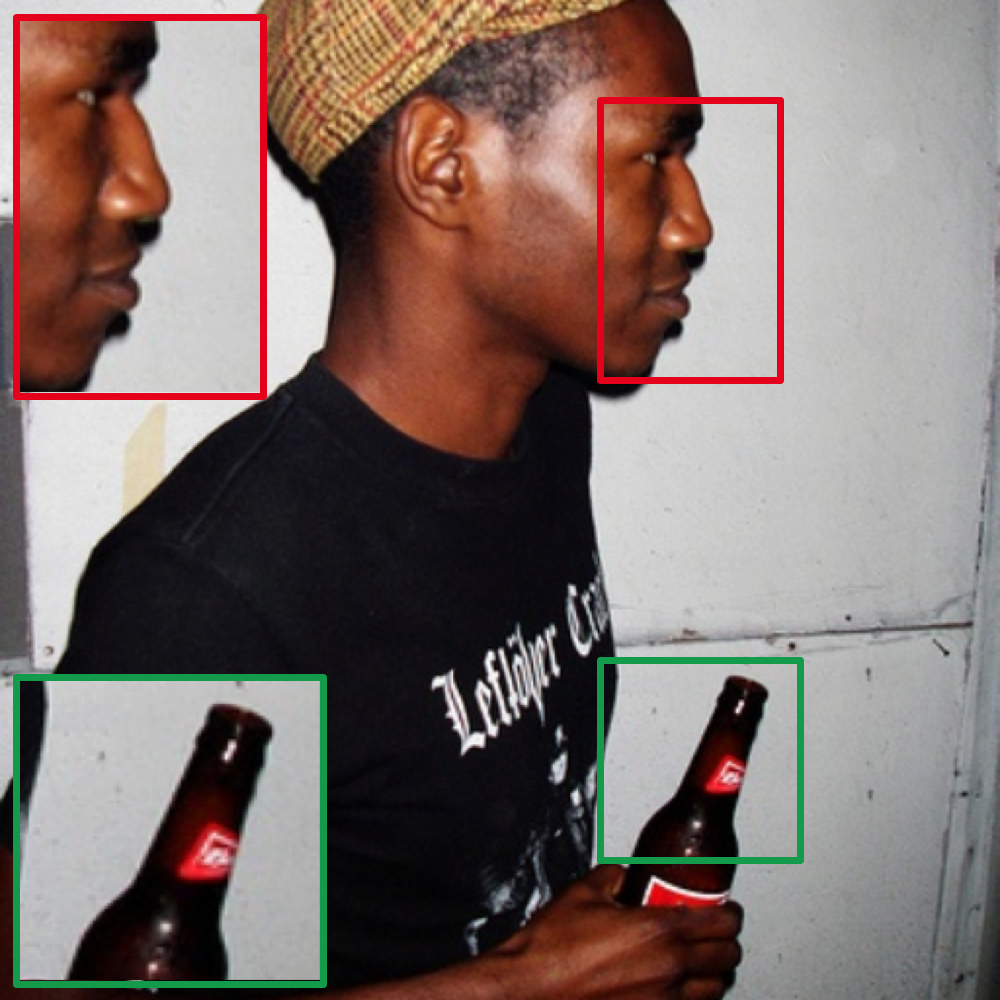}
\end{subfigure}
\begin{subfigure}[b]{0.11\textwidth}
\includegraphics[width=\textwidth]{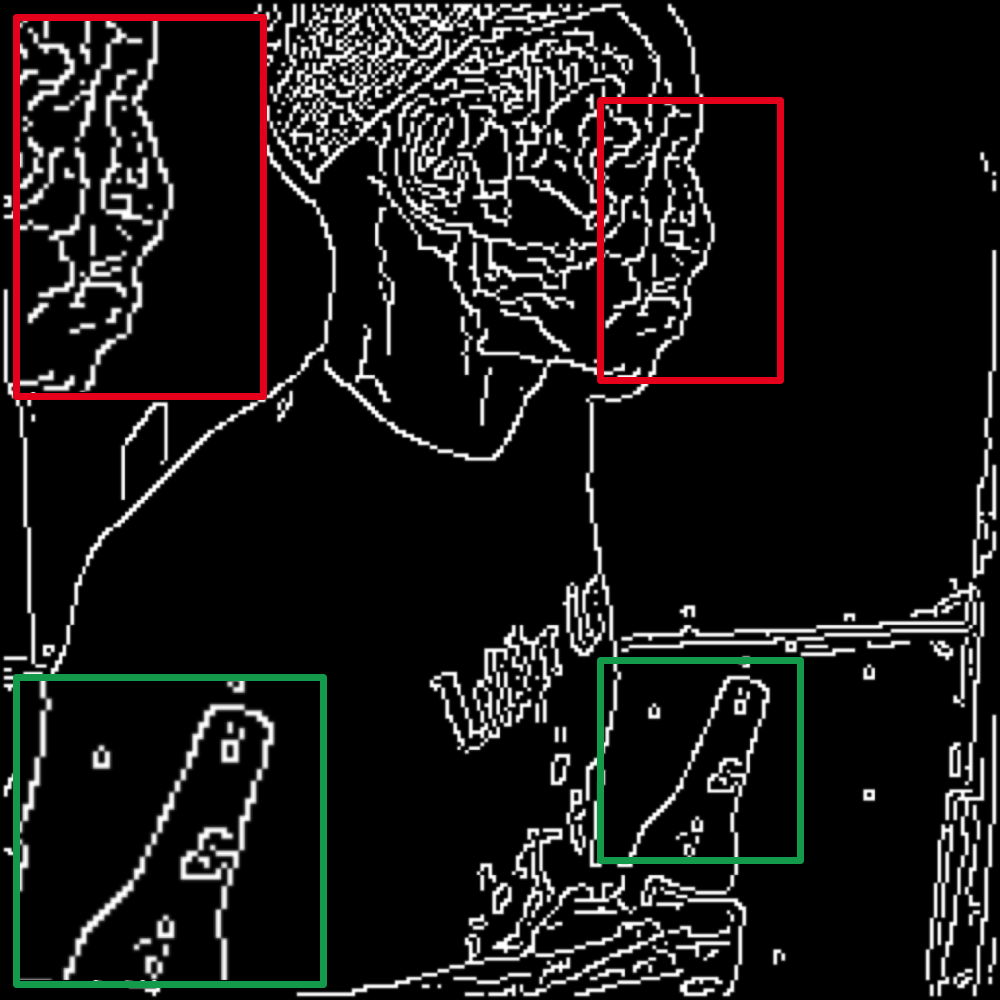}
\end{subfigure}
\begin{subfigure}[b]{0.11\textwidth}
\includegraphics[width=\textwidth]{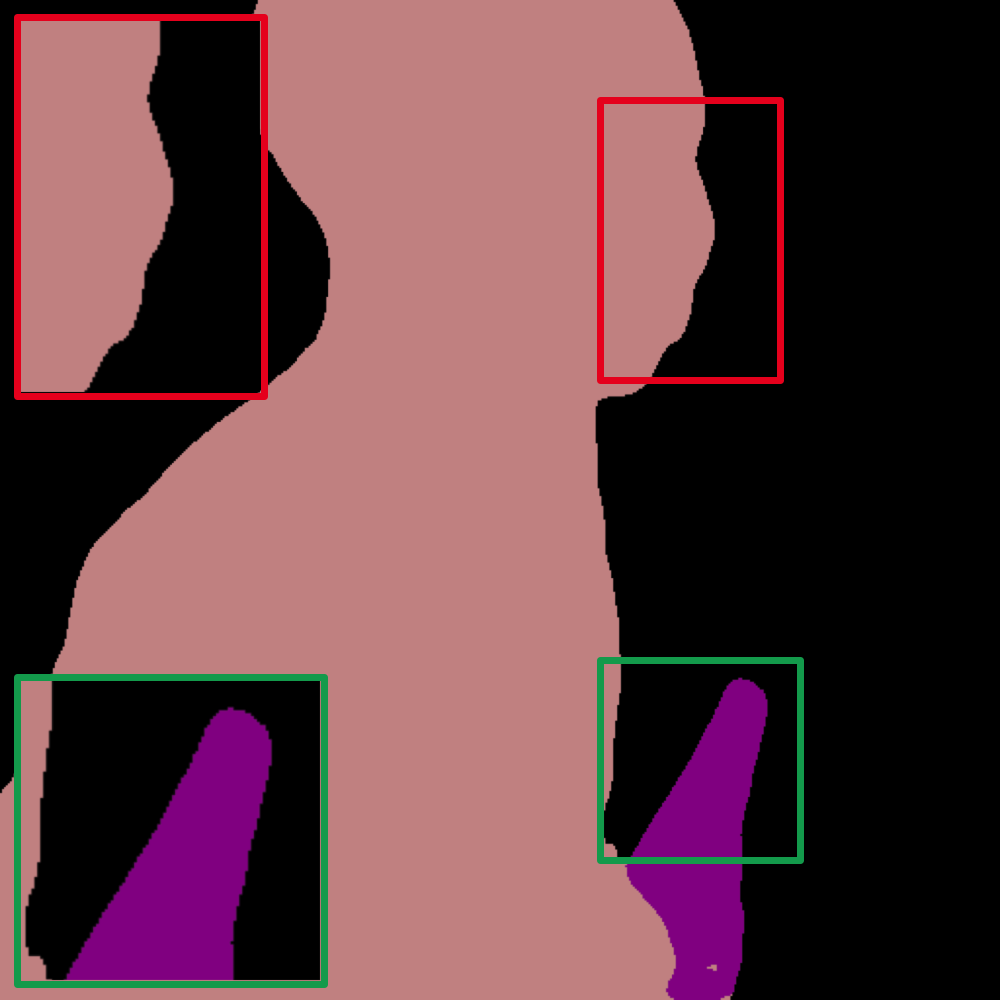}
\end{subfigure}
\begin{subfigure}[b]{0.11\textwidth}
\includegraphics[width=\textwidth]{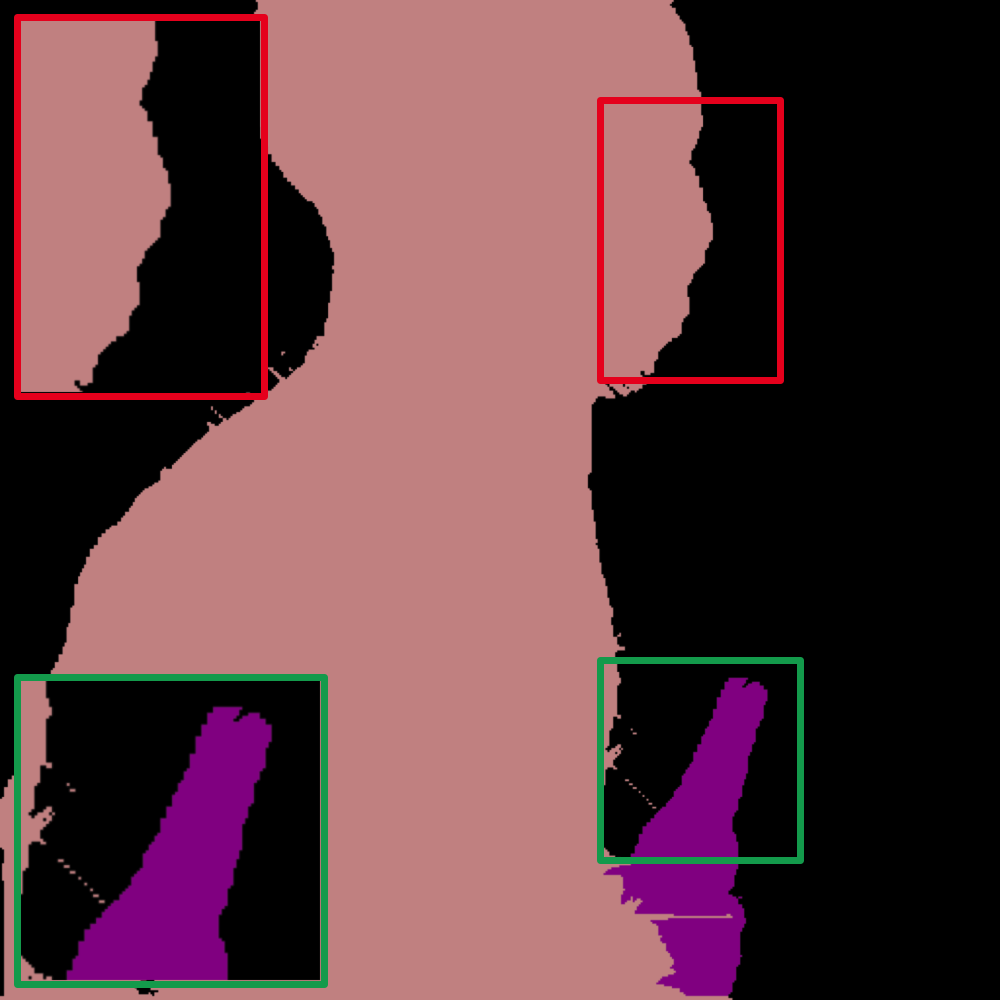}
\end{subfigure}
\begin{subfigure}[b]{0.11\textwidth}
\includegraphics[width=\textwidth]{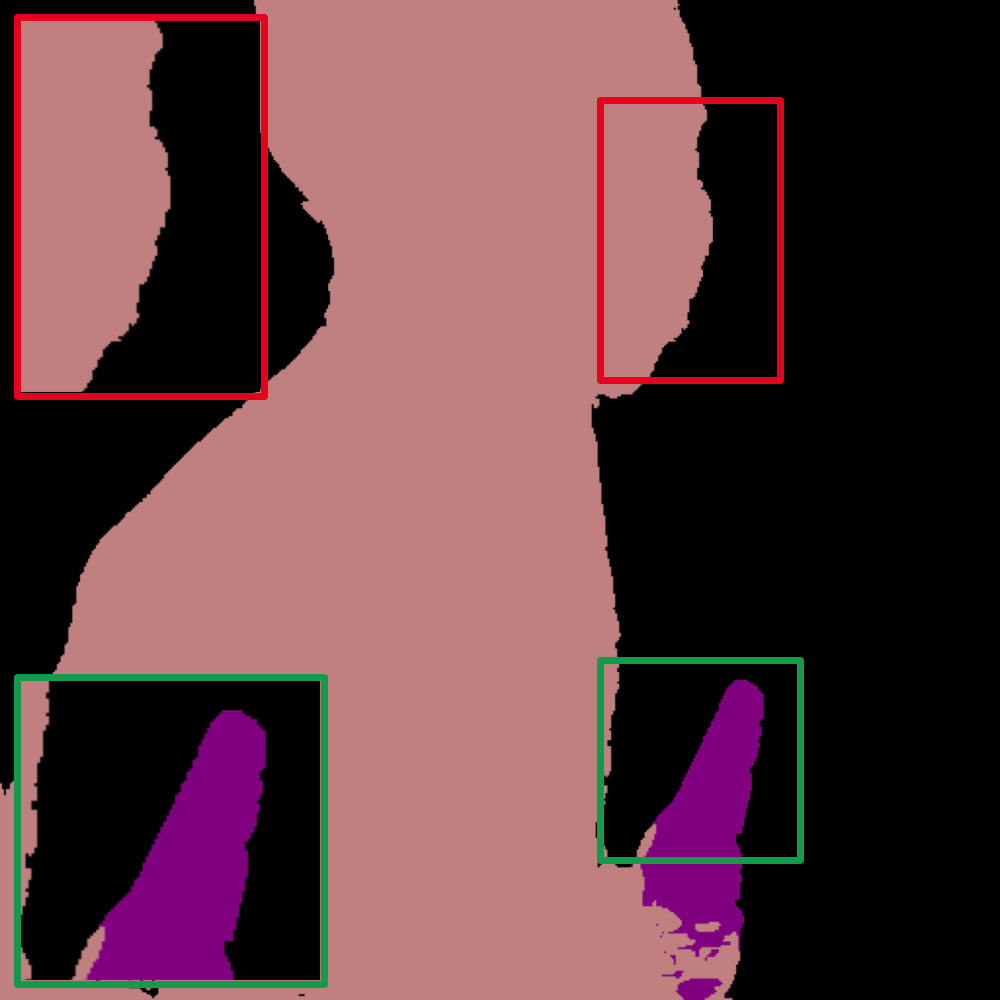}
\end{subfigure}
\begin{subfigure}[b]{0.11\textwidth}
\includegraphics[width=\textwidth]{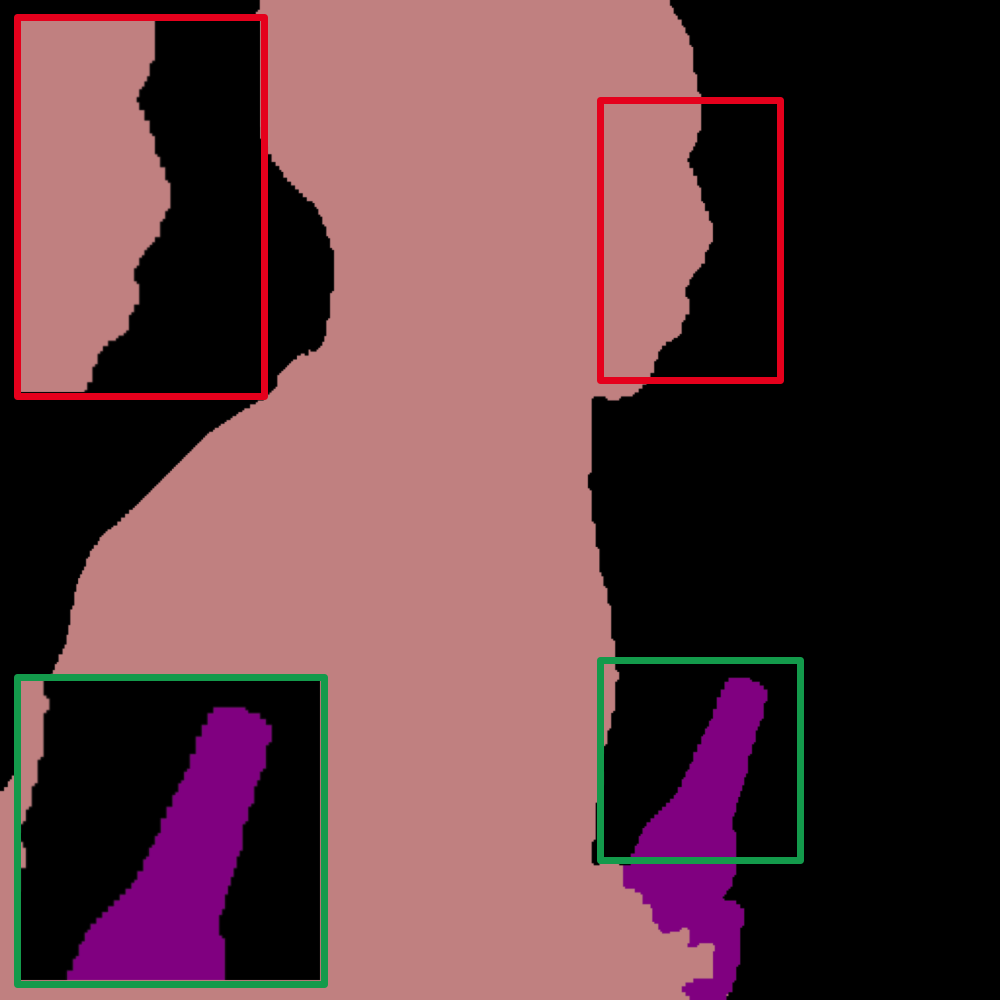}
\end{subfigure}
\begin{subfigure}[b]{0.11\textwidth}
\includegraphics[width=\textwidth]{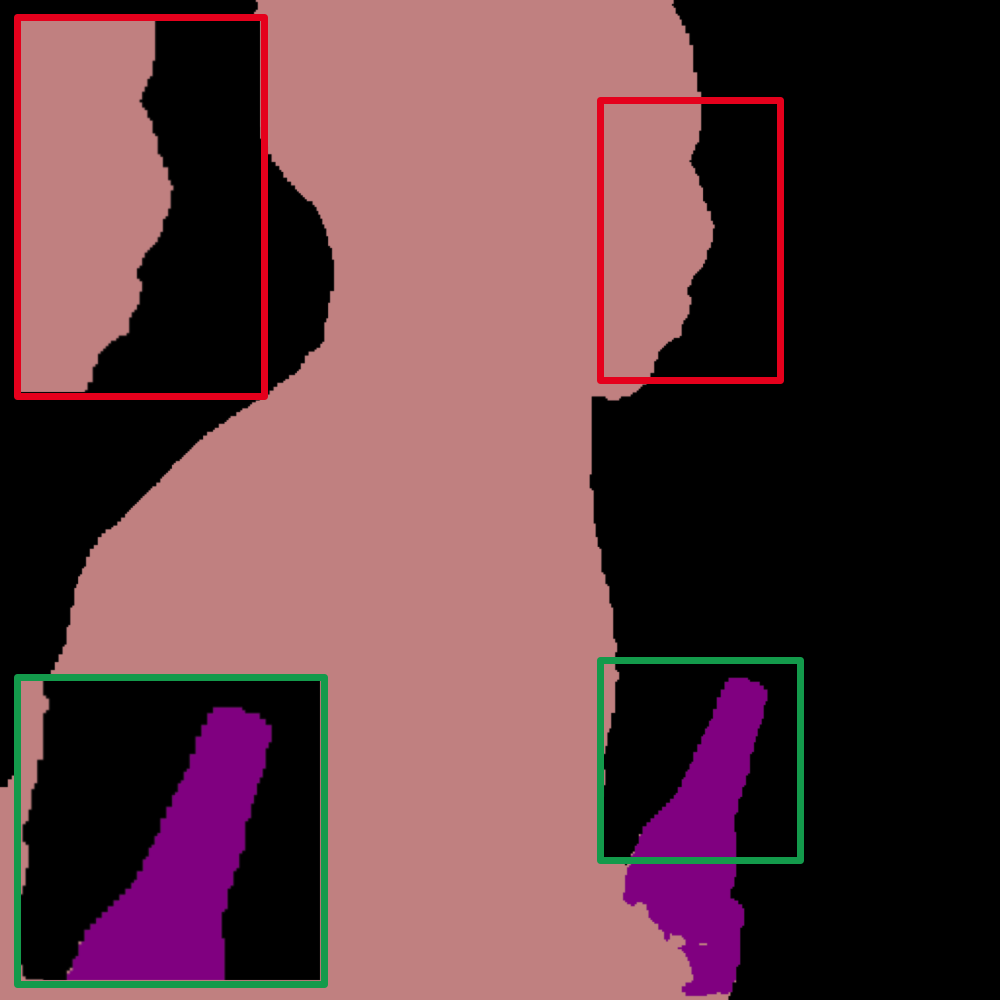}
\end{subfigure}
\begin{subfigure}[b]{0.11\textwidth}
\includegraphics[width=\textwidth]{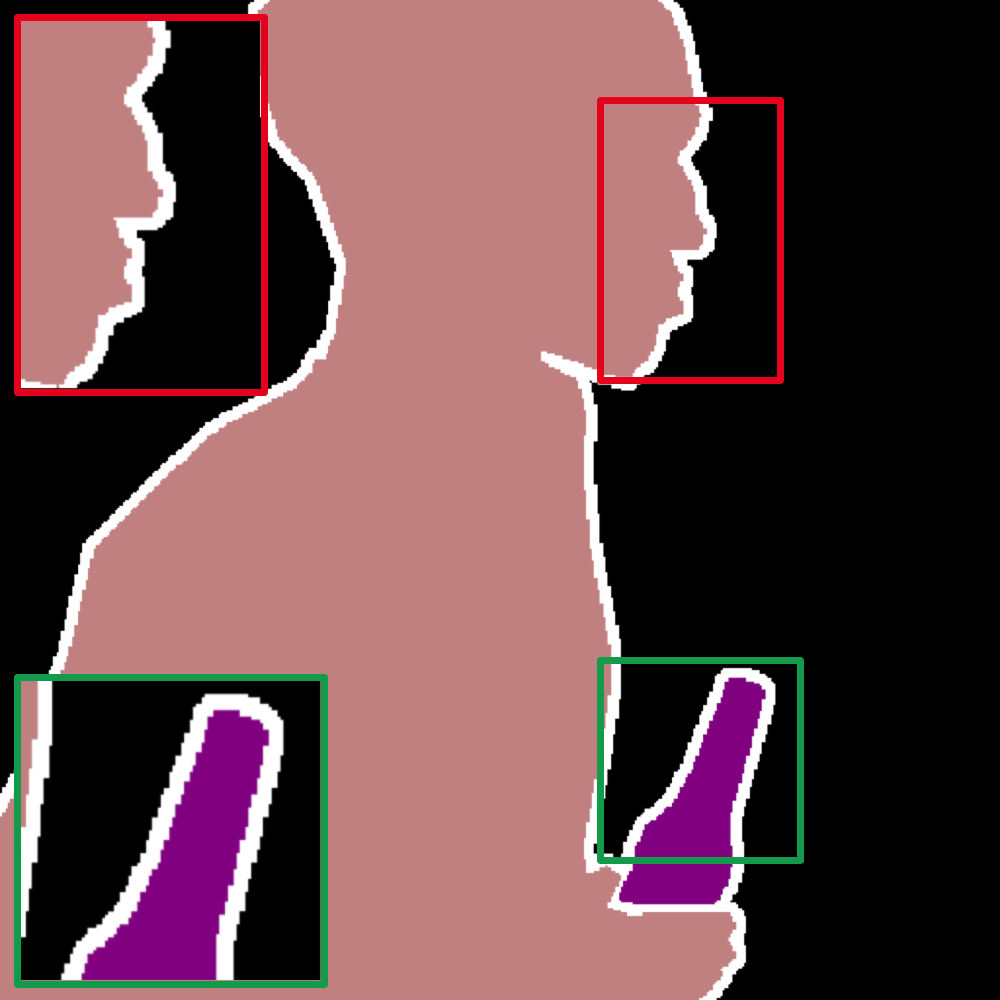}
\end{subfigure}
\\
\begin{subfigure}[b]{0.11\textwidth}
\includegraphics[width=\textwidth]{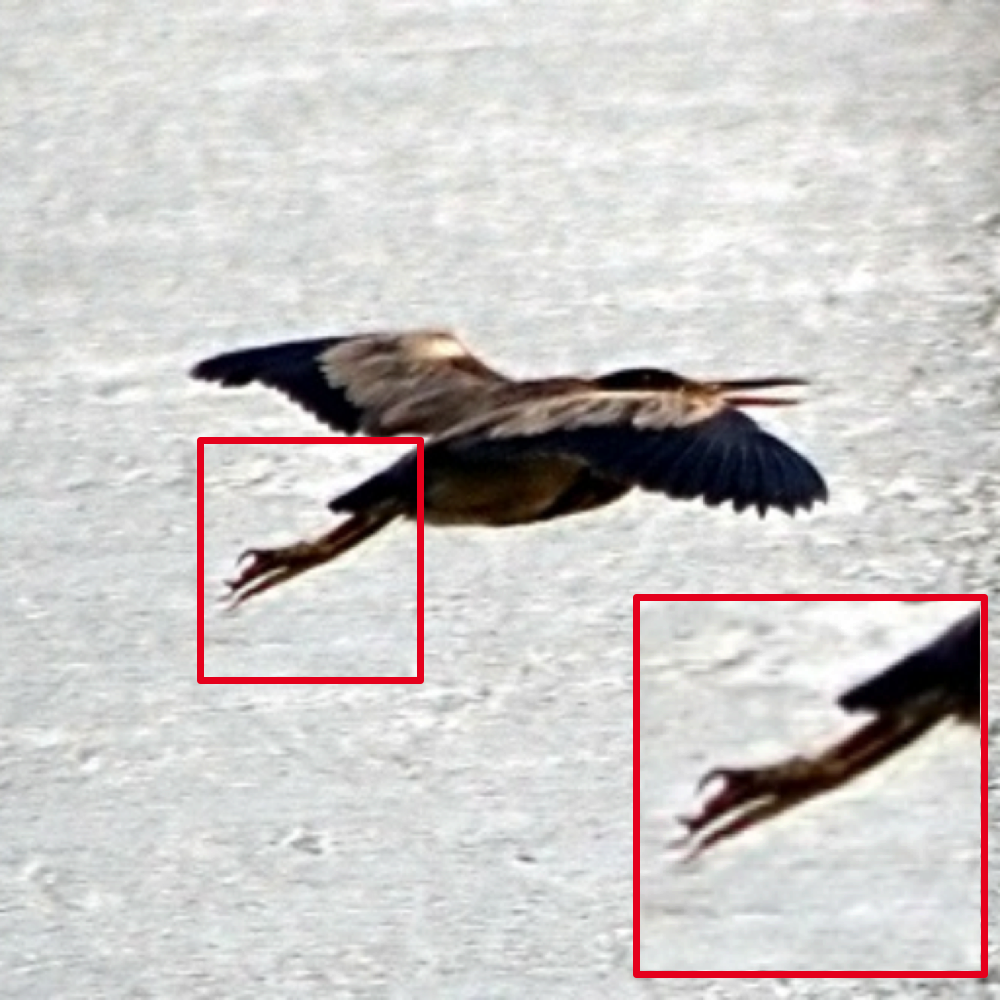}
\end{subfigure}
\begin{subfigure}[b]{0.11\textwidth}
\includegraphics[width=\textwidth]{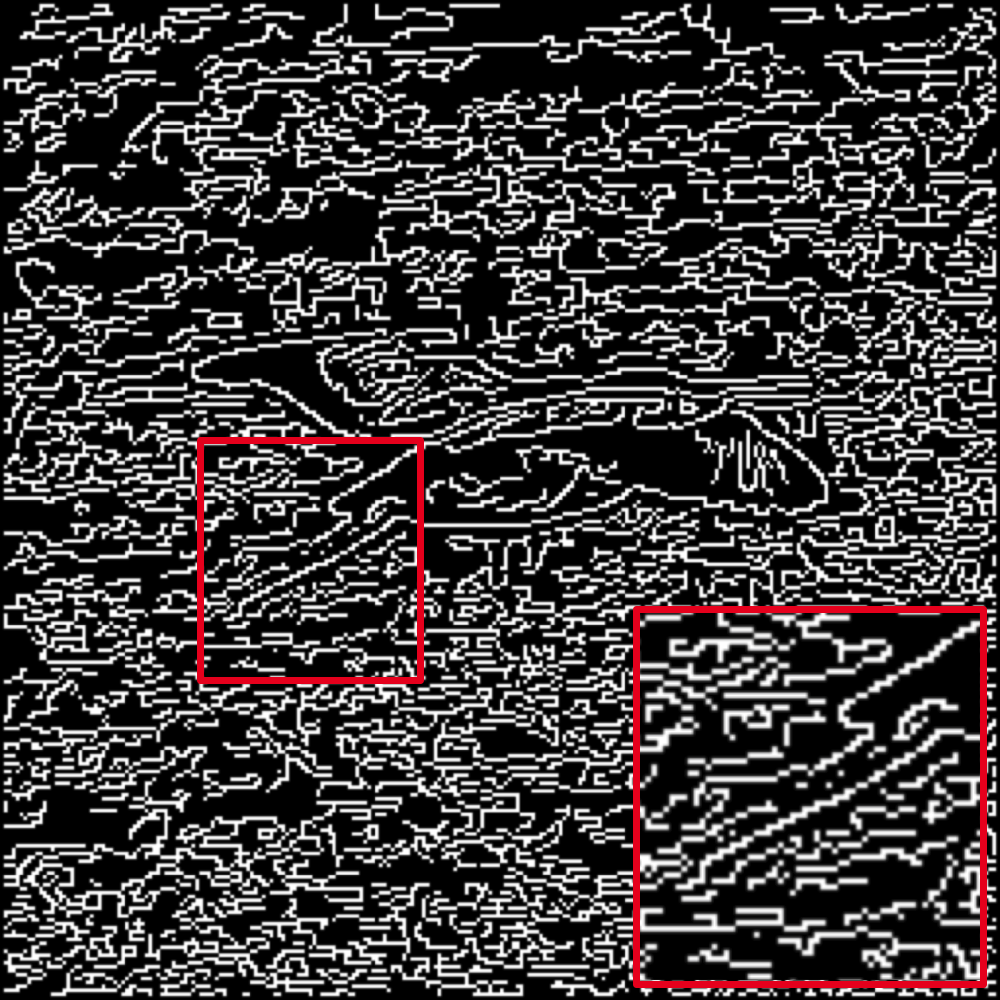}
\end{subfigure}
\begin{subfigure}[b]{0.11\textwidth}
\includegraphics[width=\textwidth]{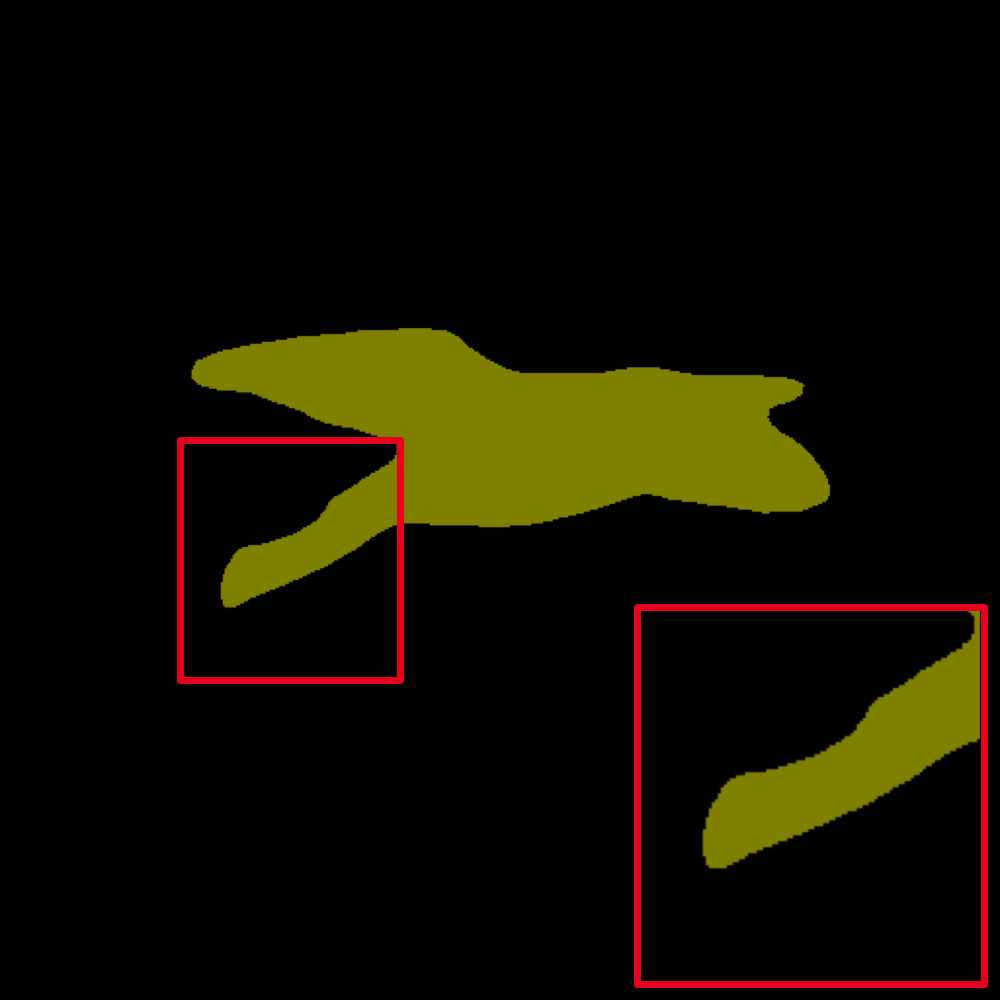}
\end{subfigure}
\begin{subfigure}[b]{0.11\textwidth}
\includegraphics[width=\textwidth]{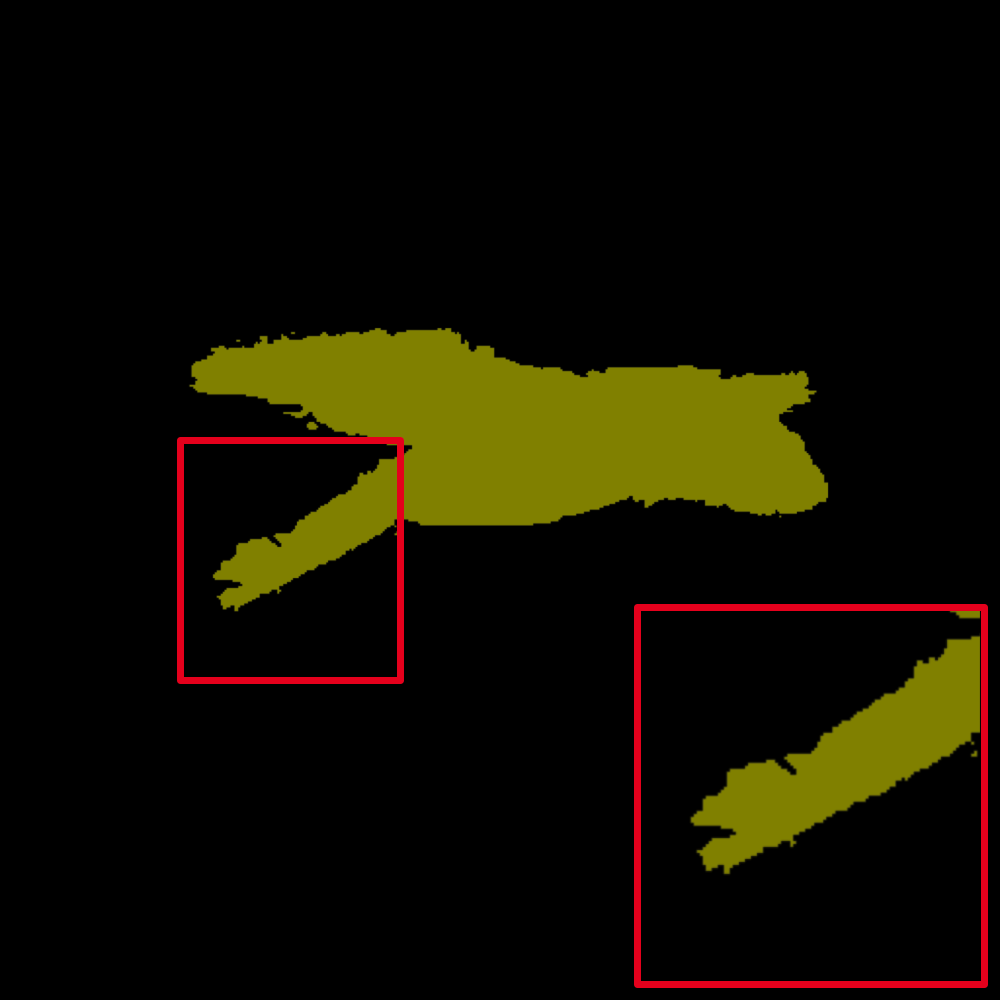}
\end{subfigure}
\begin{subfigure}[b]{0.11\textwidth}
\includegraphics[width=\textwidth]{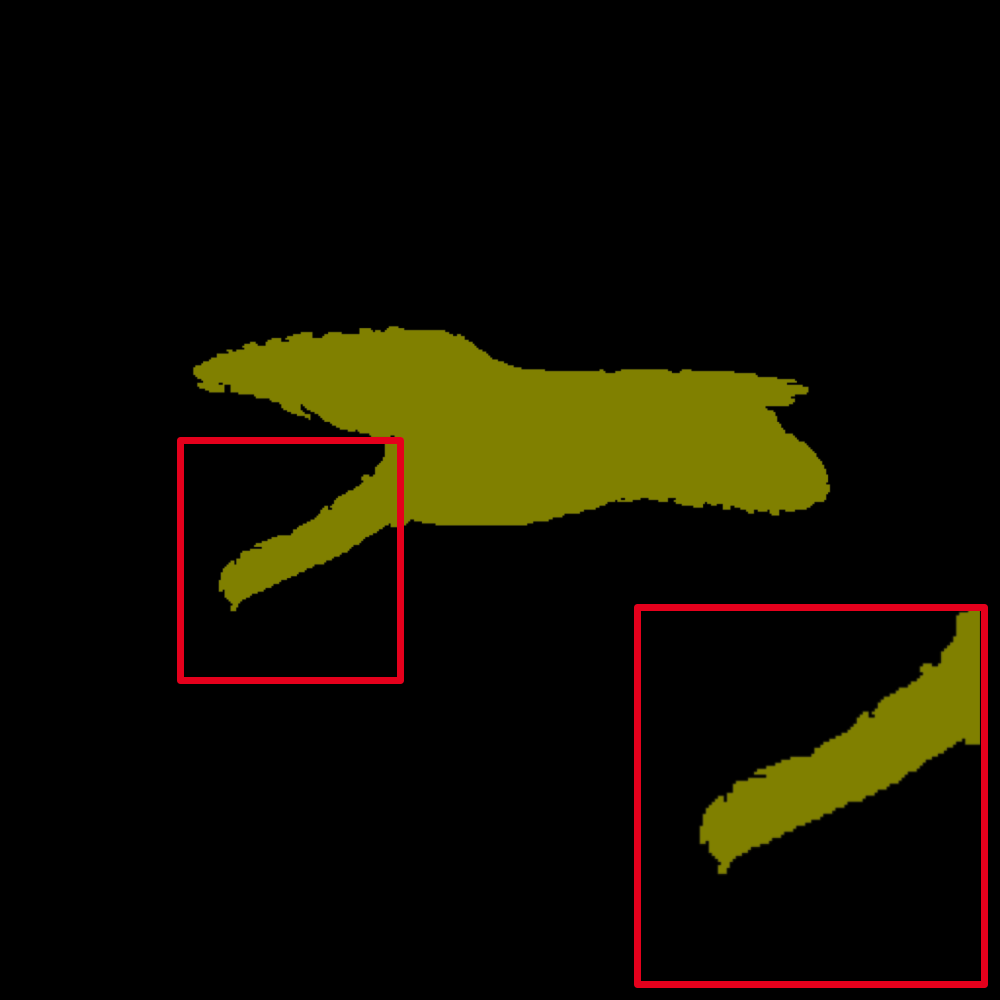}
\end{subfigure}
\begin{subfigure}[b]{0.11\textwidth}
\includegraphics[width=\textwidth]{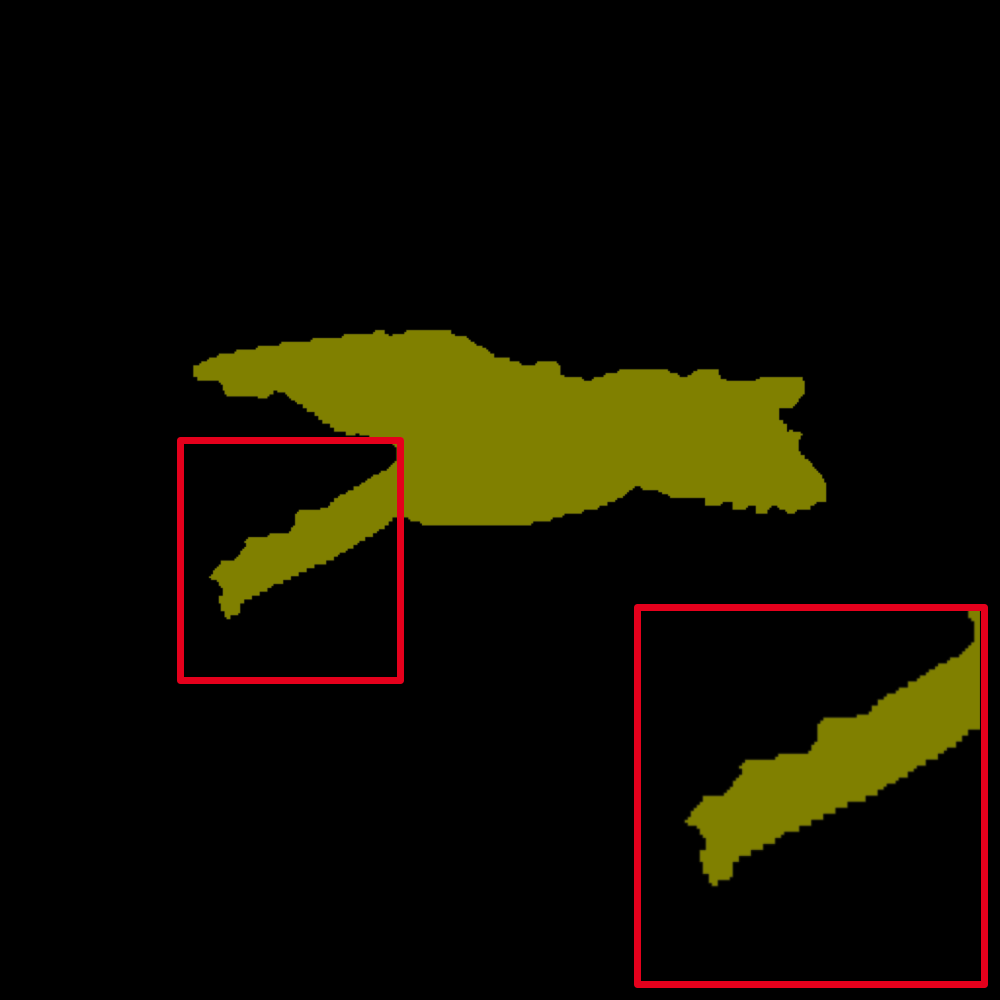}
\end{subfigure}
\begin{subfigure}[b]{0.11\textwidth}
\includegraphics[width=\textwidth]{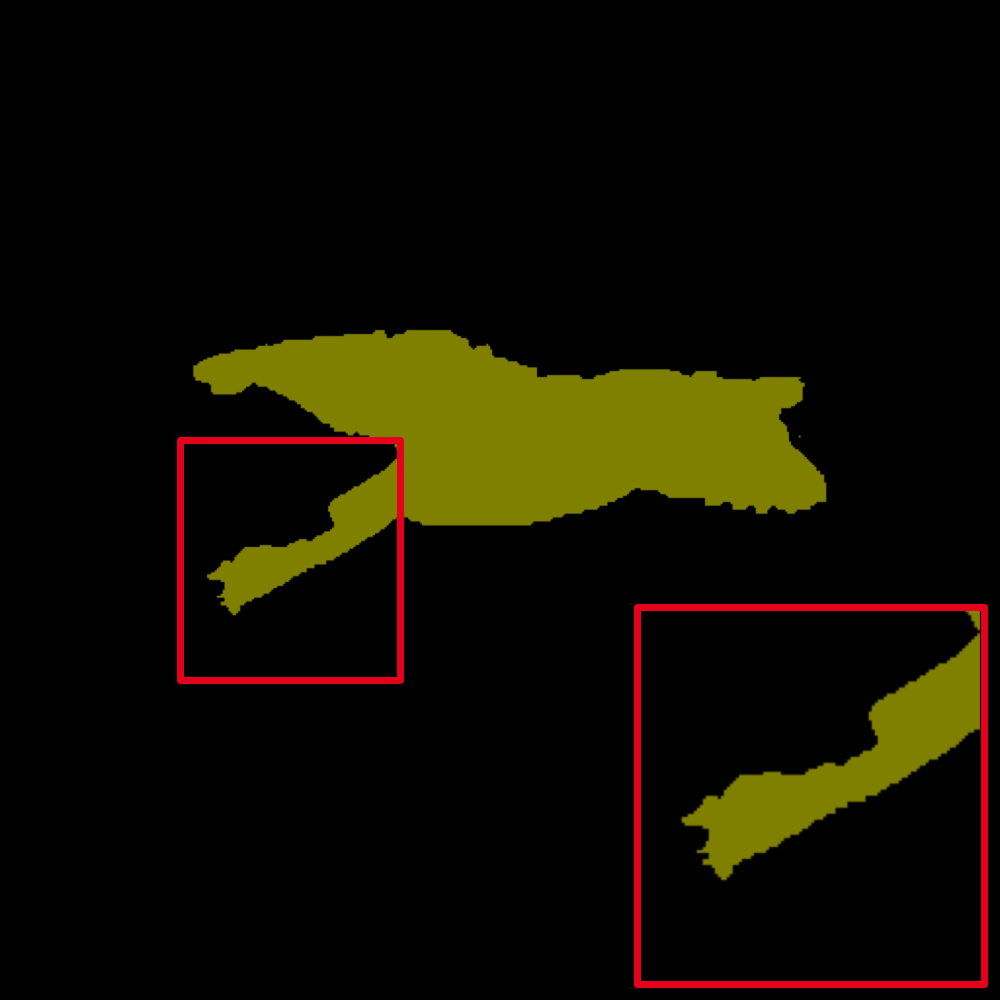}
\end{subfigure}
\begin{subfigure}[b]{0.11\textwidth}
\includegraphics[width=\textwidth]{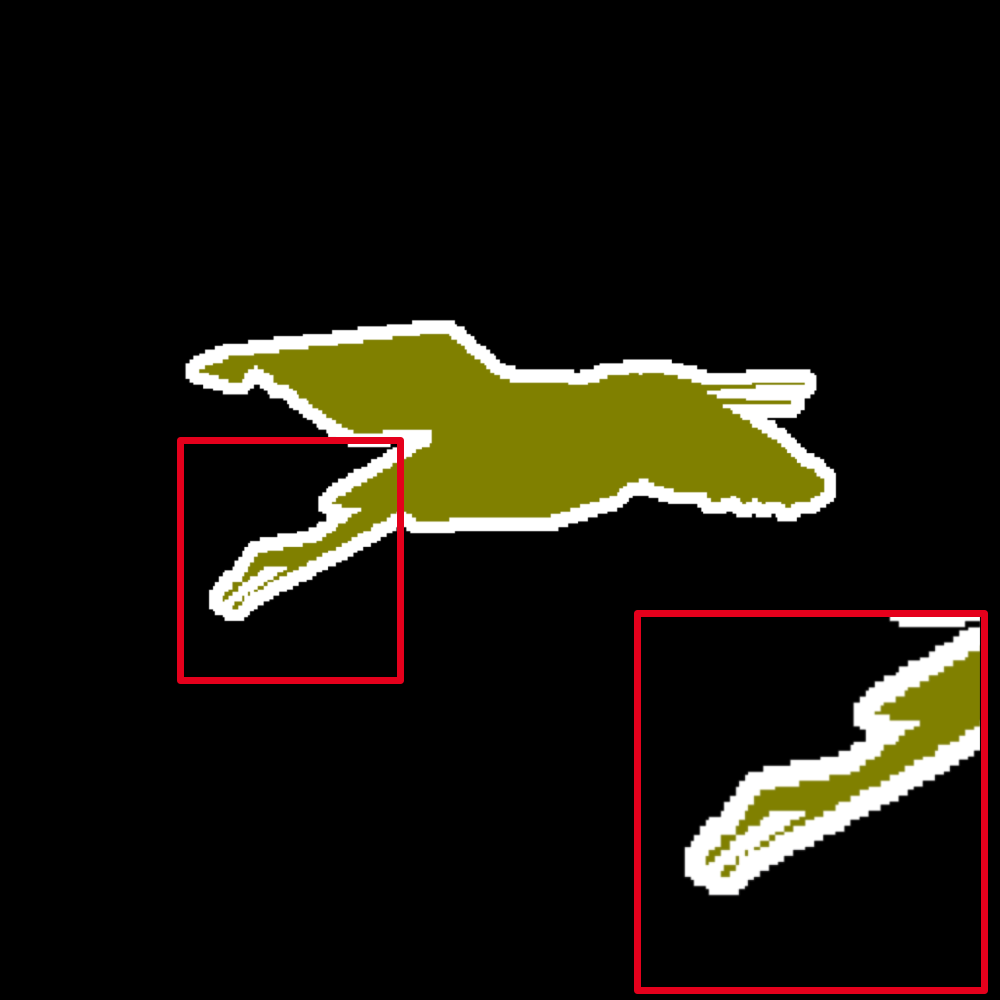}
\end{subfigure}
\\
\begin{subfigure}[b]{0.11\textwidth}
\includegraphics[width=\textwidth]{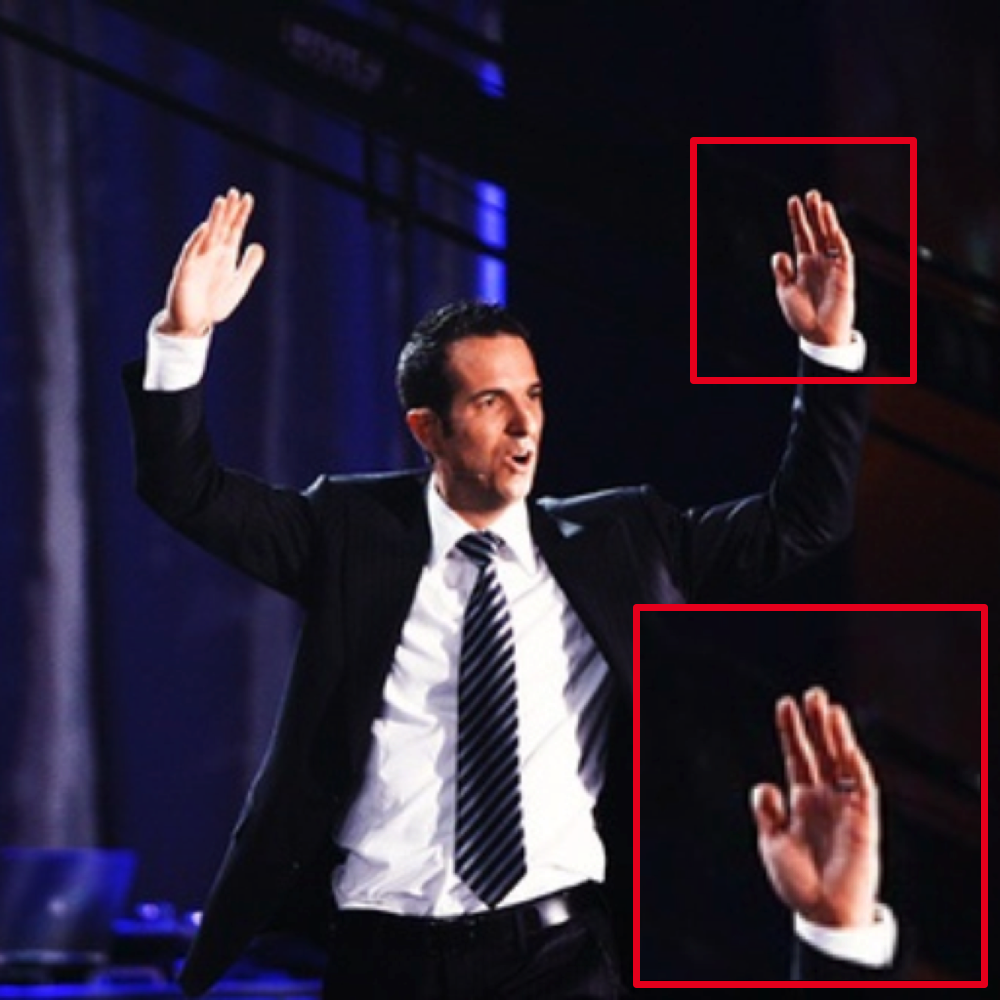}
\end{subfigure}
\begin{subfigure}[b]{0.11\textwidth}
\includegraphics[width=\textwidth]{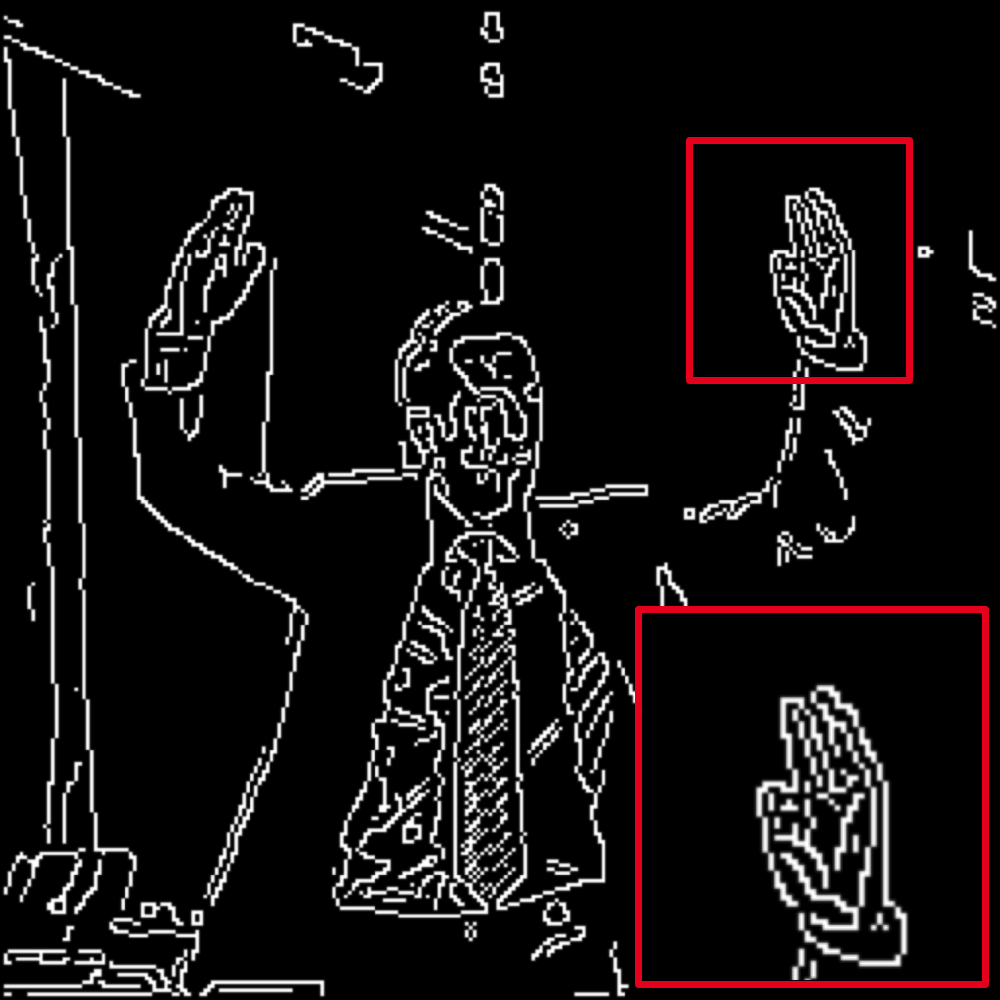}
\end{subfigure}
\begin{subfigure}[b]{0.11\textwidth}
\includegraphics[width=\textwidth]{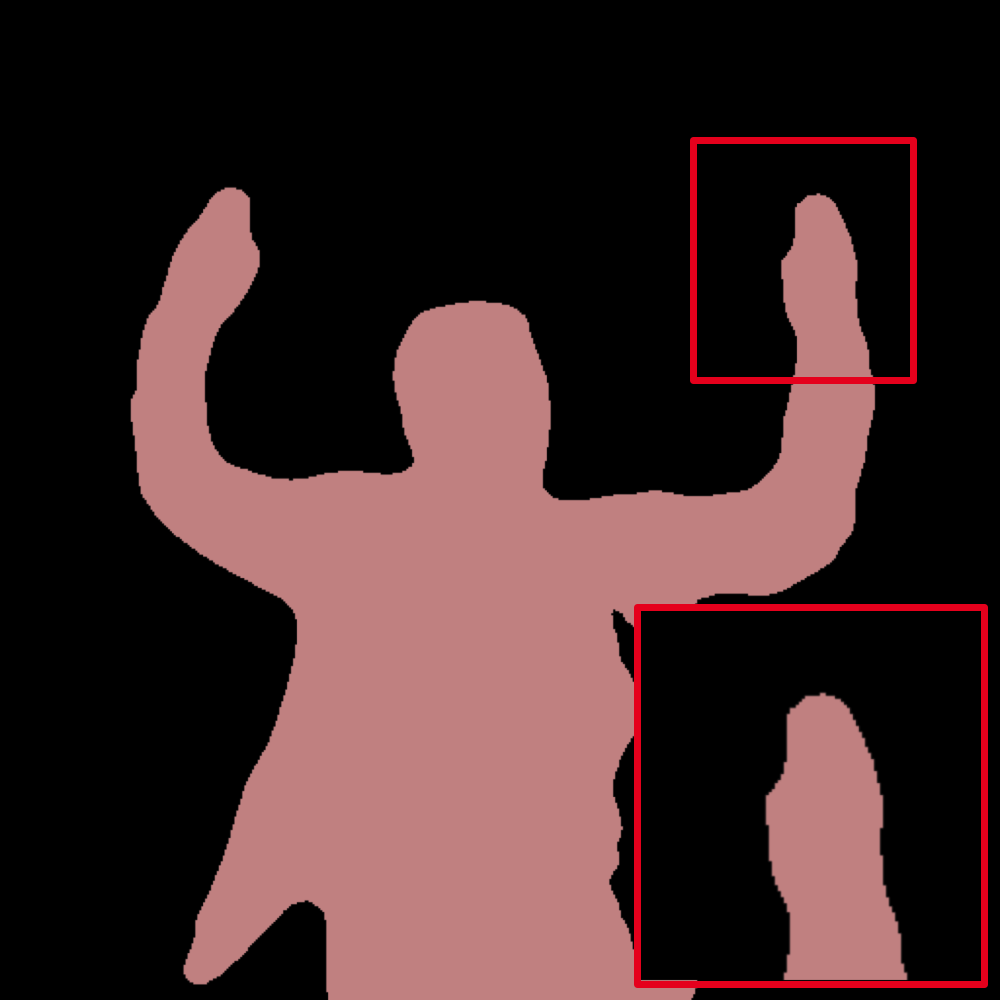}
\end{subfigure}
\begin{subfigure}[b]{0.11\textwidth}
\includegraphics[width=\textwidth]{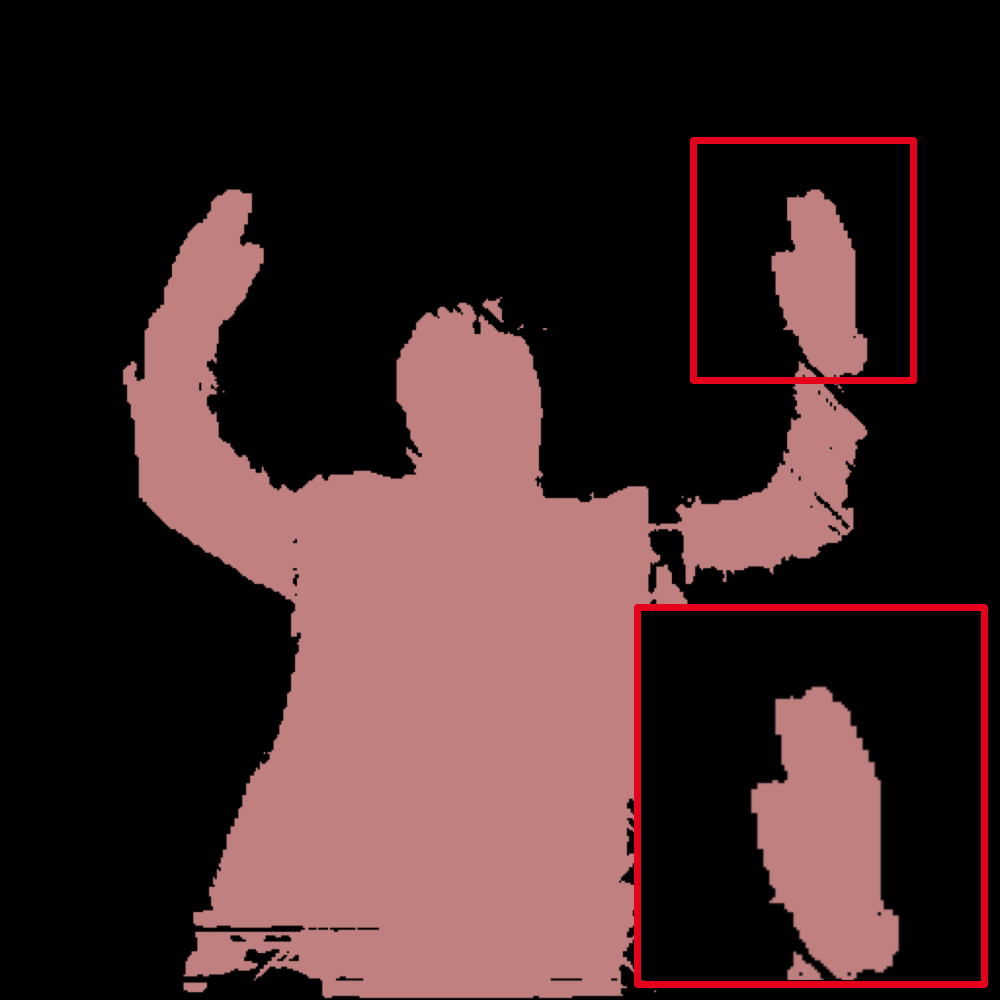}
\end{subfigure}
\begin{subfigure}[b]{0.11\textwidth}
\includegraphics[width=\textwidth]{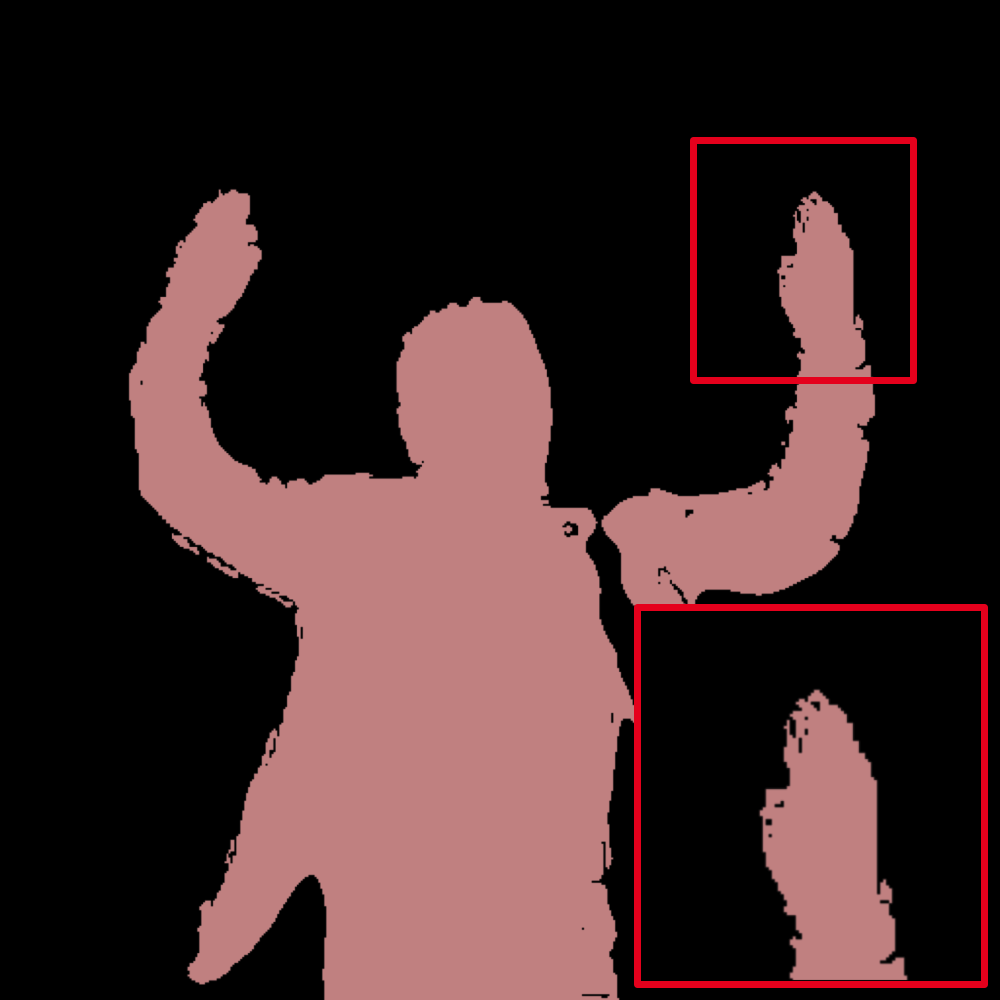}
\end{subfigure}
\begin{subfigure}[b]{0.11\textwidth}
\includegraphics[width=\textwidth]{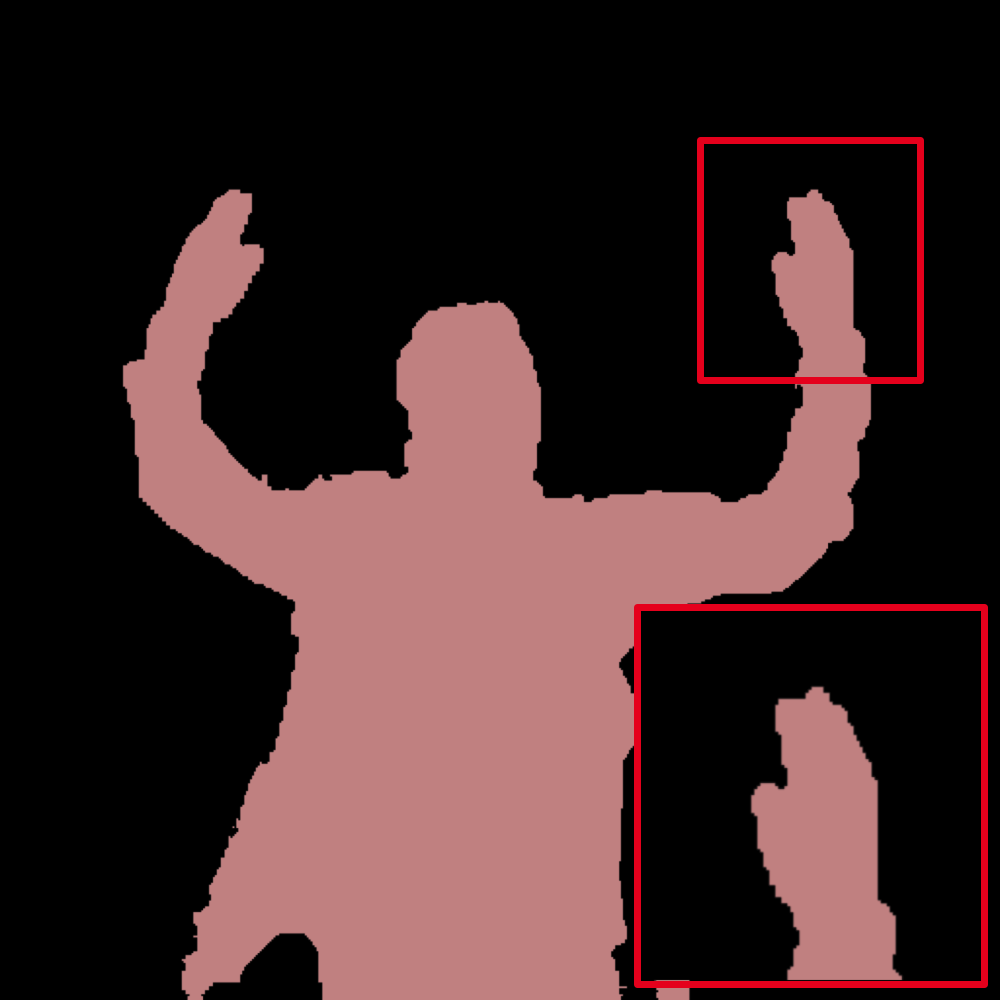}
\end{subfigure}
\begin{subfigure}[b]{0.11\textwidth}
\includegraphics[width=\textwidth]{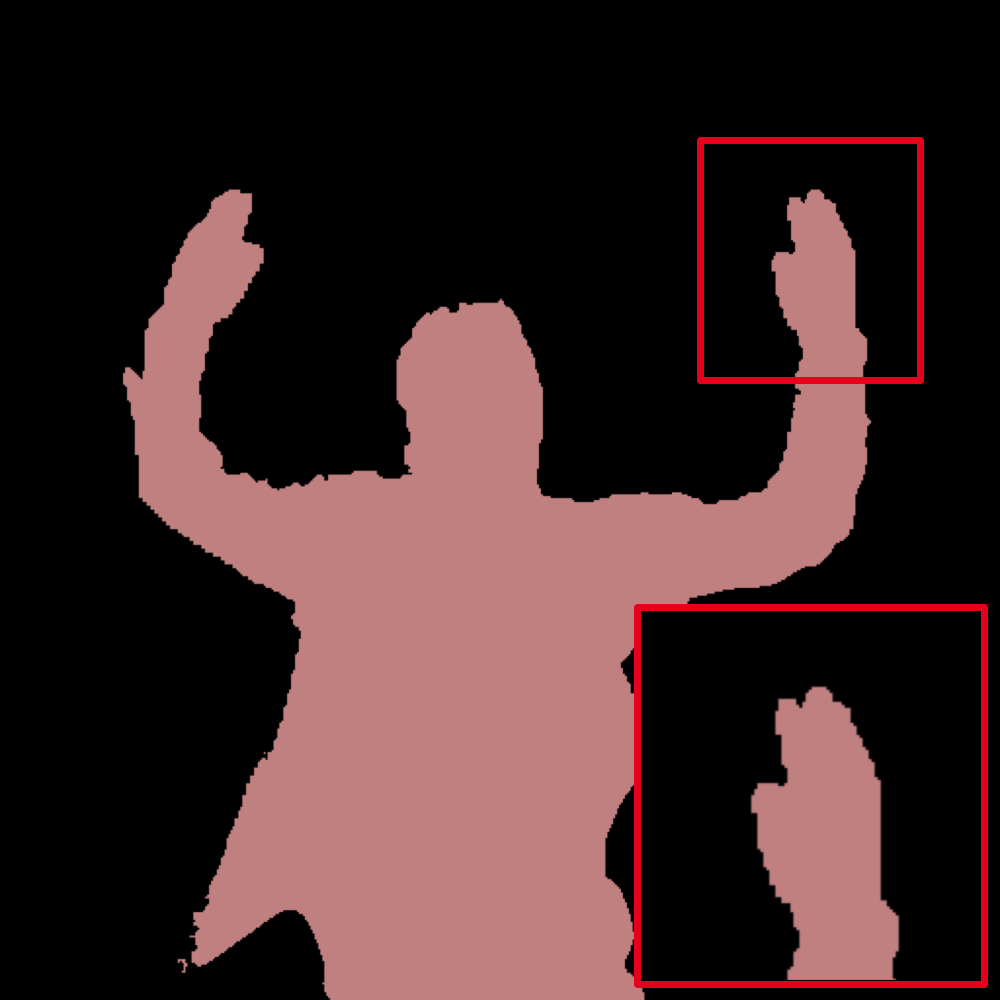}
\end{subfigure}
\begin{subfigure}[b]{0.11\textwidth}
\includegraphics[width=\textwidth]{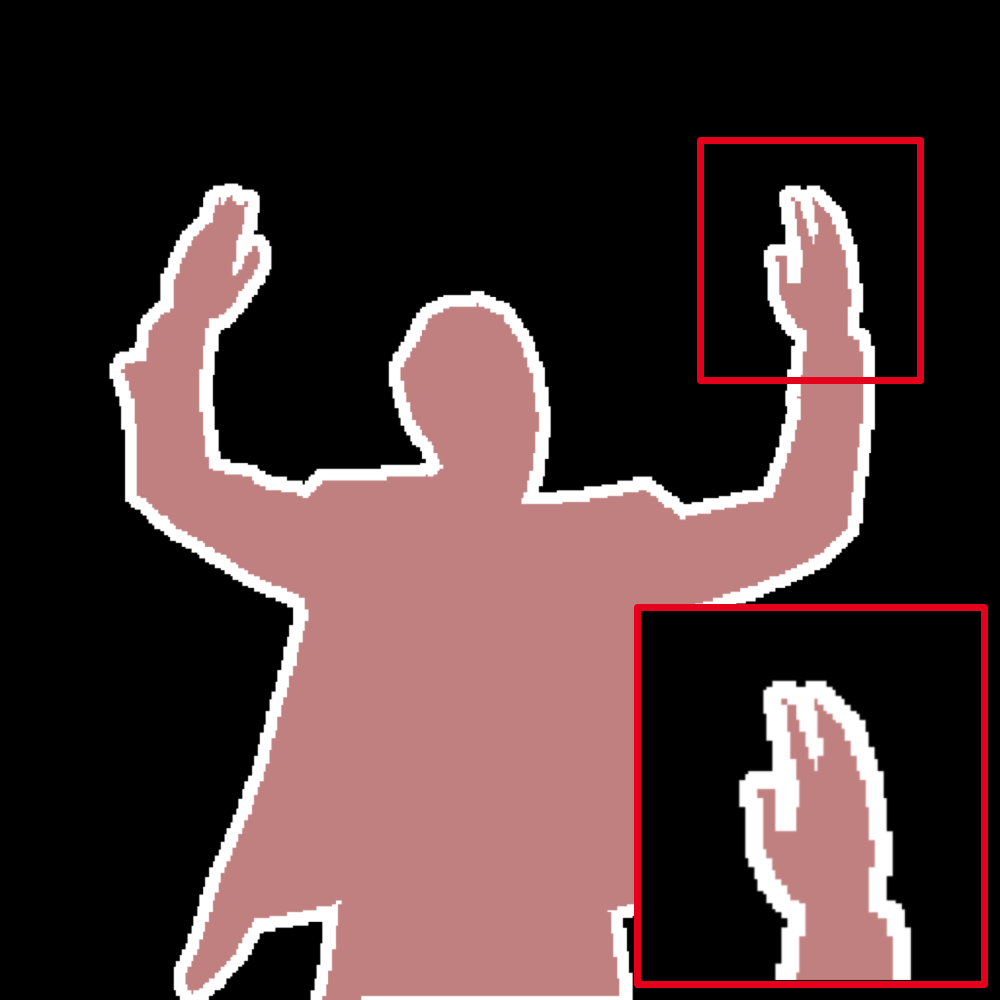}
\end{subfigure} \\
\begin{subfigure}[b]{0.11\textwidth}
\includegraphics[width=\textwidth]{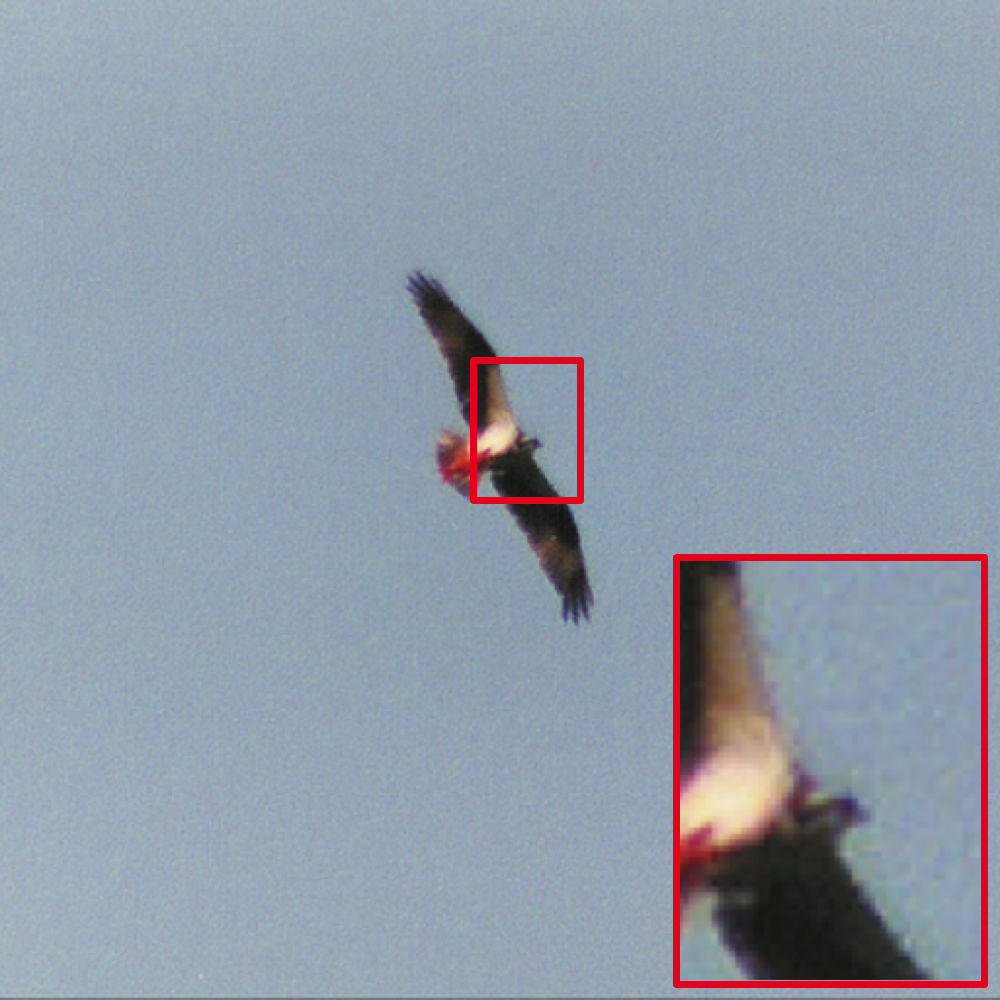}
\end{subfigure}
\begin{subfigure}[b]{0.11\textwidth}
\includegraphics[width=\textwidth]{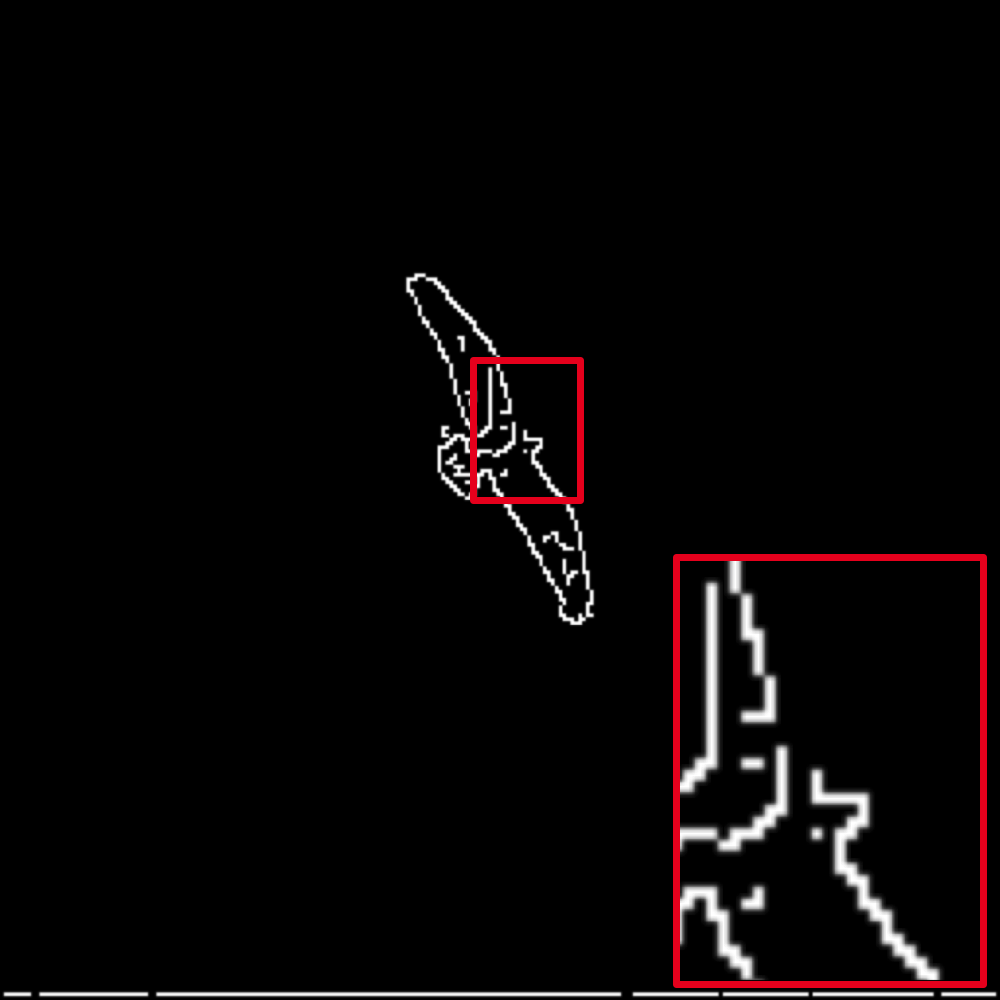}
\end{subfigure}
\begin{subfigure}[b]{0.11\textwidth}
\includegraphics[width=\textwidth]{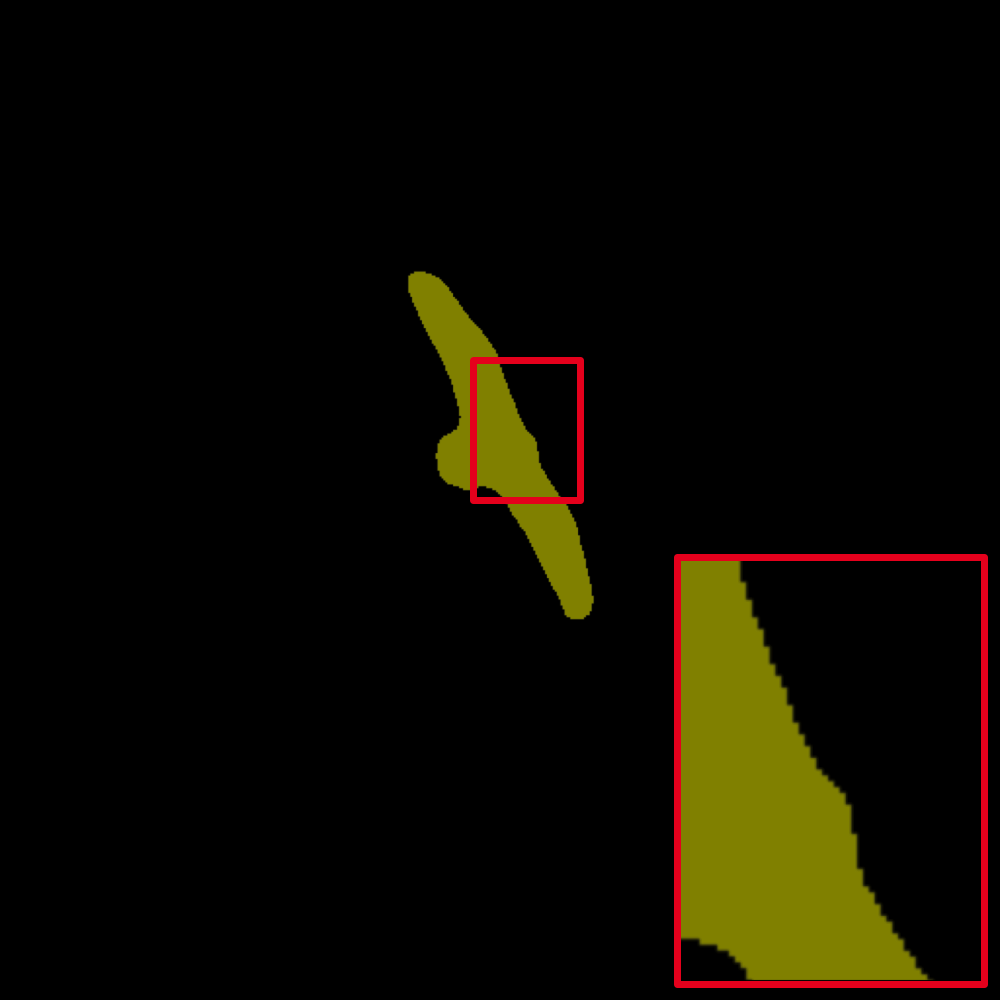}
\end{subfigure}
\begin{subfigure}[b]{0.11\textwidth}
\includegraphics[width=\textwidth]{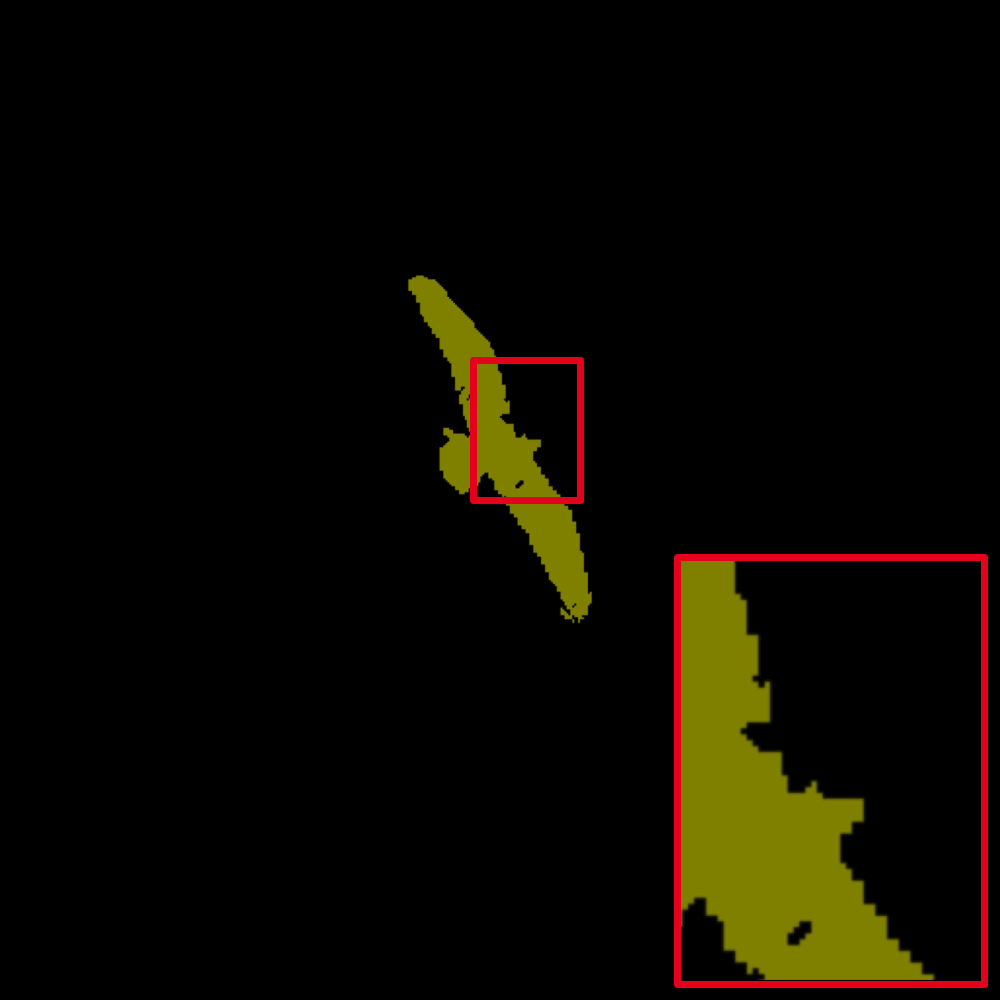}
\end{subfigure}
\begin{subfigure}[b]{0.11\textwidth}
\includegraphics[width=\textwidth]{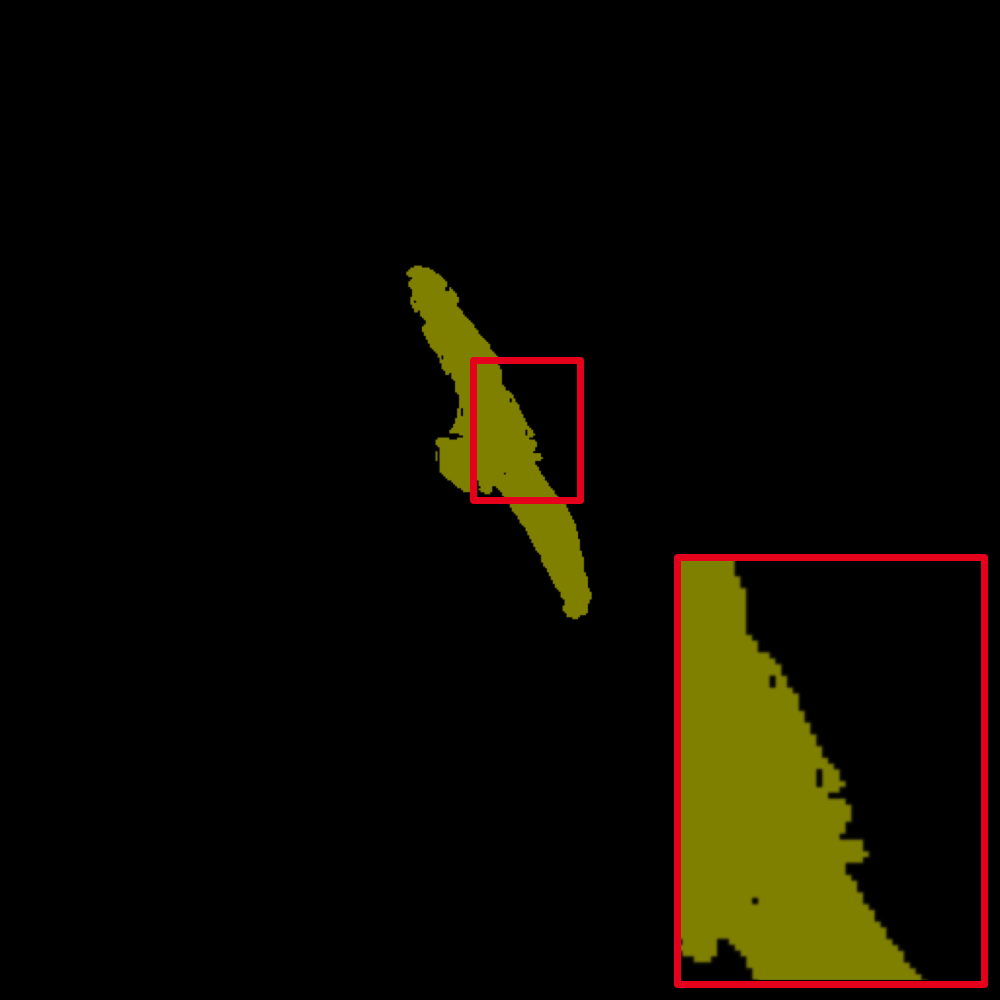}
\end{subfigure}
\begin{subfigure}[b]{0.11\textwidth}
\includegraphics[width=\textwidth]{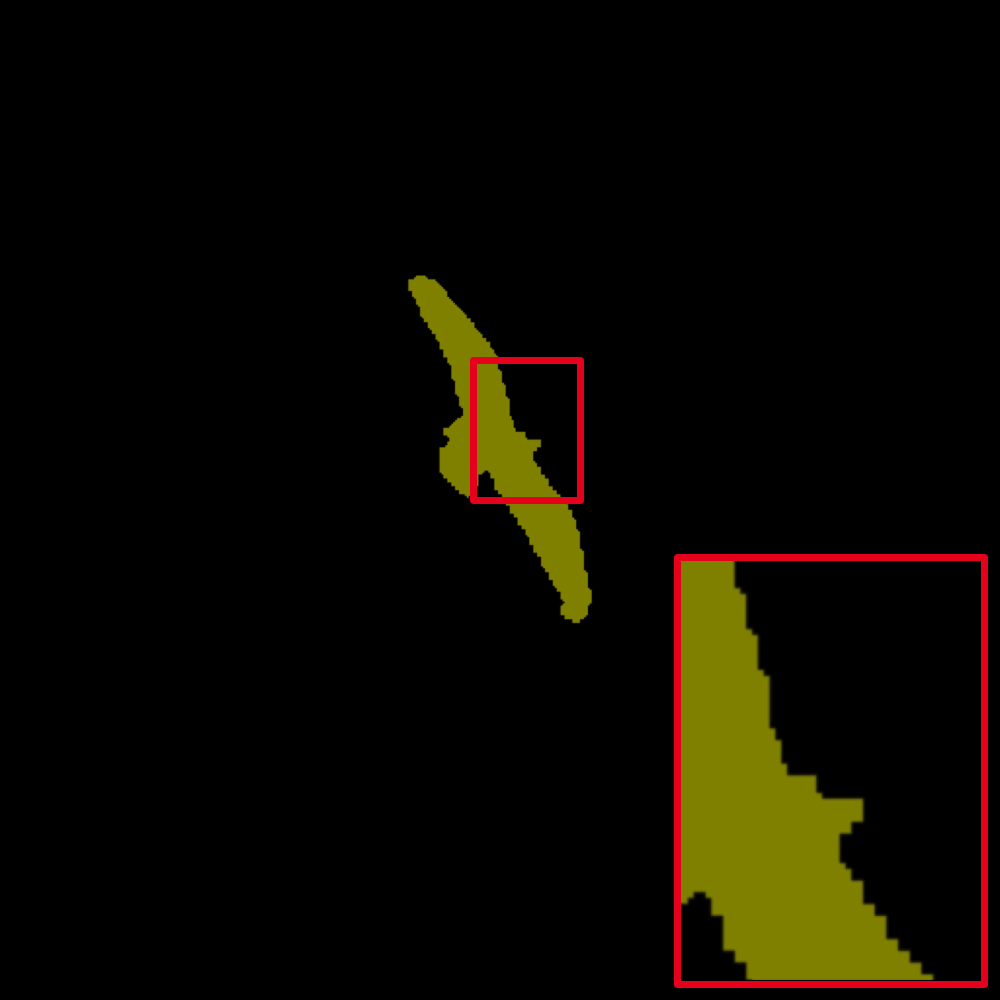}
\end{subfigure}
\begin{subfigure}[b]{0.11\textwidth}
\includegraphics[width=\textwidth]{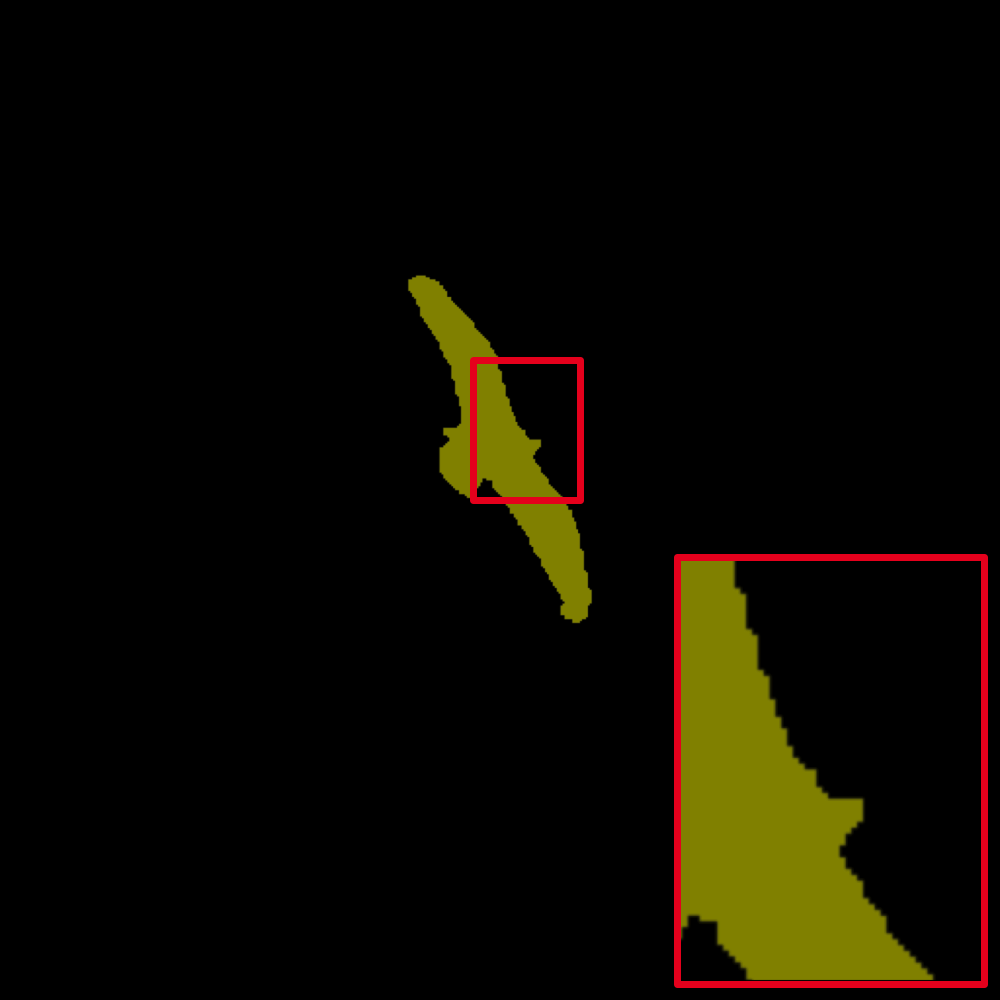}
\end{subfigure}
\begin{subfigure}[b]{0.11\textwidth}
\includegraphics[width=\textwidth]{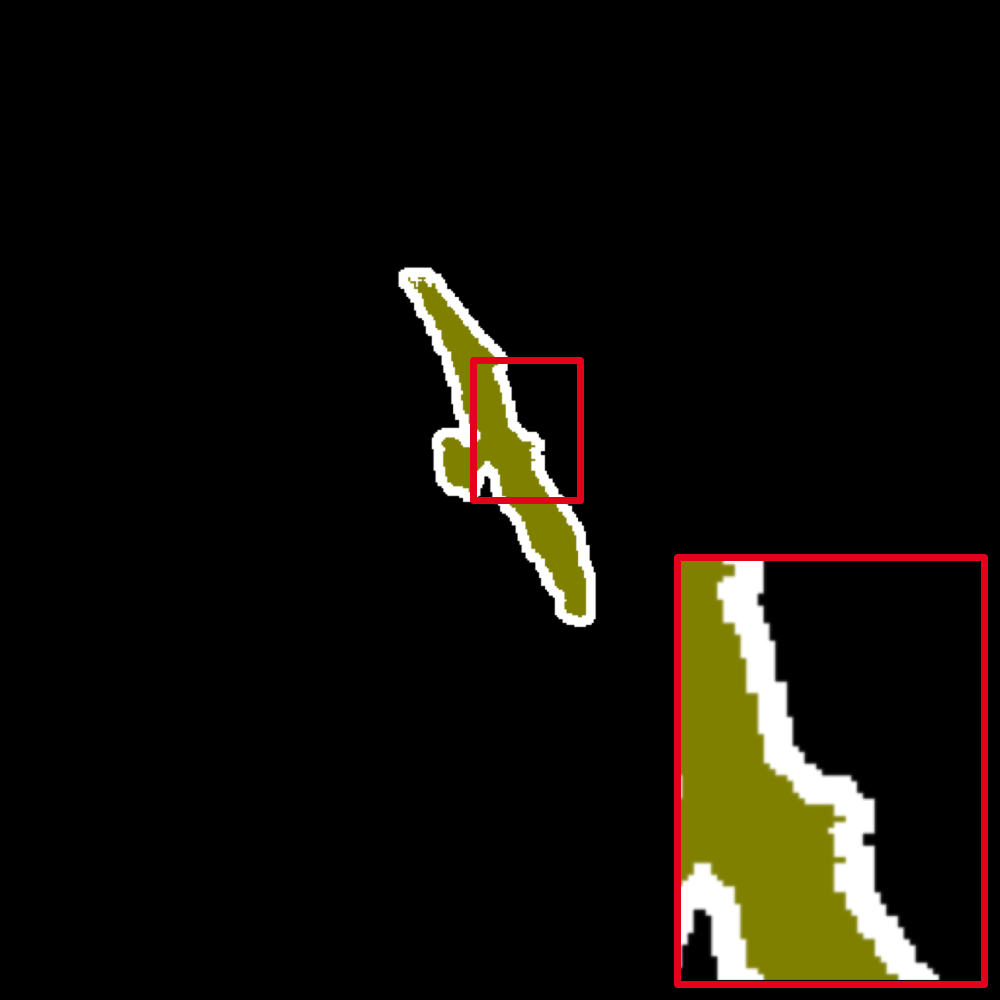}
\end{subfigure} \\
% failure case 1
\begin{subfigure}[b]{0.11\textwidth}
\includegraphics[width=\textwidth]{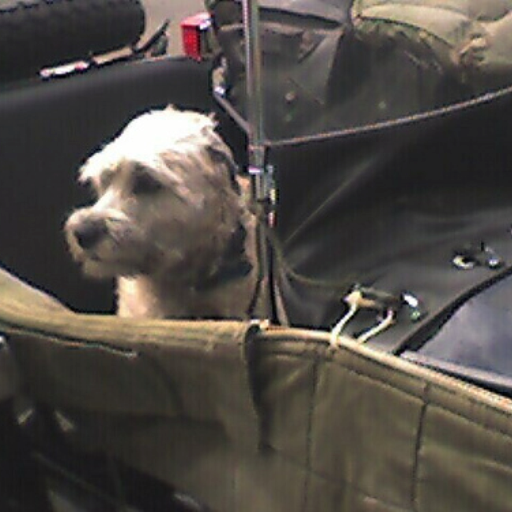}
\end{subfigure}
\begin{subfigure}[b]{0.11\textwidth}
\includegraphics[width=\textwidth]{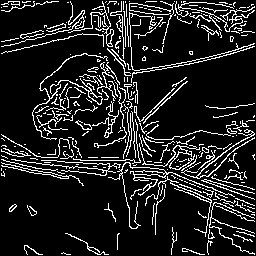}
\end{subfigure}
\begin{subfigure}[b]{0.11\textwidth}
\includegraphics[width=\textwidth]{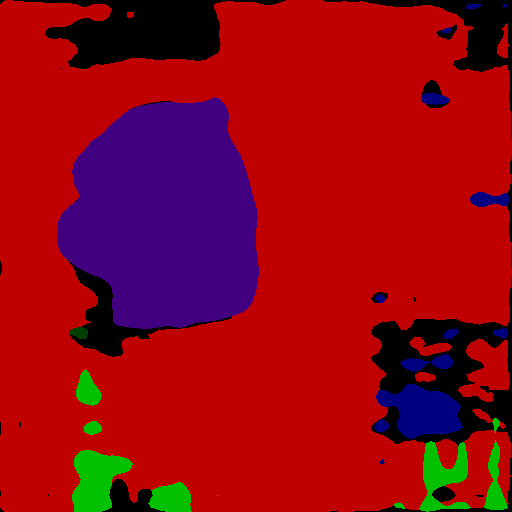}
\end{subfigure}
\begin{subfigure}[b]{0.11\textwidth}
\includegraphics[width=\textwidth]{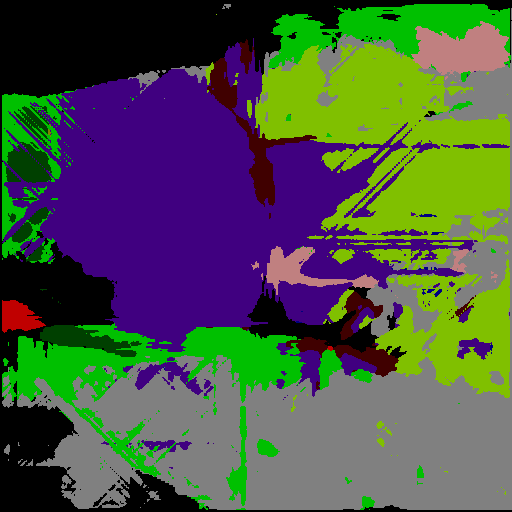}
\end{subfigure}
\begin{subfigure}[b]{0.11\textwidth}
\includegraphics[width=\textwidth]{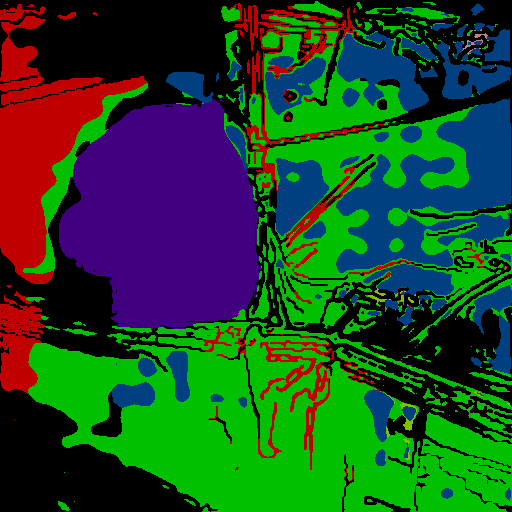}
\end{subfigure}
\begin{subfigure}[b]{0.11\textwidth}
\includegraphics[width=\textwidth]{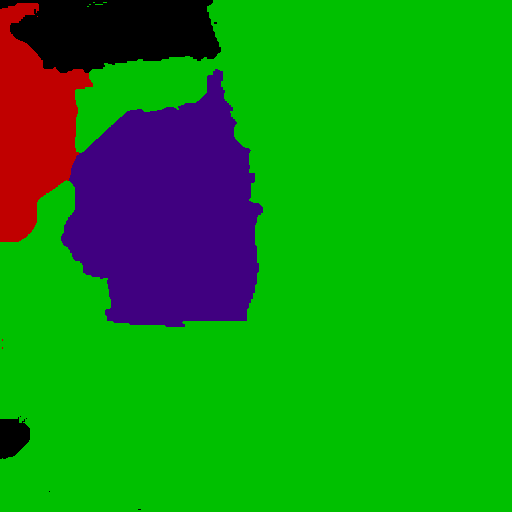}
\end{subfigure}
\begin{subfigure}[b]{0.11\textwidth}
\includegraphics[width=\textwidth]{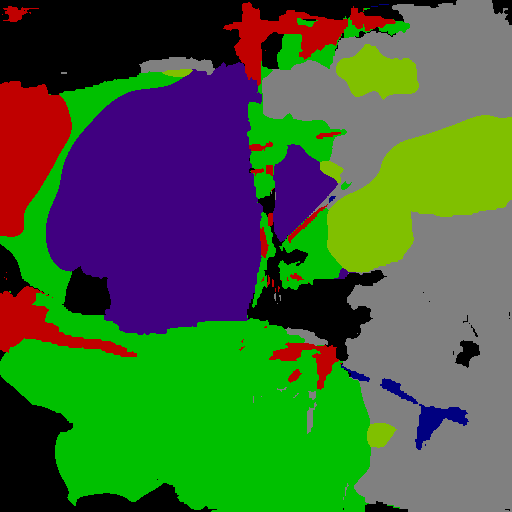}
\end{subfigure}
\begin{subfigure}[b]{0.11\textwidth}
\includegraphics[width=\textwidth]{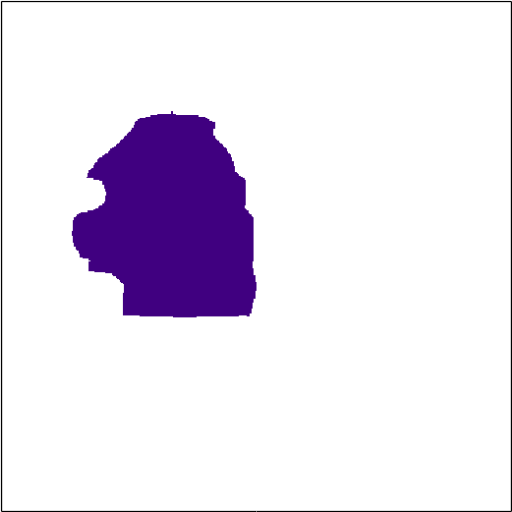}
\end{subfigure} \\
% failure case 2
\stackunder[5pt]{\includegraphics[width=0.11\textwidth]{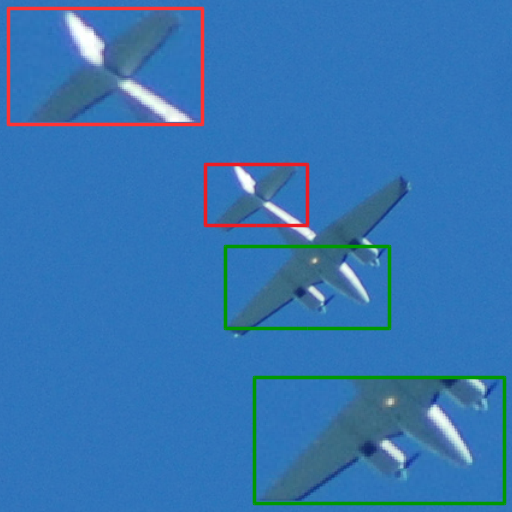}}{\scriptsize (a) RGB}
\stackunder[5pt]{\includegraphics[width=0.11\textwidth]{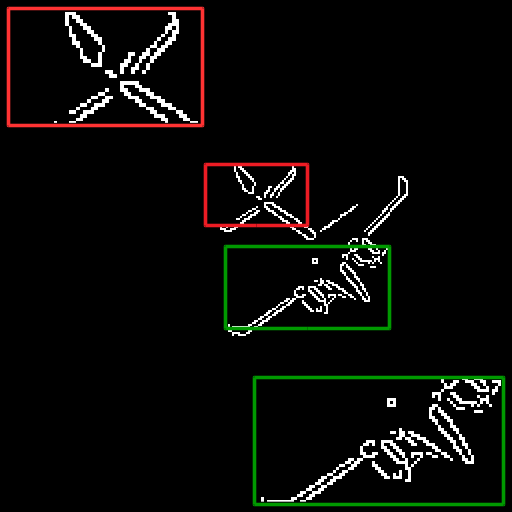}}{\scriptsize (b) Canny}
\stackunder[5pt]{\includegraphics[width=0.11\textwidth]{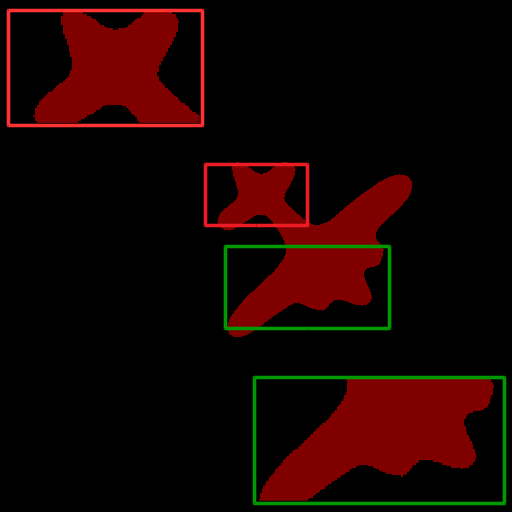}}{\scriptsize (c) baseline}
\stackunder[5pt]{\includegraphics[width=0.11\textwidth]{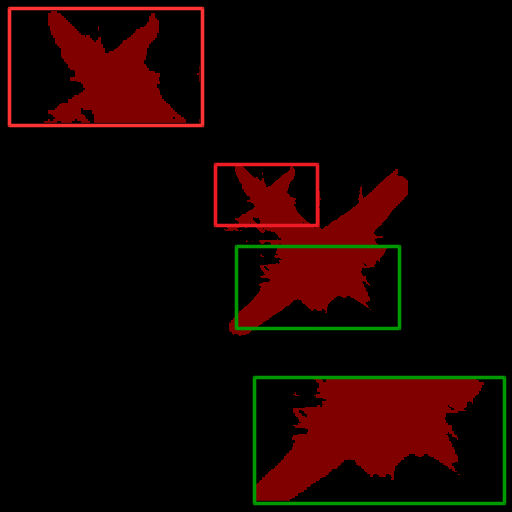}}{\scriptsize (d) SGM-8}
\stackunder[5pt]{\includegraphics[width=0.11\textwidth]{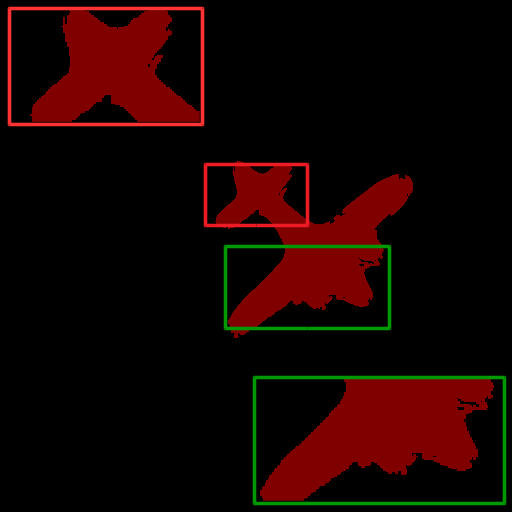}}{\scriptsize (e) MF-4}
\stackunder[5pt]{\includegraphics[width=0.11\textwidth]{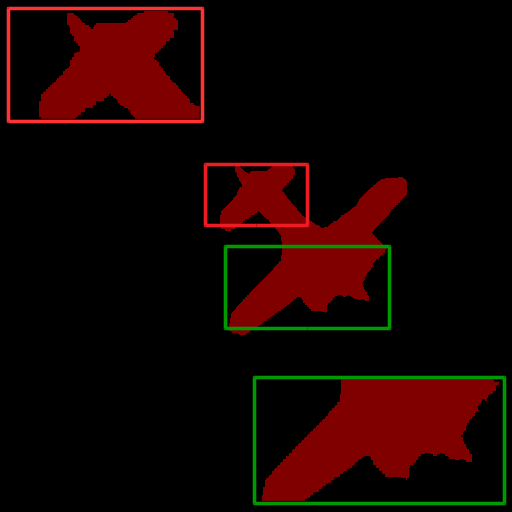}}{\scriptsize (f) ISGMR-8}
\stackunder[5pt]{\includegraphics[width=0.11\textwidth]{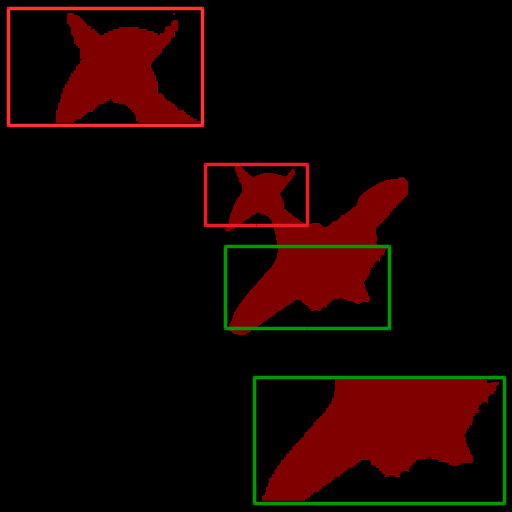}}{\scriptsize (g) TRWP-4}
\stackunder[5pt]{\includegraphics[width=0.11\textwidth]{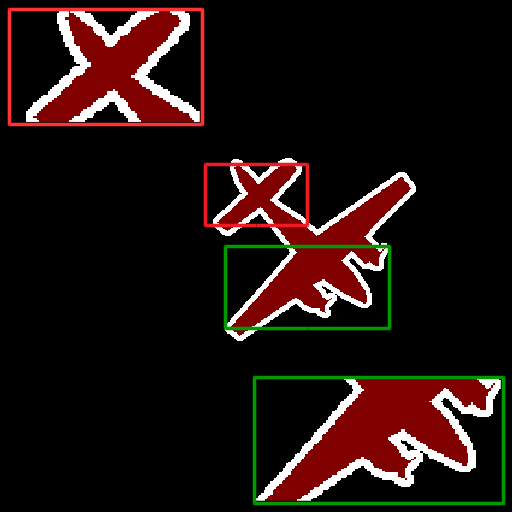}}{\scriptsize (h) GT}
\caption{Semantic segmentation on PASCAL VOC2012 val set.
Last two rows are failure cases due to poor unary terms and missing edges.
ISGMR-8 and TRWP-4 are ours.}
\label{fig:voc2012_val_vis}
\end{figure}

In \mytb{tb:voc2012_val}, ISGMR-8 and TRWP-4 outperform the baseline DeepLabV3+~\cite{DeepLabV3+},
SGM-8~\cite{sgm}, and MF-4~\cite{densecrf}.
Semantic segmentation by ISGMR-8 and TRWP-4 are more
sharp, accurate, and aligned with the Canny edges and ground-truth (GT) edges,
shown in white, than the other inference
methods, such as SGM-8 and MF-4 (see \myfig{fig:voc2012_val_vis}).

%--------------------------------------------------------------------------
\subsection{Speed Improvement} \label{sec:speed}

\begin{table}[t]
\centering
\caption{Forward propagation time with 32 and 96 
labels. 
Our CUDA version is averaged over 1000 trials; others over 100 trials.
Our CUDA version is
$7$--$32$ times faster than PyTorch GPU version.
C++ versions are with a single and 8 threads.
Unit: second.
}
\resizebox{\textwidth}{!}{
\begin{tabular}
{l||R{8mm}|R{8mm}|R{8mm}|R{6mm}|R{8mm}|R{9mm}|R{8mm}|R{8mm}|R{8mm}|R{8mm}|R{14mm}|R{14mm}}
  \hline
  \multicolumn{1}{c||}{\multirow{2}{*}{\textbf{Method}}}
  & \multicolumn{2}{c|}{\textbf{PyTorch CPU}}
  & \multicolumn{2}{c|}{\textbf{PyTorch GPU}}
  & \multicolumn{2}{c|}{\textbf{C++ single}}
  & \multicolumn{2}{c|}{\textbf{C++ multiple}}
  & \multicolumn{2}{c|}{\textbf{CUDA (ours)}}
  & \multicolumn{2}{c}{\textbf{Speed-up PyT/CUDA}} \\
  \cline{2-13}
  & \multicolumn{1}{c|}{\textbf{32}}
  & \multicolumn{1}{c|}{\textbf{96}}
  & \multicolumn{1}{c|}{\textbf{32}}
  & \multicolumn{1}{c|}{\textbf{96}}
  & \multicolumn{1}{c|}{\textbf{32}}
  & \multicolumn{1}{c|}{\textbf{96}}
  & \multicolumn{1}{c|}{\textbf{32}}
  & \multicolumn{1}{c|}{\textbf{96}}
  & \multicolumn{1}{c|}{\textbf{32}}
  & \multicolumn{1}{c|}{\textbf{96}}
  & \multicolumn{1}{c|}{\textbf{32}}
  & \multicolumn{1}{c}{\textbf{96}} \\
  \hline \hline
  TRWS-4 & \centering- & \centering- & \centering- & \centering- & 1.95
  & 13.30 & \centering- & \centering- & \centering- & \centering-
  & \centering- & \multicolumn{1}{c}{-} \\
  \hline
  ISGMR-4 & 1.43 & 11.70 & 0.96 & 1.13 & 3.23 & 25.19 & 0.88 & 5.28 & 0.03 &
0.15 & \lgray 32$\times$ & \lgray 8$\times$ \\
  ISGMR-8 & 3.18 & 24.78 & 1.59 & 1.98 & 8.25 & 71.35 & 2.12 & 15.90 & 0.07 &
0.27  & \lgray 23$\times$ & \lgray 7$\times$ \\
  ISGMR-16 & 7.89 & 52.76 & 2.34 & 4.96 & 30.76 & 273.68 & 7.70 & 62.72 & 0.13 &
0.53 & \lgray 18$\times$ & \lgray 9$\times$ \\
  \hline
  TRWP-4 & 1.40 & 11.74 & 0.87 & 1.08 & 1.84 & 15.41 & 0.76 & 4.46 & 0.03 & 0.15
& \lgray 29$\times$ & \lgray 7$\times$ \\
  TRWP-8 & 3.19 & 24.28 & 1.57 & 1.98 & 6.34 & 57.25 & 1.88 & 14.22 & 0.07 &
0.27 & \lgray 22$\times$ & \lgray 7$\times$ \\
  TRWP-16 & 7.86 & 51.85 & 2.82 & 5.08 & 28.93 & 262.28 & 7.41 & 60.45 & 0.13 &
0.52 & \lgray 22$\times$ & \lgray 10$\times$ \\
  \hline
\end{tabular}}
\label{tb:forward-time}
\end{table}

\begin{table}
\centering
\begin{minipage}[c]{0.6\textwidth}
\resizebox{\textwidth}{!}
{\begin{tabular}{l||R{8mm}|R{9mm}|R{8mm}|R{8mm}|R{14mm}|R{14mm}}
  \hline
  \multicolumn{1}{c||}{\multirow{2}{*}{\textbf{Method}}}
  & \multicolumn{2}{c|}{\textbf{PyTorch GPU}}
  & \multicolumn{2}{c|}{\textbf{CUDA (ours)}}
  & \multicolumn{2}{c}{\textbf{Speed-up PyT/CUDA}} \\
  \cline{2-7}
  & \multicolumn{1}{c|}{\textbf{32}}
  & \multicolumn{1}{c|}{\textbf{96}}
  & \multicolumn{1}{c|}{\textbf{32}}
  & \multicolumn{1}{c|}{\textbf{96}}
  & \multicolumn{1}{c|}{\textbf{32}}
  & \multicolumn{1}{c}{\textbf{96}} \\
  \hline \hline
  ISGMR-4 & 7.38 & 21.48 & 0.01 & 0.03 & \lgray 738$\times$ & \lgray 716$\times$ \\
  ISGMR-8 & 18.88 & 55.92 & 0.02 & 0.07 & \lgray 944$\times$ & \lgray 799$\times$ \\
  ISGMR-16 & 58.23 & 173.02 & 0.06 & 0.18 & \lgray 971$\times$ & \lgray 961$\times$ \\
  \hline
  TRWP-4 & 7.35 & 21.45 & 0.01 & 0.02 & \lgray 735$\times$ & \lgray 1073$\times$ \\
  TRWP-8 & 18.86 & 55.94 & 0.02 & 0.06 & \lgray 943$\times$ & \lgray 932$\times$ \\
  TRWP-16 & 58.26 & 172.95 & 0.06 & 0.16 & \lgray 971$\times$ & \lgray 1081$\times$ \\
  \hline
\end{tabular}}
\end{minipage}
\begin{minipage}[c]{0.33\textwidth}
\caption{Backpropagation time.
PyTorch GPU is averaged on 10 trials and CUDA on 1000 trials.
Ours is $716$--$1081$ times faster than PyTorch GPU.
Unit: second.}
\label{tb:backward-time}
\end{minipage}
\end{table}

Speed-up by parallelized message passing on a GPU enables a fast inference and
end-to-end learning.
To be clear, we compared forward and backward propagation times for different implementations
using 256$\times$512 size images with 32 and 96 labels.

\textbf{Forward propagation time.} In \mytb{tb:forward-time}, the forward
propagation by CUDA implementation is the fastest. Our CUDA versions of ISGMR-8
and TRWP-4 are at least 24 and 7 times faster than PyTorch GPU versions at 32
and 96 labels respectively.
In PyTorch GPU versions, we used tensor-wise tree parallelization to highly
speed it up for a fair comparison. Obviously, GPU versions are much faster than
CPU versions.

\textbf{Backpropagation time.} In \mytb{tb:backward-time}, the backpropagation
time clearly distinguishes the higher efficiency of CUDA versions than PyTorch
GPU versions. On average, the CUDA versions are at least $700$ times faster
than PyTorch GPU versions, and only a low memory is used to store indices for
backpropagation.
This makes the backpropagation much faster than the forward propagation and enables its feasibility in deep learning.
Analysis of PyTorch GPU version and our CUDA
implementation are in Appendix D.4.

%==========================================================================
\section{Conclusion}
In this paper, we introduce two fast and differentiable message passing
algorithms, namely, ISGMR and TRWP.
While ISGMR improved the effectiveness of SGM, TRWP sped up TRWS by two
orders of magnitude without loss of solution quality.
Besides, our CUDA implementations achieved at least $7$ times
and $700$ times speed-up compared to PyTorch GPU versions in the forward
and backward
propagation respectively.
These enable end-to-end learning with effective and
efficient MRF optimization algorithms.
Experiments of stereo vision and image denoising as well as
end-to-end learning for semantic segmentation validated the effectiveness
and efficiency of our proposals.\\

%--------------------------------------------------------------------------
\noindent \textbf{Acknowledgement:}
We would like to thank our colleagues Dylan Campbell and Yao Lu for the
discussion of CUDA programming.
This work is supported by the Australian Centre for Robotic Vision
(CE140100016) and Data61, CSIRO, Canberra, Australia.
    \section*{Appendix}
    \appendix
    
    \bibliographystyle{splncs}
    \bibliography{main}
\fi
\end{document}